\documentclass[10pt,twocolumn,letterpaper]{article}

\usepackage{cvpr}
\usepackage{times}
\usepackage{epsfig}
\usepackage{graphicx}
\usepackage{amsmath}
\usepackage{amssymb}
\usepackage{float}
\usepackage{algorithm}  
\usepackage{algorithmicx}
\usepackage{algpseudocode}

\usepackage{subcaption}
\usepackage{authblk}

\usepackage[breaklinks=true,bookmarks=false]{hyperref}

\cvprfinalcopy 

\def\cvprPaperID{****} 

\begin{document}

\title{\vspace{-1cm}Generative Image Inpainting with Contextual Attention}

\author[1]{Jiahui Yu}
\author[2]{Zhe Lin}
\author[2]{Jimei Yang}
\author[2]{Xiaohui Shen}
\author[2]{Xin Lu}
\author[1]{Thomas S. Huang}

\affil[1]{University of Illinois at Urbana-Champaign}
\affil[2]{Adobe Research}
\renewcommand\Authands{\ \ \ \ \ }
\renewcommand{\Authsep}{\ \ \ \ \ }

\twocolumn[{%
\renewcommand\twocolumn[1][]{#1}%
\maketitle
\vspace{-8mm}\begin{center}
\noindent
\begin{minipage}{.746\textwidth}
    \centering
    \includegraphics[width=0.496\textwidth]{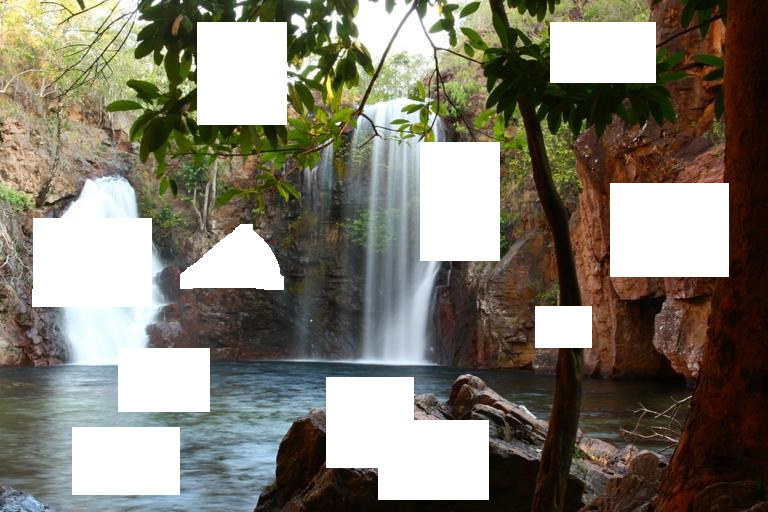}
    \includegraphics[width=0.496\textwidth]{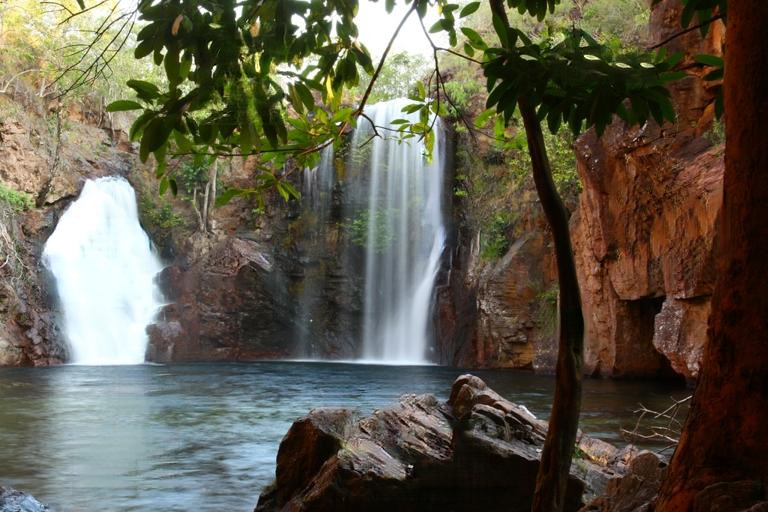}
\end{minipage}
\begin{minipage}{.2493\textwidth}
    \centering
    \includegraphics[width=0.49\textwidth]{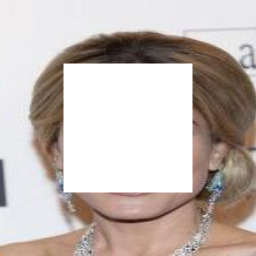}
    \includegraphics[width=0.49\textwidth]{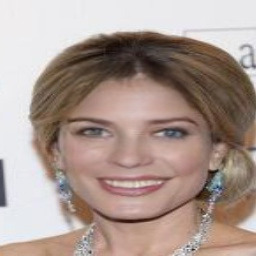}

    \includegraphics[width=0.49\textwidth]{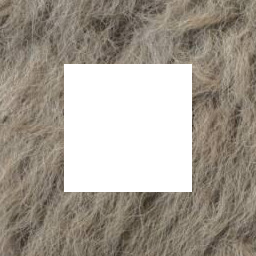}
    \includegraphics[width=0.49\textwidth]{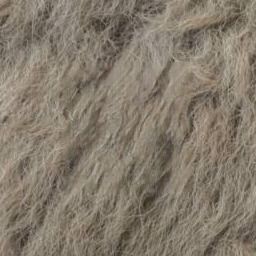}
\end{minipage}
\captionof{figure}{Example inpainting results of our method on images of natural scene, face and texture. Missing regions are shown in white. In each pair, the left is input image and right is the direct output of our trained generative neural networks without any post-processing.}
\label{fig:intro}
\end{center}

}]

\begin{abstract}
\vspace{-8mm}
Recent deep learning based approaches have shown promising results for the challenging task of inpainting large missing regions in an image. These methods can generate visually plausible image structures and textures, but often create distorted structures or blurry textures inconsistent with surrounding areas. This is mainly due to ineffectiveness of convolutional neural networks in explicitly borrowing or copying information from distant spatial locations. On the other hand, traditional texture and patch synthesis approaches are particularly suitable when it needs to borrow textures from the surrounding regions. Motivated by these observations, we propose a new deep generative model-based approach which can not only synthesize novel image structures but also explicitly utilize surrounding image features as references during network training to make better predictions. The model is a feed-forward, fully convolutional neural network which can process images with multiple holes at arbitrary locations and with variable sizes during the test time. Experiments on multiple datasets including faces (CelebA, CelebA-HQ), textures (DTD) and natural images (ImageNet, Places2) demonstrate that our proposed approach generates higher-quality inpainting results than existing ones. Code, demo and models are available at: \url{https://github.com/JiahuiYu/generative_inpainting}.

\end{abstract}

\section{Introduction}
Filling missing pixels of an image, often referred as image inpainting or completion, is an important task in computer vision. It has many applications in photo editing, image-based rendering and computational photography~\cite{barnes2009patchmatch, levin2004seamless, newson2014video, park2017transformation, simakov2008summarizing, yeh2016semantic}. The core challenge of image inpainting lies in synthesizing visually realistic and semantically plausible pixels for the missing regions that are coherent with existing ones. 

Early works~\cite{barnes2009patchmatch, hays2007scene} attempted to solve the problem using ideas similar to texture synthesis~\cite{efros2001image, efros1999texture}, i.e.\ by matching and copying background patches into holes starting from low-resolution to high-resolution or propagating from hole boundaries. These approaches work well especially in background inpainting tasks, and are widely deployed in practical applications~\cite{barnes2009patchmatch}. However, as they assume missing patches can be found somewhere in background regions, they cannot hallucinate novel image contents for challenging cases where inpainting regions involve complex, non-repetitive structures (e.g.\ faces, objects). Moreover, these methods are not able to capture high-level semantics.

Rapid progress in deep convolutional neural networks (CNN) and generative adversarial networks (GAN)~\cite{goodfellow2014generative} inspired recent works~\cite{iizuka2017globally, li2017generative, pathak2016context, yeh2016semantic} to formulate inpainting as a conditional image generation problem where high-level recognition and low-level pixel synthesis are formulated into a convolutional encoder-decoder network, jointly trained with adversarial networks to encourage the coherency between generated and existing pixels. These works are shown to generate plausible new contents in highly structured images, such as faces, objects and scenes.

Unfortunately, these CNN-based methods often create boundary artifacts, distorted structures and blurry textures inconsistent with surrounding areas. We found that this is likely due to ineffectiveness of convolutional neural networks in modeling long-term correlations between distant contextual information and the hole regions. For example, to allow a pixel being influenced by the content of 64 pixels away, it requires at least 6 layers of \(3 \times 3\) convolutions with dilation factor 2 or equivalent~\cite{iizuka2017globally, yu2015multi}. Nevertheless, a dilated convolution samples features from a regular and symmetric grid and thus may not be able to weigh the features of interest over the others. Note that a recent work~\cite{yang2016high} attempts to address the appearance discrepancy by optimizing texture similarities between generated patches and the matched patches in known regions. Although improving the visual quality, this method is being dragged by hundreds of gradient descent iterations and costs minutes to process an image with resolution \(512 \times 512\) on GPUs.

We present a unified feed-forward generative network with a novel contextual attention layer for image inpainting. Our proposed network consists of two stages. The first stage is a simple dilated convolutional network trained with reconstruction loss to rough out the missing contents. The contextual attention is integrated in the second stage. The core idea of contextual attention is to use the features of known patches as convolutional filters to process the generated patches. It is designed and implemented with convolution for matching generated patches with known contextual patches, channel-wise softmax to weigh relevant patches and deconvolution to reconstruct the generated patches with contextual patches. The contextual attention module also has spatial propagation layer to encourage spatial coherency of attention. In order to allow the network to hallucinate novel contents, we have another convolutional pathway in parallel with the contextual attention pathway. The two pathways are aggregated and fed into single decoder to obtain the final output. The whole network is trained end to end with reconstruction losses and two Wasserstein GAN losses~\cite{arjovsky2017wasserstein, gulrajani2017improved}, where one critic looks at the global image while the other looks at the local patch of the missing region.

Experiments on multiple datasets including faces, textures and natural images demonstrate that the proposed approach generates higher-quality inpainting results than existing ones. Example results are shown in Figure~\ref{fig:intro}.

Our contributions are summarized as follows:
\begin{itemize}
\setlength\itemsep{.2em}
    \item We propose a novel contextual attention layer to explicitly attend on related feature patches at distant spatial locations.
    \item We introduce several techniques including inpainting network enhancements, global and local WGANs~\cite{gulrajani2017improved} and spatially discounted reconstruction loss to improve the training stability and speed based on the current the state-of-the-art generative image inpainting network~\cite{iizuka2017globally}. As a result, we are able to train the network in a week instead of two months.
    \item Our unified feed-forward generative network achieves high-quality inpainting results on a variety of challenging datasets including CelebA faces~\cite{liu2015faceattributes}, CelebA-HQ faces~\cite{karras2017progressive}, DTD textures~\cite{cimpoi2014describing}, ImageNet~\cite{russakovsky2015imagenet} and Places2~\cite{zhou2017places}.
\end{itemize}

\section{Related Work}
\subsection{Image Inpainting}
Existing works for image inpainting can be mainly divided into two groups. The first group represents traditional diffusion-based or patch-based methods with low-level features. The second group attempts to solve the inpainting problem by a learning-based approach, e.g.\ training deep convolutional neural networks to predict pixels for the missing regions.

Traditional diffusion or patch-based approaches such as~\cite{ballester2001filling, bertalmio2000image, efros2001image, efros1999texture} typically use variational algorithms or patch similarity to propagate information from the background regions to the holes. These methods work well for stationary textures but are limited for non-stationary data such as natural images. Simakov et al.\ ~\cite{simakov2008summarizing} propose a bidirectional patch similarity-based scheme to better model non-stationary visual data for re-targeting and inpainting applications. However, dense computation of patch similarity~\cite{simakov2008summarizing} is a very expensive operation, which prohibits practical applications of such method. In order to address the challenge, a fast nearest neighbor field algorithm called PatchMatch~\cite{barnes2009patchmatch} has been proposed which has shown significant practical values for image editing applications including inpainting.



Recently, deep learning and GAN-based approaches have emerged as a promising paradigm for image inpainting. Initial efforts~\cite{kohler2014mask, xu2014deep} train convolutional neural networks for denoising and inpainting of small regions. Context Encoders~\cite{pathak2016context} firstly train deep neural networks for inpainting large holes. It is trained to complete center region of \(64 \times 64\) in a \(128 \times 128\) image, with both \(\ell_2\) pixel-wise reconstruction loss and generative adversarial loss as the objective function. More recently, Iizuka et al.\ \cite{iizuka2017globally} improve it by introducing both global and local discriminators as adversarial losses. The global discriminator assesses if completed image is coherent as a whole, while the local discriminator focus on a small area centered at the generated region to enforce the local consistency. In addition, Iizuka et al.\ \cite{iizuka2017globally} use dilated convolutions in inpainting network to replace channel-wise fully connected layer adopted in Context Encoders, both techinics are proposed for increasing receptive fields of output neurons. Meanwhile, there have been several studies focusing on generative face inpainting. Yeh et al.\ \cite{yeh2016semantic} search for the closest encoding in latent space of the corrupted image and decode to get completed image. Li et al.\ \cite{li2017generative} introduce additional face parsing loss for face completion. However, these methods typically require post processing steps such as image blending operation to enforce color coherency near the hole boundaries.

Several works~\cite{snelgrove2017high, yang2016high} follow ideas from image stylization~\cite{chen2016fast, li2016combining} to formulate the inpainting as an optimization problem. For example, Yang et al.\ \cite{yang2016high} propose a multi-scale neural patch synthesis approach based on joint optimization of image content and texture constraints, which not only preserves contextual structures but also produces high-frequency details by matching and adapting patches with the most similar mid-layer feature correlations of a deep classification network. This approach shows promising visual results but is very slow due to the optimization process.

\subsection{Attention Modeling}
There have been many studies on learning spatial attention in deep convolutional neural networks. Here, we select to review a few representative ones related to the proposed contextual attention model. Jaderberg et al.~\cite{jaderberg2015spatial} firstly propose a parametric spatial attention module called spatial transformer network (STN) for object classification tasks. The model has a localization module to predict parameters of global affine transformation to warp features. However, this model assumes a global transformation so is not suitable for modeling patch-wise attention. Zhou et al.~\cite{zhou2016view} introduce an appearance flow to predict offset vectors specifying which pixels in the input view should be moved to reconstruct the target view for novel view synthesis. This method is shown to be effective for matching related views of the same objects but is not effective in predicting a flow field from the background region to the hole, according to our experiments. Recently, Dai et al.\ \cite{dai2017deformable} and Jeon et al.\ \cite{jeon2017active} propose to learn spatially attentive or active convolutional kernels. These methods can potentially better leverage information to deform the convolutional kernel shape during training but may still be limited when we need to borrow exact features from the background.

\section{Improved Generative Inpainting Network}
\begin{figure*}[h]
\begin{center}
\includegraphics[width=1.\linewidth]{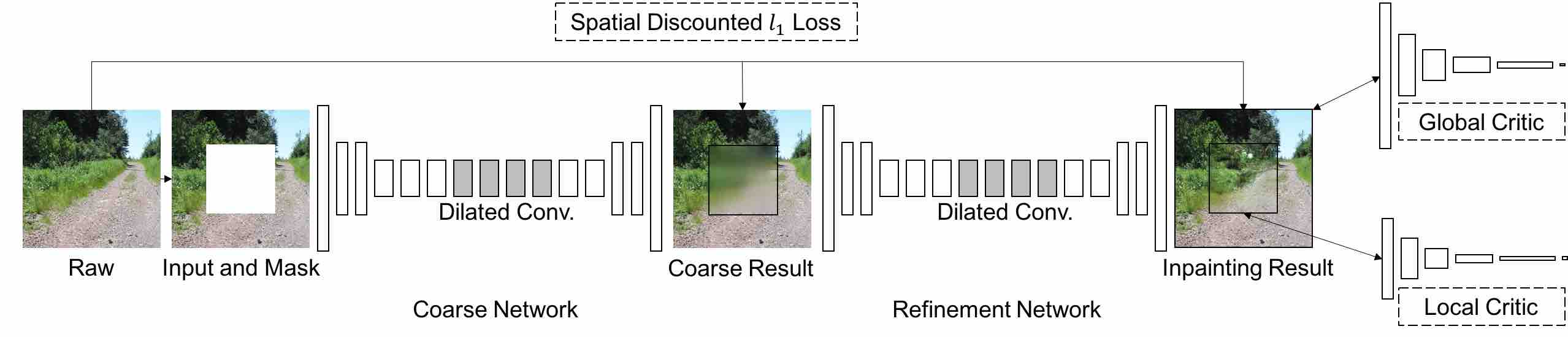}
\end{center}
   \caption{Overview of our improved generative inpainting framework. The coarse network is trained with reconstruction loss explicitly, while the refinement network is trained with reconstruction loss, global and local WGAN-GP adversarial loss.}
\label{fig:framework}
\end{figure*}
We first construct our baseline generative image inpainting network by reproducing and making several improvements to the recent state-of-the-art inpainting model~\cite{iizuka2017globally} which has shown promising visual results for inpainting images of faces, building facades and natural images. 

\textbf{Coarse-to-fine network architecture} 
The network architecture of our improved model is shown in Figure~\ref{fig:framework}. We follow the same input and output configurations as in~\cite{iizuka2017globally} for training and inference, i.e. the generator network takes an image with white pixels filled in the holes and a binary mask indicating the hole regions as input pairs, and outputs the final completed image. We pair the input with a corresponding binary mask to handle holes with variable sizes, shapes and locations. The input to the network is a \(256 \times 256\) image with a rectangle missing region sampled randomly during training, and the trained model can take an image of different sizes with multiple holes in it.

In image inpainting tasks, the size of the receptive fields should be sufficiently large, and Iizuka et al.\ \cite{iizuka2017globally} adopt dilated convolution for that purpose. To further enlarge the receptive fields and stabilize training, we introduce a two-stage coarse-to-fine network architecture where the first network makes an initial coarse prediction, and the second network takes the coarse prediction as inputs and predict refined results. The coarse network is trained with the reconstruction loss explicitly, while the refinement network is trained with the reconstruction as well as GAN losses. Intuitively, the refinement network sees a more complete scene than the original image with missing regions, so its encoder can learn better feature representation than the coarse network. This two-stage network architecture is similar in spirits to residual learning~\cite{he2016deep} or deep supervision~\cite{lee2015deeply}.

Also, our inpainting network is designed in a thin and deep scheme for efficiency purpose and has fewer parameters than the one in ~\cite{iizuka2017globally}. In terms of layer implementations, we use mirror padding for all convolution layers and remove batch normalization layers~\cite{ioffe2015batch} (which we found deteriorates color coherence). Also, we use ELUs~\cite{clevert2015fast} as activation functions instead of ReLU in~\cite{iizuka2017globally}, and clip the output filter values instead of using \(tanh\) or \(sigmoid\) functions. In addition, we found separating global and local feature representations for GAN training works better than feature concatenation in~\cite{iizuka2017globally}. More details can be found in the supplementary materials.

\textbf{Global and local Wasserstein GANs} \label{subsec:wgan}
Different from previous generative inpainting networks~\cite{iizuka2017globally, li2017generative, pathak2016context} which rely on DCGAN~\cite{radford2015unsupervised} for adversarial supervision, we propose to use a modified version of WGAN-GP~\cite{arjovsky2017wasserstein, gulrajani2017improved}. We attach the WGAN-GP loss to both global and local outputs of the second-stage refinement network to enforce global and local consistency, inspired by~\cite{iizuka2017globally}. WGAN-GP loss is well-known to outperform existing GAN losses for image generation tasks, and it works well when combined with \(\ell_1\) reconstruction loss as they both use the \(\ell_1\) distance metric.

Specifically, WGAN uses the \textit{Earth-Mover} distance (a.k.a.\ \textit{Wasserstein-1}) distance \(W(\mathbb{P}_r, \mathbb{P}_g)\) for comparing the generated and real data distributions. Its objective function is constructed by applying the \textit{Kantorovich-Rubinstein} duality:
\[ \min_G\max_{D \in \mathcal{D}} E_{\mathbf{x} \sim \mathbb{P}_r}[D(\mathbf{x})] - E_{\tilde{\mathbf{x}} \sim \mathbb{P}_g}[D(\tilde{\mathbf{x}})],\]
where \(\mathcal{D}\) is the set of 1-Lipschitz functions and \(\mathbb{P}_g\) is the model distribution implicitly defined by \(\tilde{\mathbf{x}} = G(\mathbf{z})\). \(\mathbf{z}\) is the input to the generator.

Gulrajani et al.~\cite{gulrajani2017improved} proposed an improved version of WGAN with a gradient penalty term
\[\lambda E_{\hat{\mathbf{x}} \sim \mathbb{P}_{\hat{\mathbf{x}}}}(\lVert \nabla_{\hat{\mathbf{x}}} D(\hat{\mathbf{x}})\rVert_2 - 1)^2,\]
where \(\mathbf{\hat x}\) is sampled from the straight line between points sampled from distribution \(\mathbb{P}_g\) and \(\mathbb{P}_r\). The reason is that the gradient of \(D^*\) at all points \(\hat{\mathbf{x}} = (1 - t)\mathbf{x} + t\tilde{\mathbf{x}}\) on the straight line should point directly towards current sample \(\tilde{\mathbf{x}}\), meaning \(\nabla_{\mathbf{\hat x}} D^*(\hat{\mathbf{x}}) = \frac{\tilde{\mathbf{x}}-\hat{\mathbf{x}}}{\lVert \tilde{\mathbf{x}}-\hat{\mathbf{x}}\rVert}\). 

For image inpainting, we only try to predict hole regions, thus the gradient penalty should be applied only to pixels inside the holes. This can be implemented with multiplication of gradients and input mask \(\mathbf{m}\) as follows:
\[\lambda E_{\hat{\mathbf{x}} \sim \mathbb{P}_{\hat{\mathbf{x}}}}(\lVert \nabla_{\hat{\mathbf{x}}} D(\hat{\mathbf{x}}) \odot (\mathbf{1} - \mathbf{m})\rVert_2 - 1)^2,\]
where the mask value is \(0\) for missing pixels and \(1\) for elsewhere. \(\lambda\) is set to 10 in all experiments.

We use a weighted sum of pixel-wise \(\ell_1\) loss (instead of mean-square-error as in~\cite{iizuka2017globally}) and WGAN adversarial losses. Note that in primal space, \textit{Wasserstein-1} distance in WGAN is based on \(\ell_1\) ground distance:
\[ W(\mathbb{P}_r, \mathbb{P}_g) = \inf_{\gamma \in \prod(\mathbb{P}_r, \mathbb{P}_g)} E_{(\mathbf{x},\mathbf{y}) \sim \gamma}[\lVert \mathbf{x} - \mathbf{y} \rVert],\]
where \(\prod(\mathbb{P}_r, \mathbb{P}_g)\) denotes the set of all joint distributions \(\gamma(\mathbf{x}, \mathbf{y})\) whose marginals are respectively \(\mathbb{P}_r\) and \(\mathbb{P}_g\). Intuitively, the pixel-wise reconstruction loss directly regresses holes to the current ground truth image, while WGANs implicitly learn to match potentially correct images and train the generator with adversarial gradients. As both losses measure pixel-wise \(\ell_1\) distances, the combined loss is easier to train and makes the optimization process stabler.

\textbf{Spatially discounted reconstruction loss}
Inpainting problems involve hallucination of pixels, so it could have many plausible solutions for any given context. In challenging cases, a plausible completed image can have patches or pixels that are very different from those in the original image. As we use the original image as the only ground truth to compute a reconstruction loss, strong enforcement of reconstruction loss in those pixels may mislead the training process of convolutional network. 

Intuitively, missing pixels near the hole boundaries have much less ambiguity than those pixels closer to the center of the hole. This is similar to the issue observed in reinforcement learning. When long-term rewards have large variations during sampling, people use temporal discounted rewards over sampled trajectories~\cite{sutton1998reinforcement}. Inspired by this, we introduce spatially discounted reconstruction loss using a weight mask \(\mathbf{M}\). The weight of each pixel in the mask is computed as \(\gamma^{l}\), where \(l\) is the distance of the pixel to the nearest known pixel. \(\gamma\) is set to 0.99 in all experiments.

Similar weighting ideas are also explored in~\cite{pathak2016context, yeh2016semantic}. Importance weighted context loss, proposed in~\cite{yeh2016semantic}, is spatially weighted by the ratio of uncorrupted pixels within a fixed window (e.g. \(7 \times 7\)). Pathak et al.\ \cite{pathak2016context} predict a slightly larger patch with higher loss weighting (\(\times 10\)) in the border area. For inpainting large hole, the proposed discounted loss is more effective for improving the visual quality. We use discounted \(\ell_1\) reconstruction loss in our implementation.

With all the above improvements, our baseline generative inpainting model converges much faster than~\cite{iizuka2017globally} and result in more accurate inpainting results. For Places2~\cite{zhou2017places}, we reduce the training time from 11,520 GPU-hours (K80) reported by~\cite{iizuka2017globally} to 120 GPU-hours (GTX 1080) which is almost \(100 \times\) speedup. Moreover, the post-processing step (image blending)~\cite{iizuka2017globally} is no longer necessary. 

\section{Image Inpainting with Contextual Attention}
Convolutional neural networks process image features with local convolutional kernel layer by layer thus are not effective for borrowing features from distant spatial locations. To overcome the limitation, we consider attention mechanism and introduce a novel contextual attention layer in the deep generative network. In this section, we first discuss details of the contextual attention layer, and then address how we integrate it into our unified inpainting network.

\subsection{Contextual Attention} \label{subsec:attention}
\begin{figure}[t]
\centering
    \includegraphics[width=1.\linewidth]{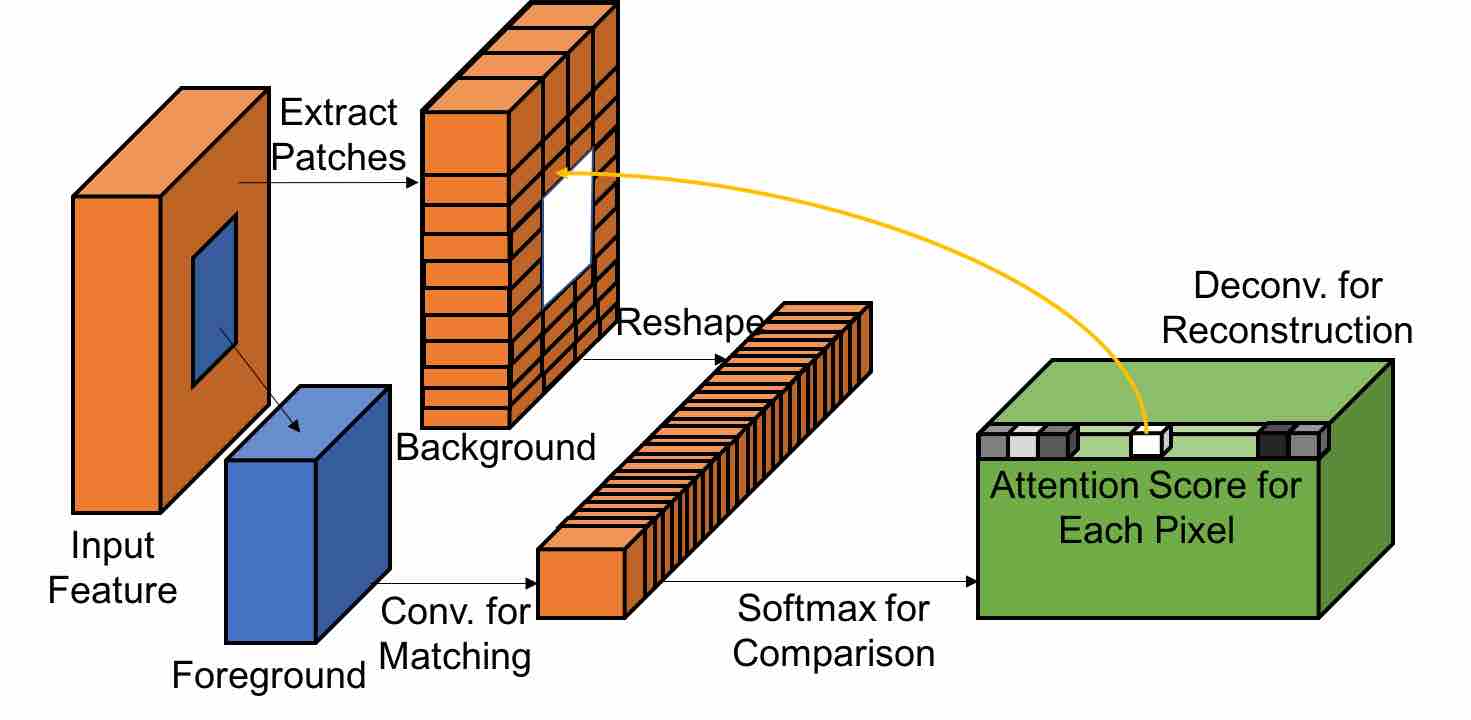}
    \caption{Illustration of the contextual attention layer. Firstly we use convolution to compute matching score of foreground patches with background patches (as convolutional filters). Then we apply softmax to compare and get attention score for each pixel. Finally we reconstruct foreground patches with background patches by performing deconvolution on attention score. The contextual attention layer is differentiable and fully-convolutional.}
    \label{fig:contextual_attention_layer}
\end{figure}
The contextual attention layer learns where to borrow or copy feature information from known background patches to generate missing patches. It is differentiable, thus can be trained in deep models, and fully-convolutional, which allows testing on arbitrary resolutions.

\textbf{Match and attend}
We consider the problem where we want to match features of missing pixels (foreground) to surroundings (background). As shown in Figure~\ref{fig:contextual_attention_layer}, we first extract patches (\(3 \times 3\)) in background and reshape them as convolutional filters. To match foreground patches \(\{f_{x, y}\}\) with backgrounds ones \(\{b_{x',y'}\}\), we measure with normalized inner product (cosine similarity)
\[s_{x,y,x',y'} = \langle \frac{f_{x,y}}{||f_{x,y}||}, \frac{b_{x',y'}}{||b_{x',y'}||}\rangle,\]
where \(s_{x,y,x',y'}\) represents similarity of patch centered in background \((x',y')\) and foreground \((x, y)\). Then we weigh the similarity with scaled softmax along \(x'y'\)-dimension to get attention score for each pixel \(s^*_{x,y,x',y'} = \textit{softmax}_{x',y'} (\lambda s_{x,y,x',y'})\), where \(\lambda\) is a constant value. This is efficiently implemented as convolution and channel-wise softmax. Finally, we reuse extracted patches \(\{b_{x',y'}\}\) as deconvolutional filters to reconstruct foregrounds. Values of overlapped pixels are averaged.

\textbf{Attention propagation}
We further encourage coherency of attention by propagation (fusion). The idea of coherency is that a shift in foreground patch is likely corresponding to an equal shift in background patch for attention. For example, \(s^*_{x, y, x', y'}\) usually have close value with \(s^*_{x+1, y, x'+1, y'}\). To model and encourage coherency of attention maps, we do a left-right propagation followed by a top-down propagation with kernel size of \(k\). Take left-right propagation as an example, we get new attention score with:
\[\hat{s}_{x, y, x', y'} = \sum_{i \in \{-k, ..., k\}}s^*_{x+i, y, x'+i, y'}.\]
The propagation is efficiently implemented as convolution with identity matrix as kernels. Attention propagation significantly improves inpainting results in testing and enriches gradients in training.

\textbf{Memory efficiency}
Assuming that a \(64 \times 64\) region is missing in a \(128 \times 128\) feature map, then the number of convolutional filters extracted from backgrounds is 12,288. This may cause memory overhead for GPUs. To overcome this issue, we introduce two options: 1) extracting background patches with strides to reduce the number of filters and 2) downscaling resolution of foreground inputs before convolution and upscaling attention map after propagation.

\subsection{Unified Inpainting Network} \label{subsec:unified}
\begin{figure}[t]
\centering
    \includegraphics[width=1.\linewidth]{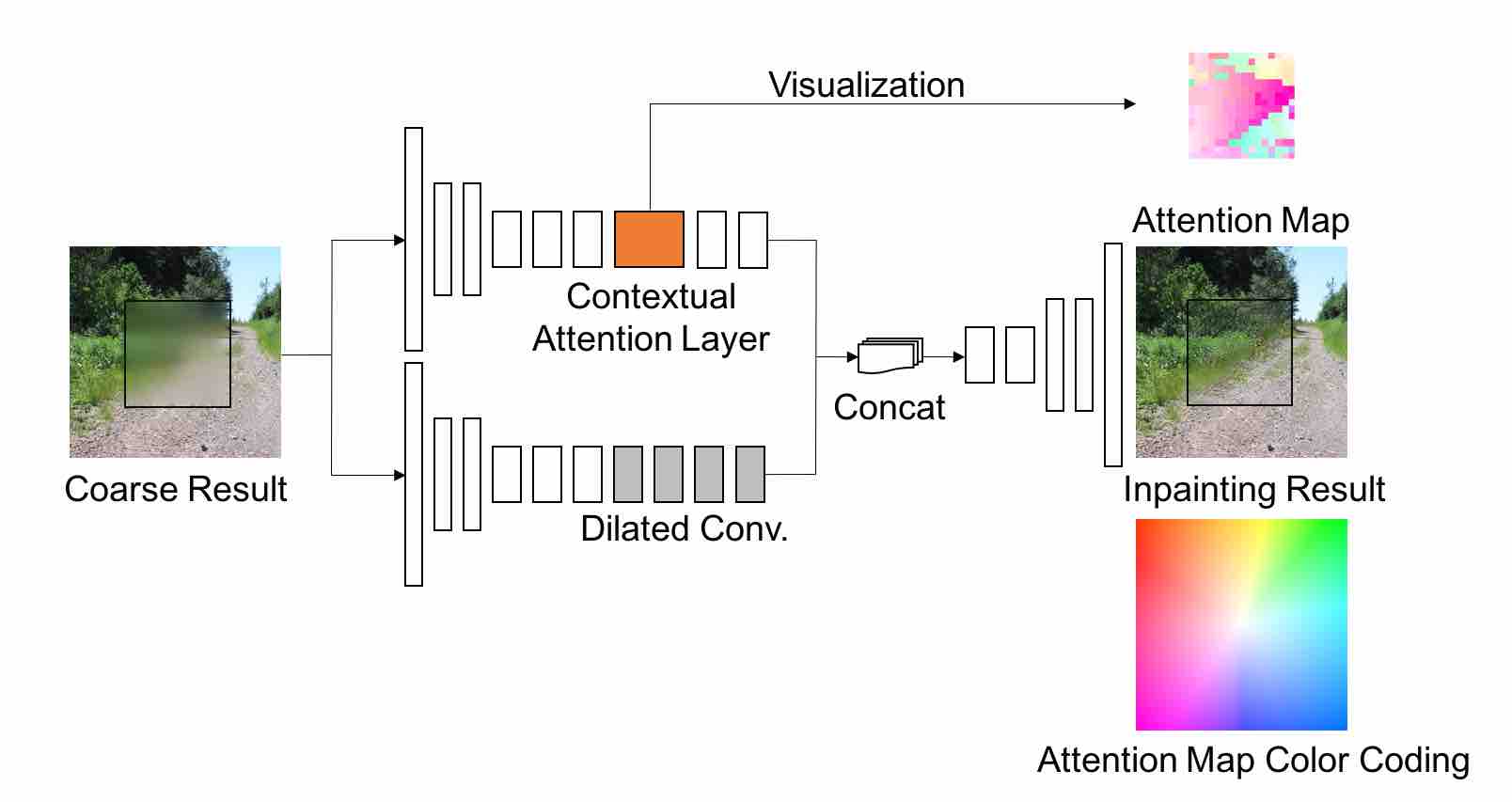}
    \caption{Based on coarse result from the first encoder-decoder network, two parallel encoders are introduced and then merged to single decoder to get inpainting result. For visualization of  attention map, color indicates relative location of the most interested background patch for each pixel in foreground. For examples, white (center of color coding map) means the pixel attends on itself, pink on bottom-left, green means on top-right.}
    \label{fig:unified_attention}
\end{figure}
To integrate attention module, we introduce two parallel encoders as shown in Figure~\ref{fig:unified_attention} based on Figure~\ref{fig:framework}. The bottom encoder specifically focuses on hallucinating contents with layer-by-layer (dilated) convolution, while the top one tries to attend on background features of interest. Output features from two encoders are aggregated and fed into a single decoder to obtain the final output. To interpret contextual attention, we visualize it in a way shown in Figure~\ref{fig:unified_attention}. We use color to indicate the relative location of the most interested background patch for each foreground pixel. For examples, white (center of color coding map) means the pixel attends on itself, pink on bottom-left, green on top-right. The offset value is scaled differently for different images to best visualize the most interesting range.

For training, given a raw image \(\mathbf{x}\), we sample a binary image mask \(\mathbf{m}\) at a random location. Input image \(\mathbf{z}\) is corrupted from the raw image as \(\mathbf{z} = \mathbf{x} \odot \mathbf{m}\). Inpainting network \(G\) takes concatenation of \(\mathbf{z}\) and \(\mathbf{m}\) as input, and output predicted image \(\mathbf{x'} = G(\mathbf{z, m})\) with the same size as input. Pasting the masked region of \(\mathbf{x'}\) to input image, we get the inpainting output \(\tilde{\mathbf{x}} = \mathbf{z} + \mathbf{x'} \odot \mathbf{(1-m)}\). Image values of input and output are linearly scaled to \([-1, 1]\) in all experiments. Training procedure is shown in Algorithm~\ref{algo:algo}.

\begin{algorithm}[ht]
\caption{Training of our proposed framework.}
    \begin{algorithmic}[1]
        \While {G has not converged}
            \For {\(i = 1, ..., 5\)}
                \State{Sample batch images \(\mathbf{x}\) from training data;}
                \State{Generate random masks \(\mathbf{m}\) for \(\mathbf{x}\);}
                \State{Construct inputs \(\mathbf{z} \gets \mathbf{x} \odot \mathbf{m}\);}
                \State{Get predictions \(\tilde{\mathbf{x}} \gets \mathbf{z} + G(\mathbf{z}, \mathbf{m}) \odot (\mathbf{1-m})\);}
                \State{Sample \(t \sim U[0,1]\) and \(\hat{\mathbf{x}} \gets (1 - t)\mathbf{x} + t\tilde{\mathbf{x}}\);}
                \State{Update two critics with \(\mathbf{x}\), \(\tilde{\mathbf{x}}\) and \(\hat{\mathbf{x}}\);}
            \EndFor
            \State{Sample batch images \(\mathbf{x}\) from training data;}
            \State{Generate random masks \(\mathbf{m}\) for \(\mathbf{x}\);}
            \State{Update inpainting network G with spatial dis-}
            \State{counted \(\ell_1\) loss and two adversarial critic losses;}
        \EndWhile
    \end{algorithmic}
\label{algo:algo}
\end{algorithm}

\section{Experiments} \label{sec:expr}
We evaluate the proposed inpainting model on four datasets including Places2~\cite{zhou2017places}, CelebA faces~\cite{liu2015faceattributes}, CelebA-HQ faces~\cite{karras2017progressive}, DTD textures~\cite{cimpoi2014describing} and ImageNet~\cite{russakovsky2015imagenet}. 

\begin{figure*}[h]
\centering

\includegraphics[width=0.322\linewidth]{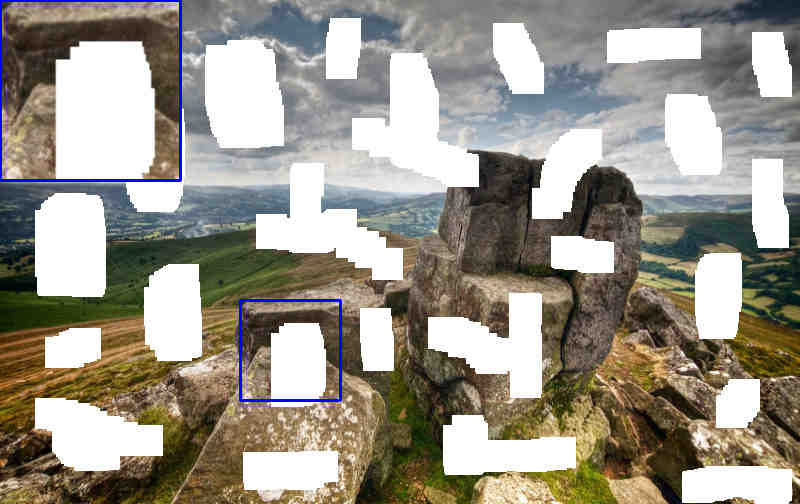}
\includegraphics[width=0.322\linewidth]{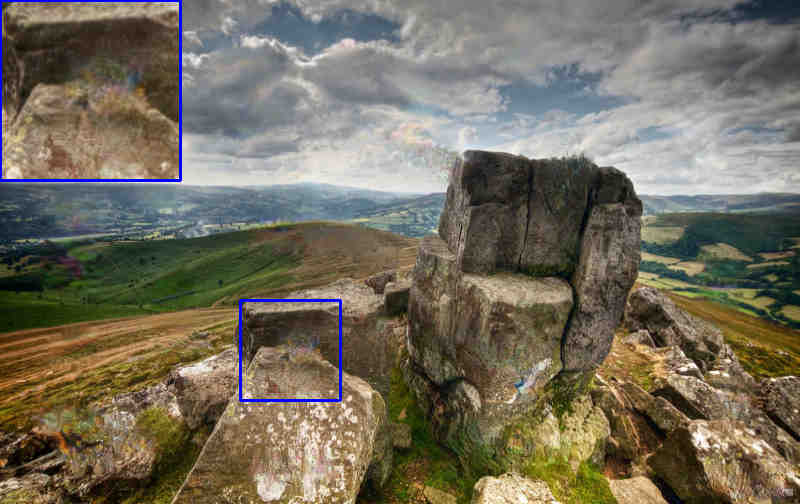}
\includegraphics[width=0.322\linewidth]{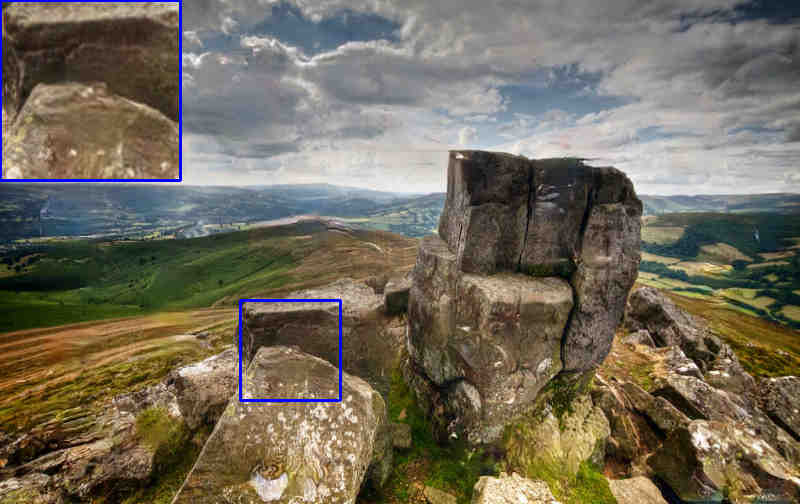}

\caption{Comparison of our baseline model with Iizuka et al.\ \cite{iizuka2017globally}. From left to right, we show the input image, result copied from main paper of work~\cite{iizuka2017globally}, and result of our baseline model. Note that no post-processing step is performed for our baseline model, while image blending is applied for the result of~\cite{iizuka2017globally}. Best viewed with zoom-in.}
\label{fig:expr_siggraph}
\end{figure*}

\textbf{Qualitative comparisons}
First, we show in Figure~\ref{fig:expr_siggraph} that our baseline model generates comparable inpainting results with the previous state-of-the-art~\cite{iizuka2017globally} by comparing our output result and result copied from their main paper. Note that no post-processing step is performed for our baseline model, while image blending is applied in result of~\cite{iizuka2017globally}.

\begin{figure*}[h]
\centering
\includegraphics[width=0.1901016\linewidth]{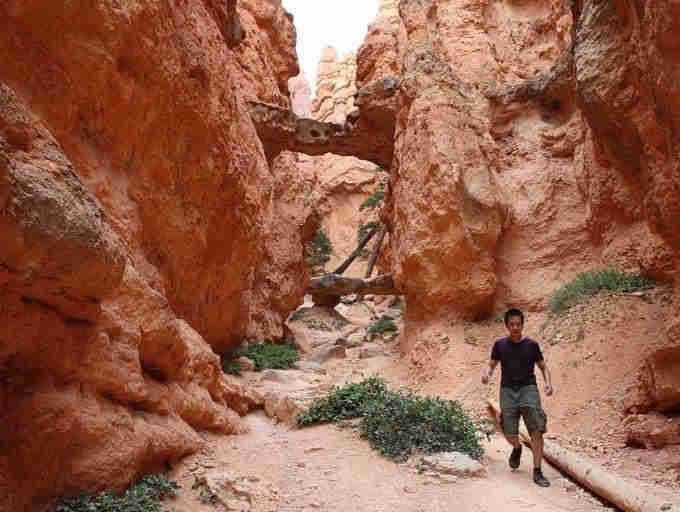}
\includegraphics[width=0.1901016\linewidth]{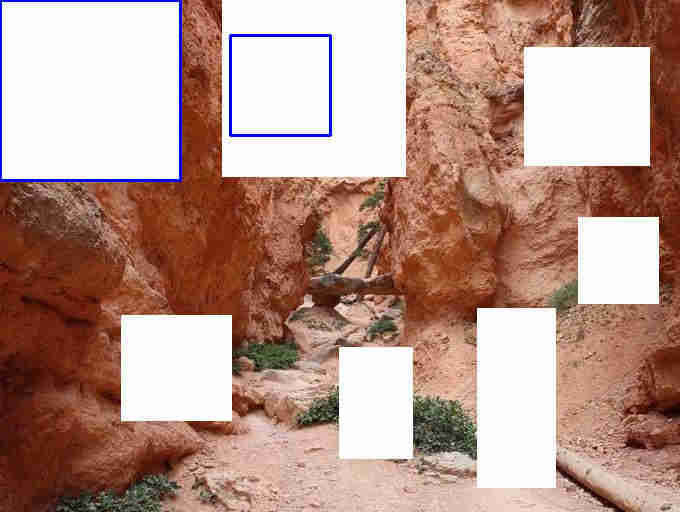}
\includegraphics[width=0.1901016\linewidth]{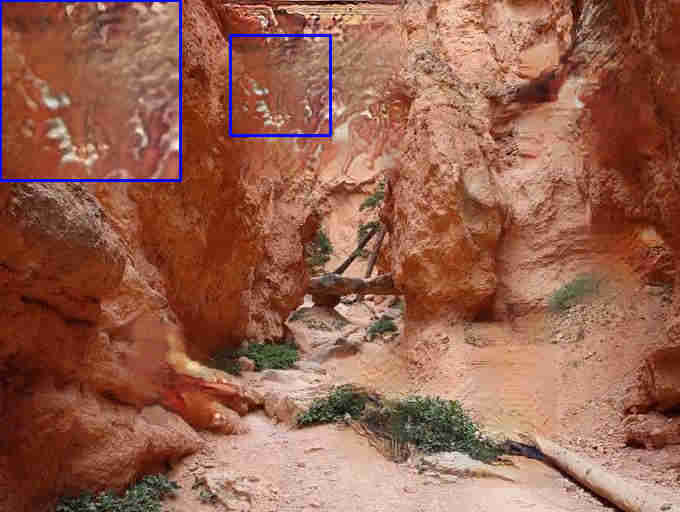}
\includegraphics[width=0.1901016\linewidth]{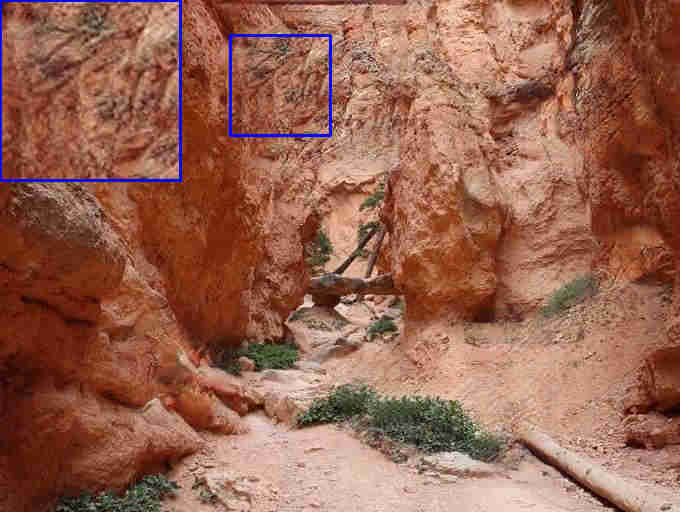}
\includegraphics[width=0.1901016\linewidth]{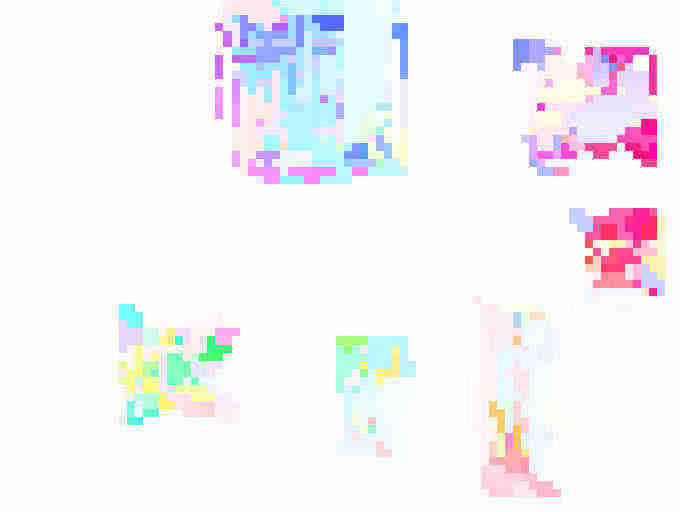}

\includegraphics[width=0.1901016\linewidth]{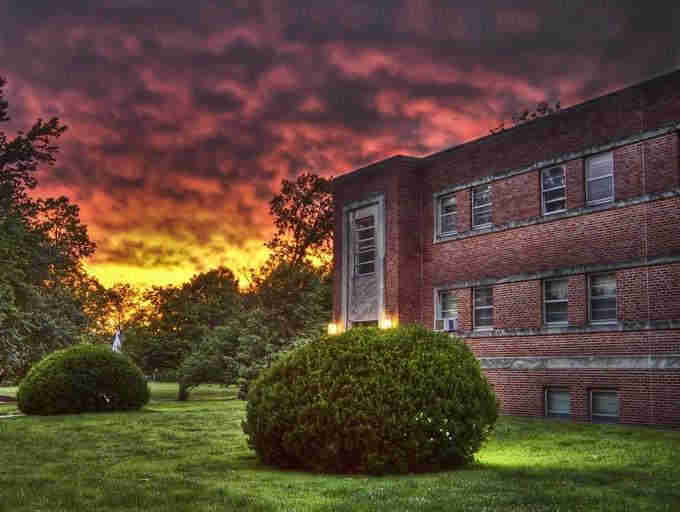}
\includegraphics[width=0.1901016\linewidth]{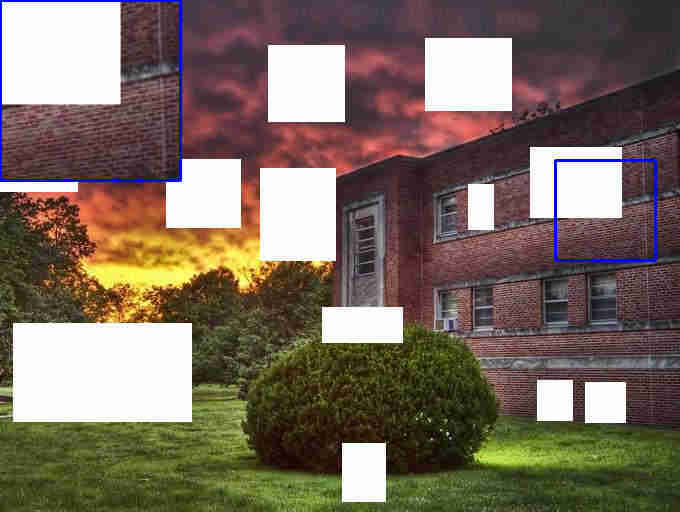}
\includegraphics[width=0.1901016\linewidth]{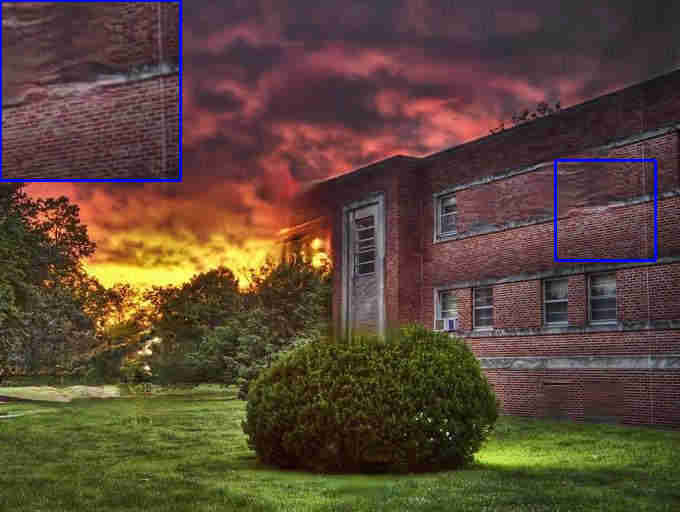}
\includegraphics[width=0.1901016\linewidth]{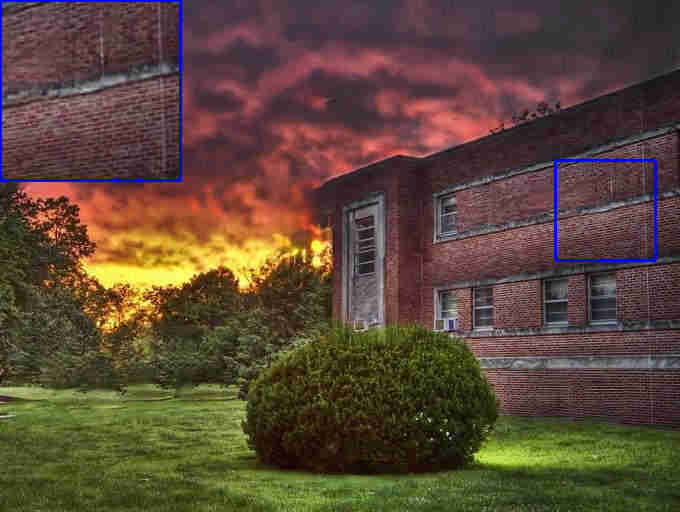}
\includegraphics[width=0.1901016\linewidth]{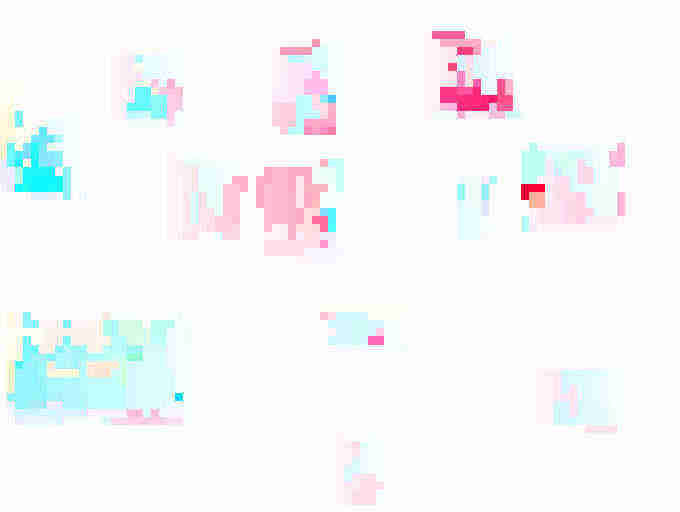}

\includegraphics[width=0.1901016\linewidth]{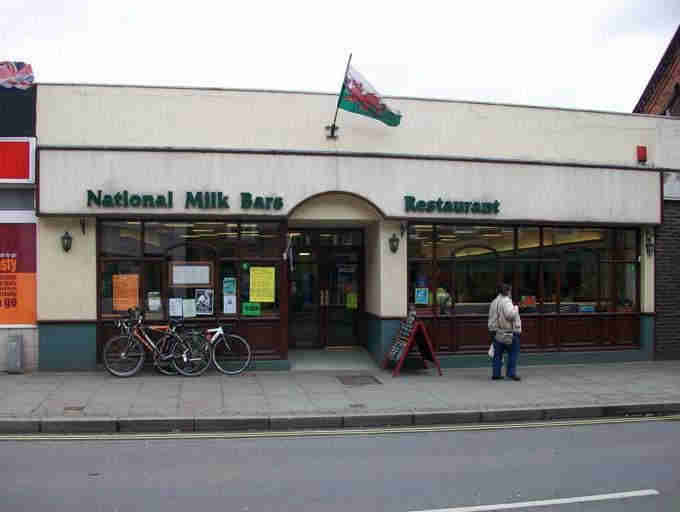}
\includegraphics[width=0.1901016\linewidth]{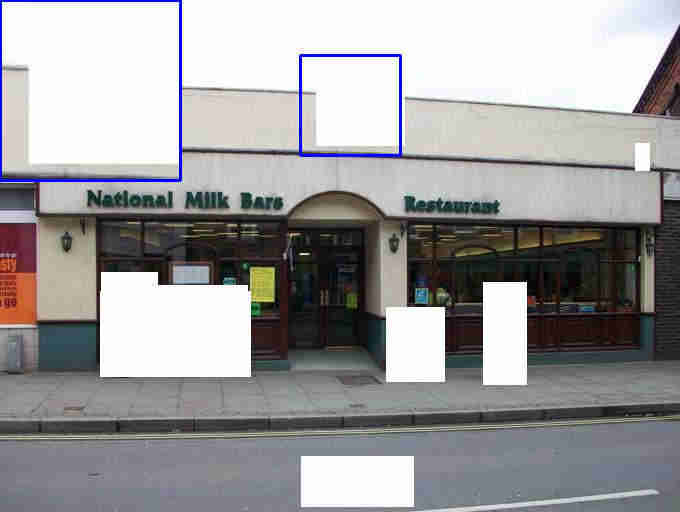}
\includegraphics[width=0.1901016\linewidth]{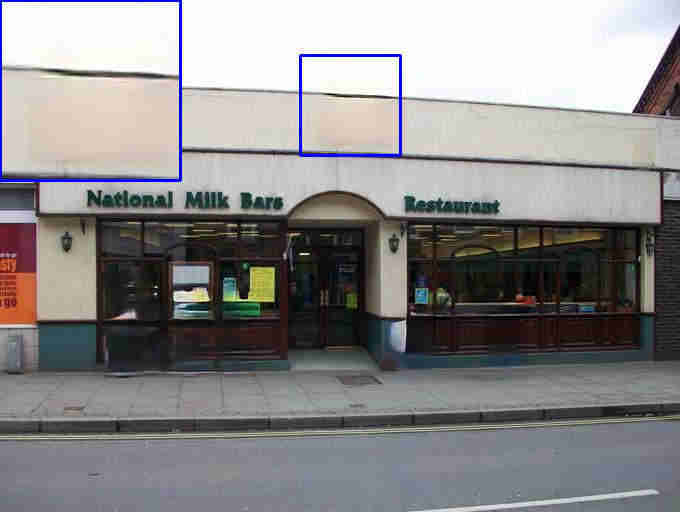}
\includegraphics[width=0.1901016\linewidth]{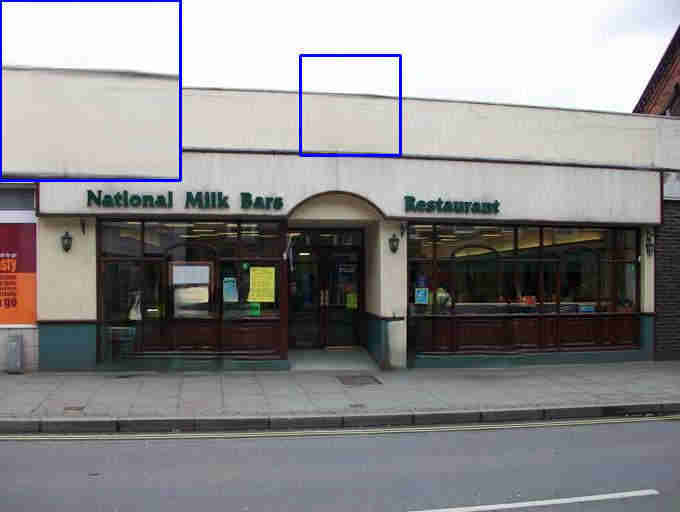}
\includegraphics[width=0.1901016\linewidth]{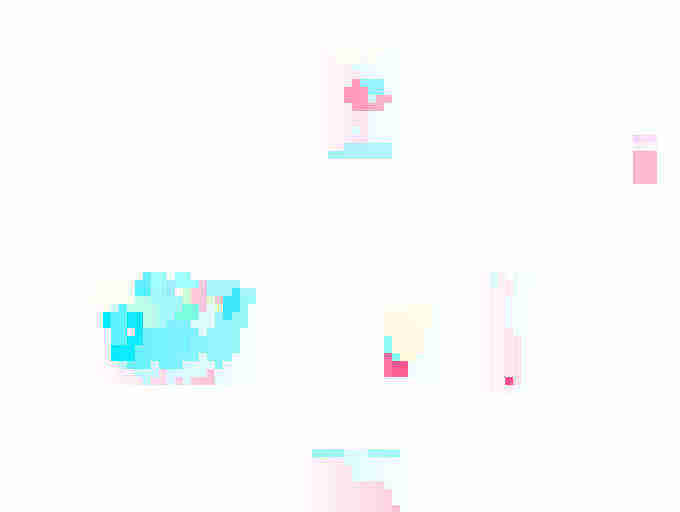}

\includegraphics[width=0.1901016\linewidth]{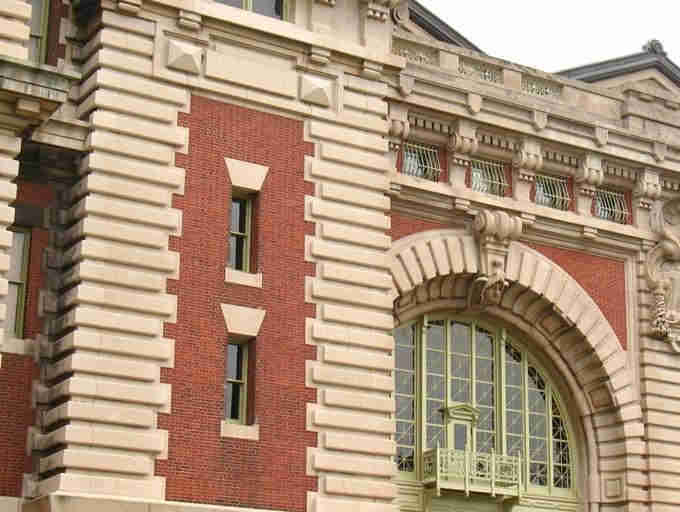}
\includegraphics[width=0.1901016\linewidth]{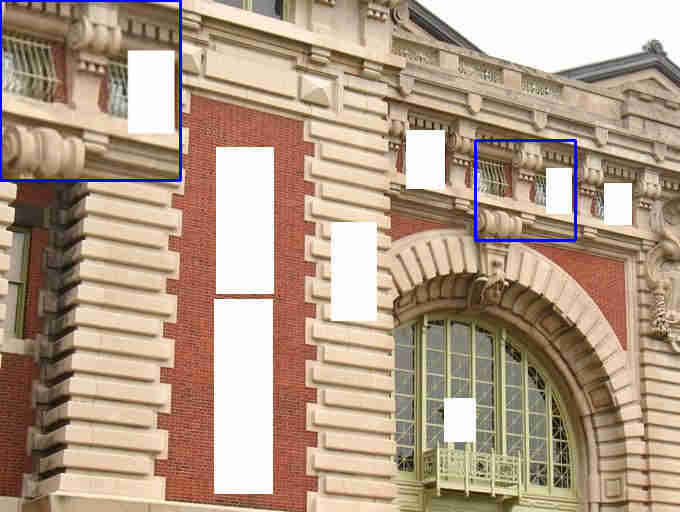}
\includegraphics[width=0.1901016\linewidth]{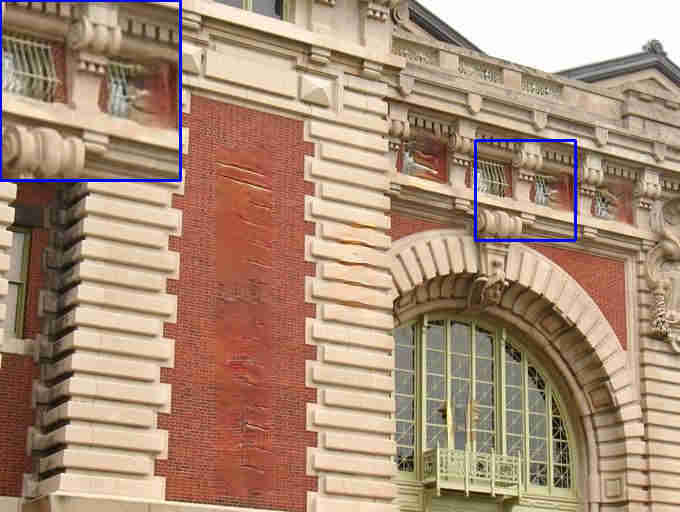}
\includegraphics[width=0.1901016\linewidth]{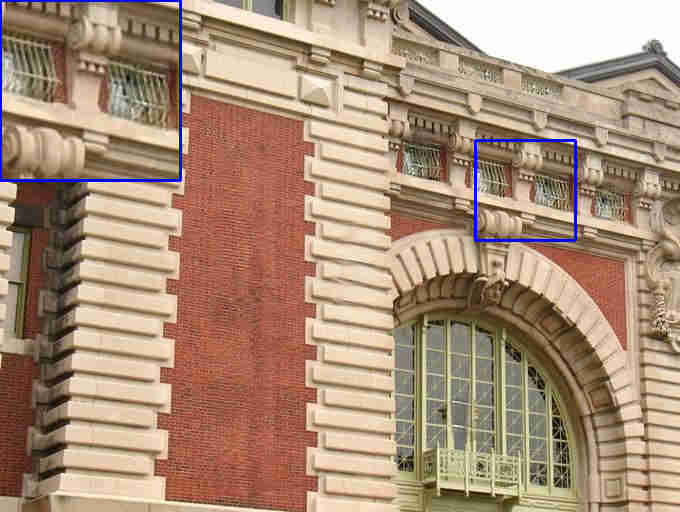}
\includegraphics[width=0.1901016\linewidth]{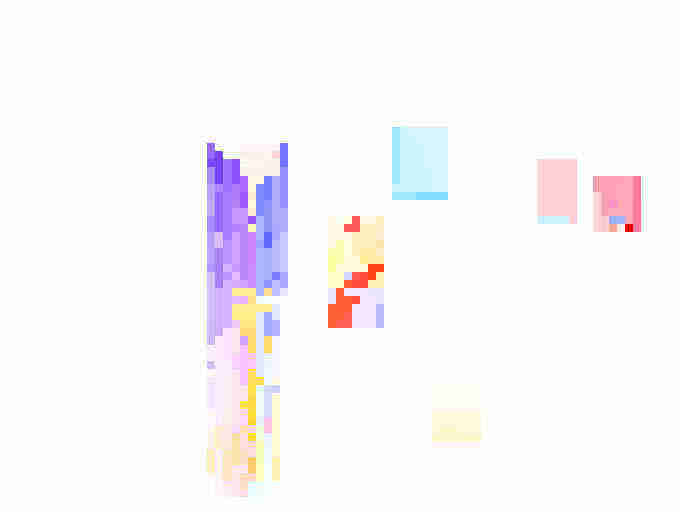}

\includegraphics[width=0.1901016\linewidth]{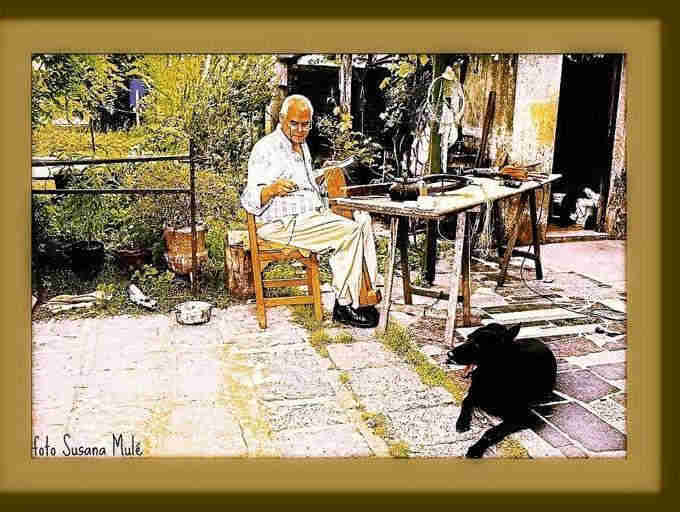}
\includegraphics[width=0.1901016\linewidth]{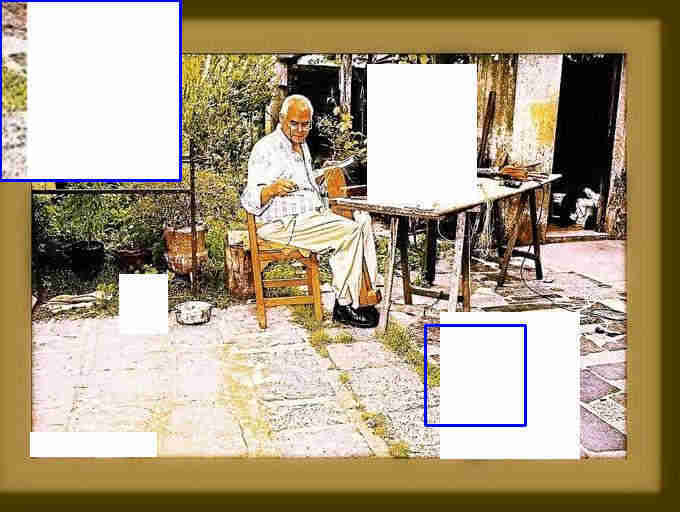}
\includegraphics[width=0.1901016\linewidth]{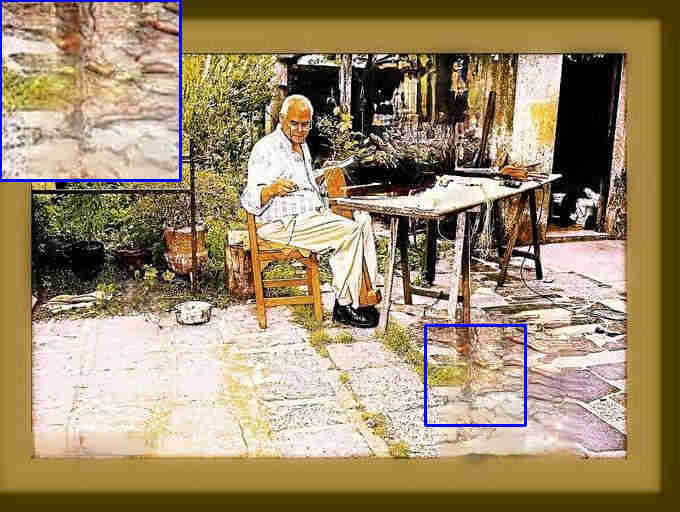}
\includegraphics[width=0.1901016\linewidth]{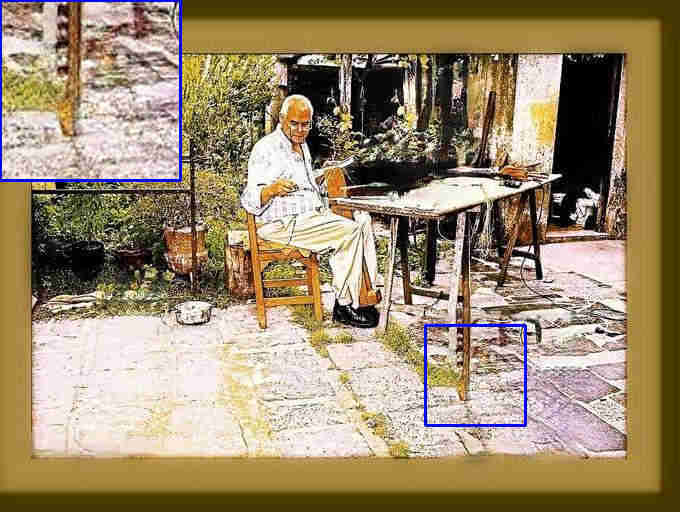}
\includegraphics[width=0.1901016\linewidth]{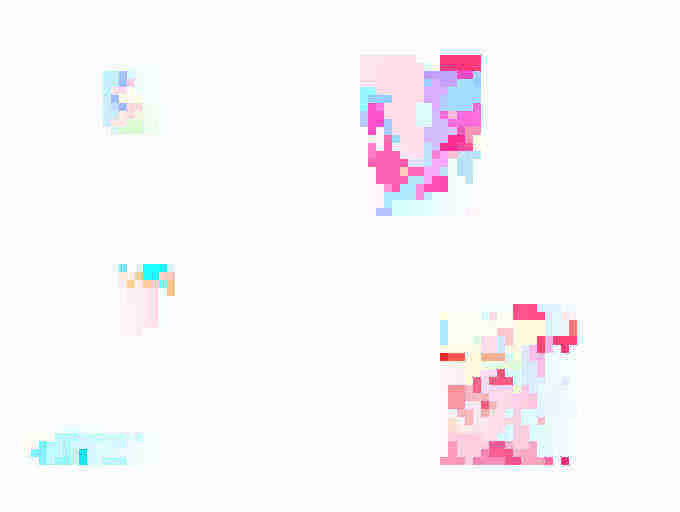}

\includegraphics[width=0.1901016\linewidth]{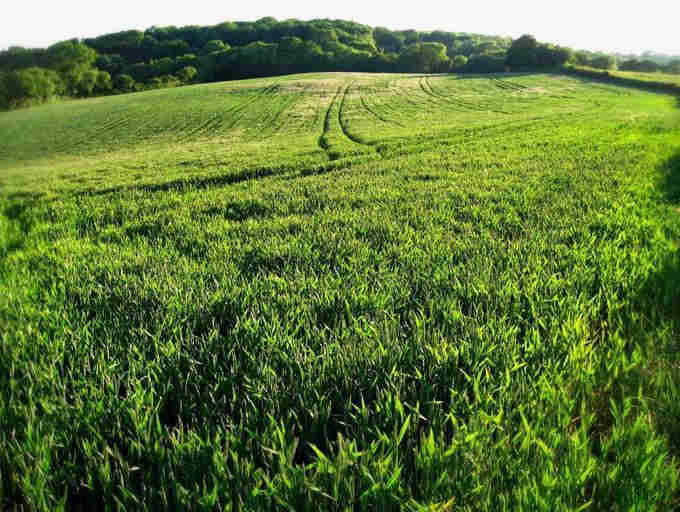}
\includegraphics[width=0.1901016\linewidth]{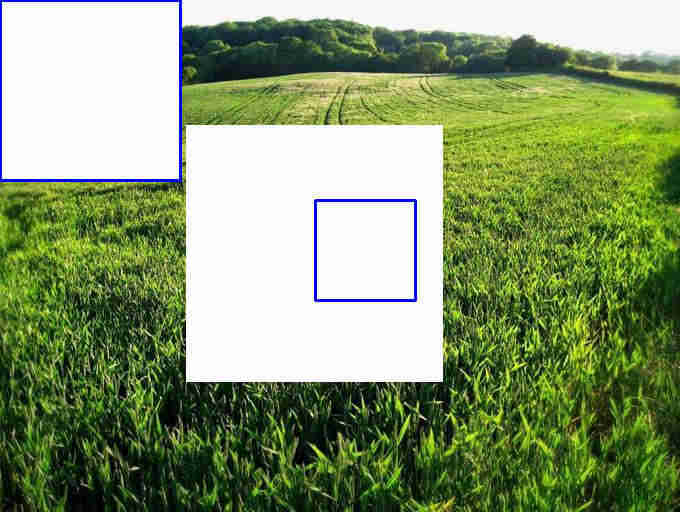}
\includegraphics[width=0.1901016\linewidth]{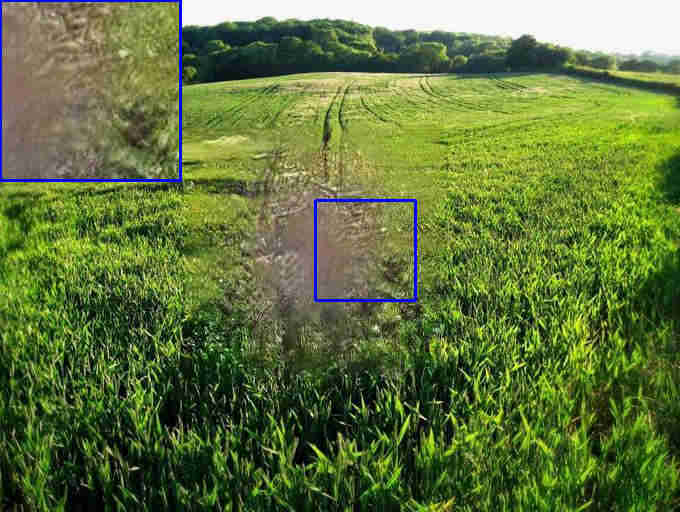}
\includegraphics[width=0.1901016\linewidth]{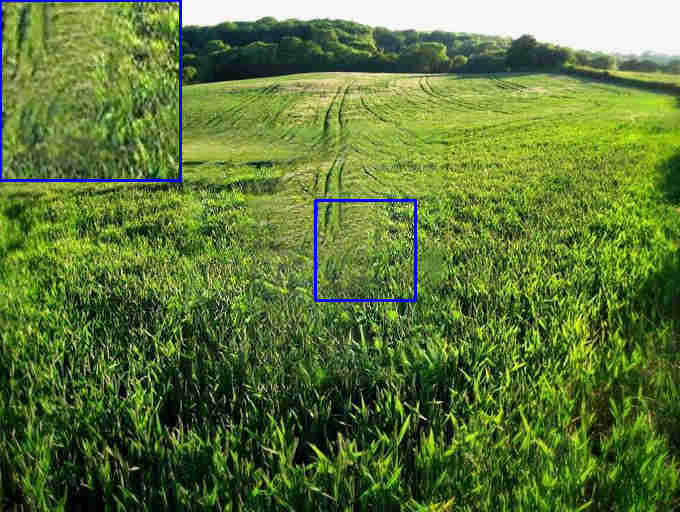}
\includegraphics[width=0.1901016\linewidth]{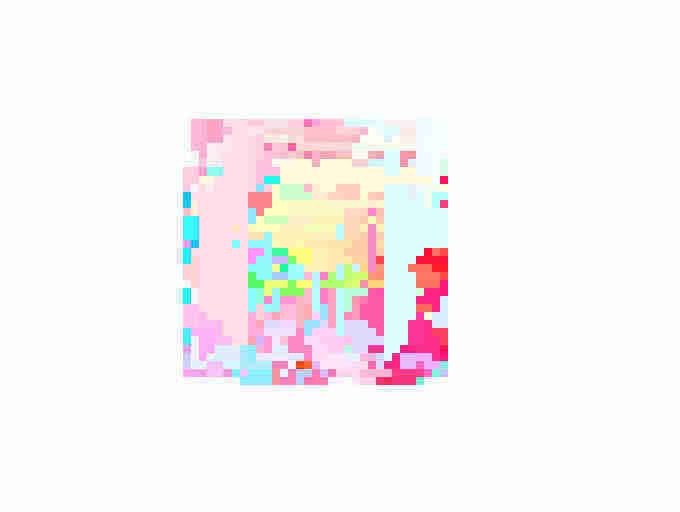}

\caption{Qualitative results and comparisons to the baseline model. We show from left to right the original image, input image, result of our baseline model, result and attention map (upscaled \(4 \times\)) of our full model. Best viewed with zoom-in.}
\label{fig:expr_main_results}
\end{figure*}

Next we use the most challenging Places2 dataset to evaluate our full model with contextual attention by comparing to our baseline two-stage model which is extended from the previous state-of-the-art~\cite{iizuka2017globally}. For training, we use images of resolution \(256 \times 256\) with largest hole size \(128 \times 128\) described in Section~\ref{subsec:unified}. Both methods are based on fully-convolutional neural networks thus can fill in multiple holes on images of different resolutions. Visual comparisons on a variety of complex scenes from the validation set are shown in Figure~\ref{fig:expr_main_results}. Those test images are all with size \(512 \times 680\) for consistency of testing. All the results reported are direct outputs from the trained models without using any post-processing. For each example, we also visualize latent attention map for our model in the last column (color coding is explained in Section~ \ref{subsec:unified}).


As shown in the figure, our full model with contextual attention can leverage the surrounding textures and structures and consequently generates more realistic results with much less artifacts than the baseline model. Visualizations of attention maps reveal that our method is aware of contextual image structures and can adaptively borrow information from surrounding areas to help the synthesis and generation.

\begin{figure}[h]
\centering
\includegraphics[width=0.24\linewidth]{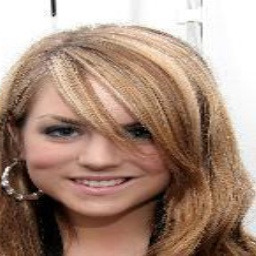}
\includegraphics[width=0.24\linewidth]{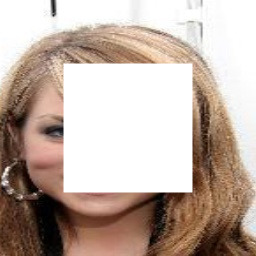}
\includegraphics[width=0.24\linewidth]{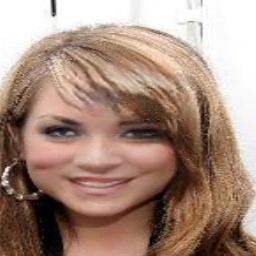}
\includegraphics[width=0.24\linewidth]{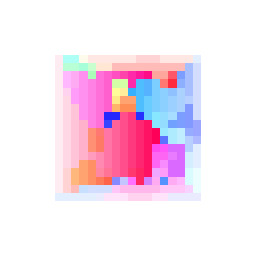}

\includegraphics[width=0.24\linewidth]{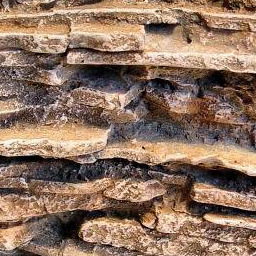}
\includegraphics[width=0.24\linewidth]{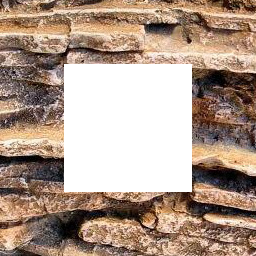}
\includegraphics[width=0.24\linewidth]{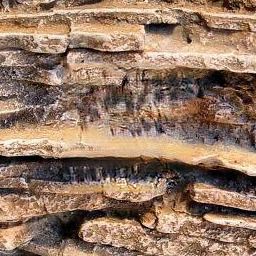}
\includegraphics[width=0.24\linewidth]{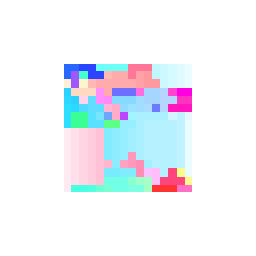}

\includegraphics[width=0.24\linewidth]{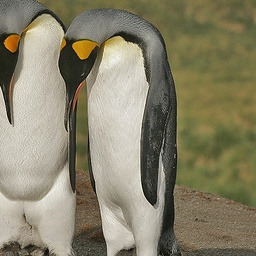}
\includegraphics[width=0.24\linewidth]{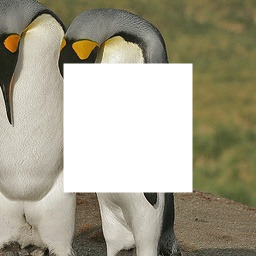}
\includegraphics[width=0.24\linewidth]{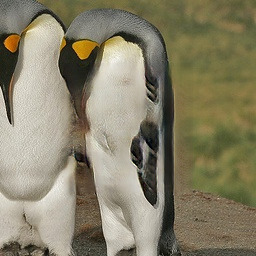}
\includegraphics[width=0.24\linewidth]{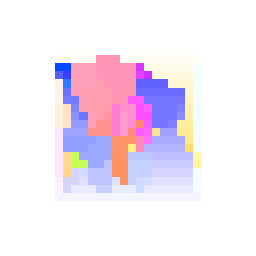}

\caption{Sample results of our model on CelebA faces, DTD textures and ImageNet from top to bottom. Each row, from left to right, shows original image, input image, result and attention map (upscaled \(4 \times\)), respectively.}
\label{fig:expr_main_others}
\end{figure}

In Figure~\ref{fig:expr_main_others}, we also show some example results and attention maps of our full model trained on CelebA, DTD and ImageNet. Due to space limitation, we include more results for these datasets in the supplementary material.

\textbf{Quantitative comparisons}
Like other image generation tasks, image inpainting lacks good quantitative evaluation metrics. Inception score~\cite{salimans2016improved} introduced for evaluating GAN models is not a good metric for evaluating image inpainting methods as inpainting mostly focuses on background filling (e.g.\ object removal case), not on its ability to generate a variety classes of objects.

Evaluation metrics in terms of reconstruction errors are also not perfect as there are many possible solutions different from the original image content. Nevertheless, we report our evaluation in terms of mean \(\ell_1\) error, mean \(\ell_2\) error, peak signal-to-noise ratio (PSNR) and total variation (TV) loss on validation set on Places2 just for reference in Table~\ref{tab:quantitative}. As shown in the table, learning-based methods perform better in terms of \(\ell_1\), \(\ell_2\) errors and PSNR, while methods directly copying raw image patches have lower total variation loss.
\begin{table}[h]
\begin{center}
\begin{tabular}{c c c c c}
\hline
Method & \(\ell_1\) loss & \(\ell_2\) loss & PSNR & TV loss\\
\hline
PatchMatch~\cite{barnes2009patchmatch} & 16.1\% & 3.9\% & 16.62 & \textbf{25.0}\%\\
Baseline model & 9.4\% & 2.4\% & 18.15 & 25.7\%\\
Our method & \textbf{8.6\%} & \textbf{2.1\%} & \textbf{18.91} & 25.3\%\\
\hline
\end{tabular}
\end{center}
\caption{Results of mean \(\ell_1\) eror, mean \(\ell_2\) error, PSNR and TV loss on validation set on Places2 for reference.}
\label{tab:quantitative}
\end{table}

Our full model has a total of \textbf{2.9M} parameters, which is roughly half of model proposed in~\cite{iizuka2017globally}. Models are implemented on TensorFlow v1.3, CUDNN v6.0, CUDA v8.0, and run on hardware with CPU Intel(R) Xeon(R) CPU E5-2697 v3 (2.60GHz) and GPU GTX 1080 Ti. Our full model runs at \textbf{0.2} seconds per frame on GPU and \textbf{1.5} seconds per frame on CPU for images of resolution \(\mathbf{512 \times 512}\) on average.


\subsection{Ablation study}

\begin{figure}[h]
\centering
\includegraphics[width=0.13\linewidth]{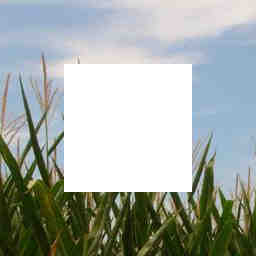}
\hspace{0.mm}
\includegraphics[width=0.13\linewidth]{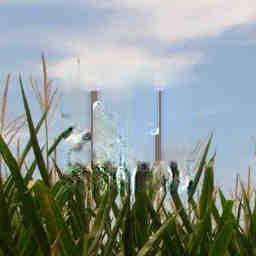}
\includegraphics[width=0.13\linewidth]{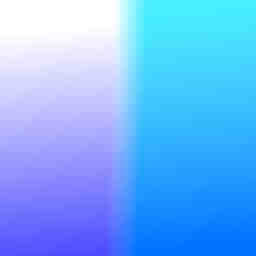}
\hspace{0.mm}
\includegraphics[width=0.13\linewidth]{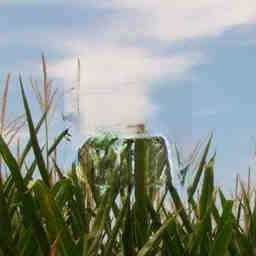}
\includegraphics[width=0.13\linewidth]{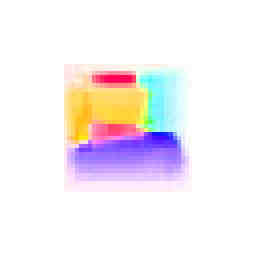}
\hspace{0.mm}
\includegraphics[width=0.13\linewidth]{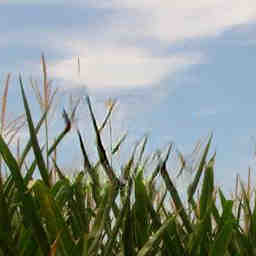}
\includegraphics[width=0.13\linewidth]{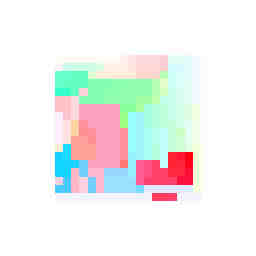}

\includegraphics[width=0.13\linewidth]{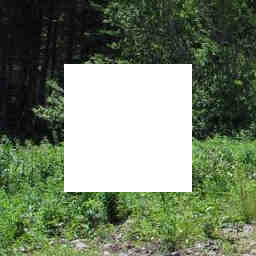}
\hspace{0.mm}
\includegraphics[width=0.13\linewidth]{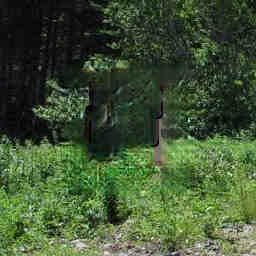}
\includegraphics[width=0.13\linewidth]{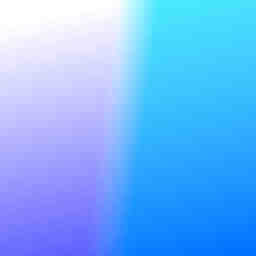}
\hspace{0.mm}
\includegraphics[width=0.13\linewidth]{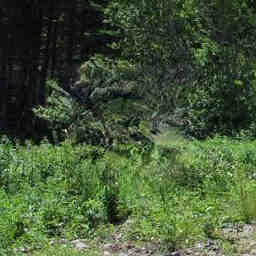}
\includegraphics[width=0.13\linewidth]{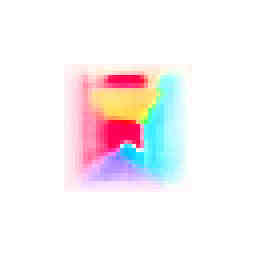}
\hspace{0.mm}
\includegraphics[width=0.13\linewidth]{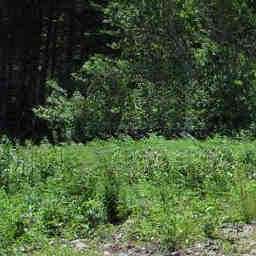}
\includegraphics[width=0.13\linewidth]{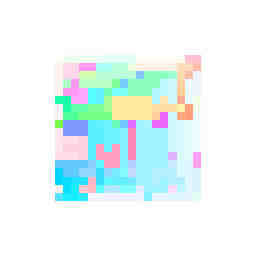}

\caption{We show input image, result and attention map using three different attention modules: spatial transformer network (left), appearance flow (middle), our contextual attention (right).}
\label{fig:expr_pixel_flow}
\end{figure}
\textbf{Contextual attention \vs spatial transformer network and appearance flow}
We investigate the effectiveness of contextual attention comparing to other spatial attention modules including appearance flow~\cite{zhou2016view} and spatial transformer network~\cite{jaderberg2015spatial} for image inpainting. For appearance flow~\cite{zhou2016view}, we train on the same framework except that the contextual attention layer is replaced with a convolution layer to directly predict 2-D pixel offsets as attention. As shown in Figure~\ref{fig:expr_pixel_flow}, for a very different test image pair, appearance flow returns very similar attention maps, meaning that the network may stuck in a bad local minima. To improve results of appearance flow, we also investigated ideas of multiple attention aggregation and patch-based attention. None of these ideas work well enough to improve the inpainting results. Also, we show the results with the spatial transformer network~\cite{jaderberg2015spatial} as attention in our framework in Figure~\ref{fig:expr_pixel_flow}. As shown in the figure, STN-based attention does not work well for inpainting as its global affine transformation is too coarse.

\begin{figure}[h]
\centering
\includegraphics[width=0.155\linewidth]{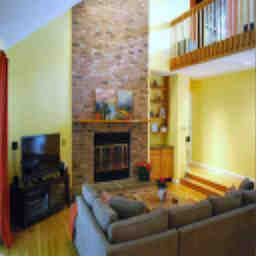}
\includegraphics[width=0.155\linewidth]{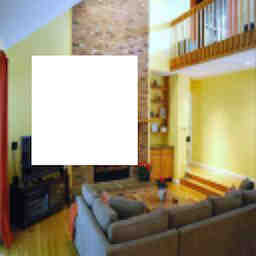}
\includegraphics[width=0.155\linewidth]{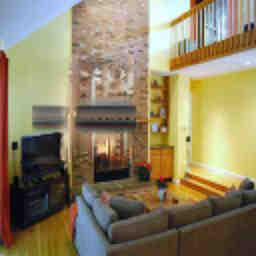}
\hspace{0.mm}
\includegraphics[width=0.155\linewidth]{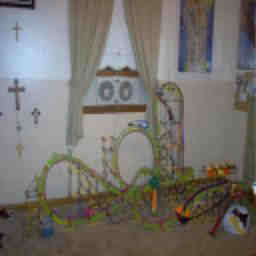}
\includegraphics[width=0.155\linewidth]{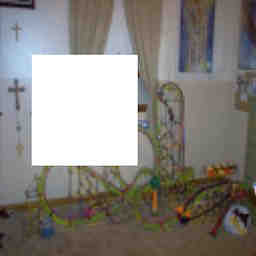}
\includegraphics[width=0.155\linewidth]{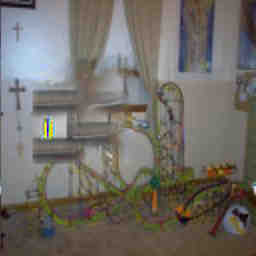}

\includegraphics[width=0.155\linewidth]{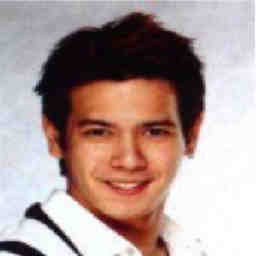}
\includegraphics[width=0.155\linewidth]{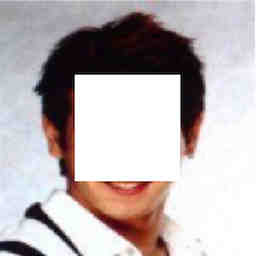}
\includegraphics[width=0.155\linewidth]{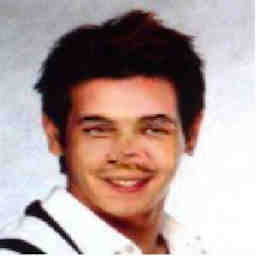}
\hspace{0.mm}
\includegraphics[width=0.155\linewidth]{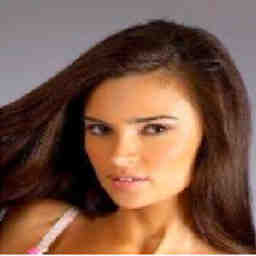}
\includegraphics[width=0.155\linewidth]{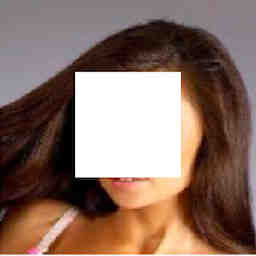}
\includegraphics[width=0.155\linewidth]{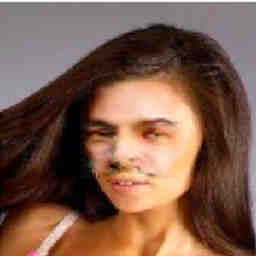}
\caption{Inpainting results of the model trained with DCGAN on Places2 (top) and CelebA (bottom) when modes collapse.}
\label{fig:expr_gan_comp}
\end{figure}
\textbf{Choice of the GAN loss for image inpainting}
Our inpainting framework benefits greatly from the WGAN-GP loss as validated by its learning curves and faster/stabler convergence behaviors. The same model trained with DCGAN sometimes collapses to limited modes for the inpainting task, as shown in Figure~\ref{fig:expr_gan_comp}. We also experimented with LSGAN~\cite{mao2016least}, and the results were worse.

\textbf{Essential reconstruction loss}
We also performed testing if we could drop out the \(\ell_1\) reconstruction loss and purely rely on the adversarial loss (i.e.\ improved WGANs) to generate good results. To draw a conclusion, we train our inpainting model without \(\ell_1\) reconstruction loss in the refinement network. Our conclusion is that the pixel-wise reconstruction loss, although tends to make the result blurry, is an essential ingredient for image inpainting. The reconstruction loss is helpful in capturing content structures and serves as a powerful regularization term for training GANs. 

\textbf{Perceptual loss, style loss and total variation loss} We have not found perceptual loss (reconstruction loss on VGG features), style loss (squared Frobenius norm of Gram matrix computed on the VGG features)~\cite{johnson2016perceptual} and total variation (TV) loss bring noticeable improvements for image inpainting in our framework, thus are not used.

\section{Conclusion}
We proposed a coarse-to-fine generative image inpainting framework and introduced our baseline model as well as full model with a novel contextual attention module. We showed that the contextual attention module significantly improves image inpainting results by learning feature representations for explicitly matching and attending to relevant background patches. As a future work, we plan to extend the method to very high-resolution inpainting applications using ideas similar to progressive growing of GANs~\cite{karras2017progressive}. The proposed inpainting framework and contextual attention module can also be applied on conditional image generation, image editing and computational photography tasks including image-based rendering, image super-resolution, guided editing and many others.

{\small
\bibliographystyle{ieee}
\bibliography{egbib}
}

\appendix

\section{More Results on CelebA, CelebA-HQ, DTD and ImageNet}
\begin{figure*}[h]
\centering

\includegraphics[width=0.1\linewidth]{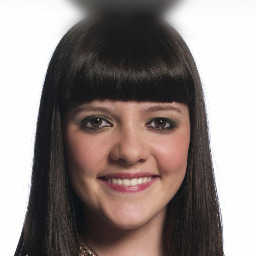}
\includegraphics[width=0.1\linewidth]{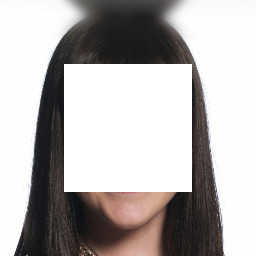}
\includegraphics[width=0.1\linewidth]{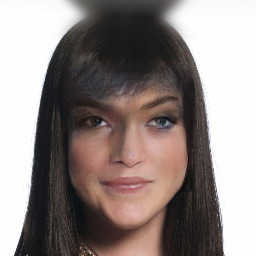}
\hspace{0.mm}
\includegraphics[width=0.1\linewidth]{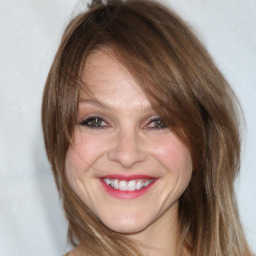}
\includegraphics[width=0.1\linewidth]{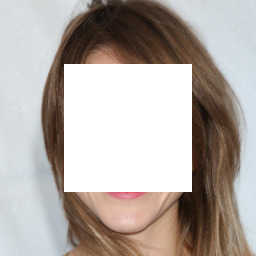}
\includegraphics[width=0.1\linewidth]{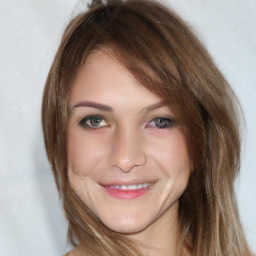}
\hspace{0.mm}
\includegraphics[width=0.1\linewidth]{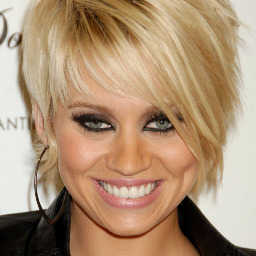}
\includegraphics[width=0.1\linewidth]{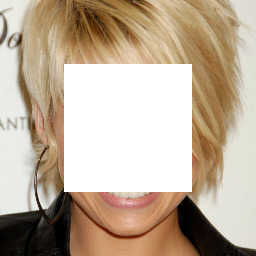}
\includegraphics[width=0.1\linewidth]{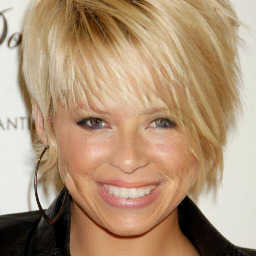}
\hspace{0.mm}

\includegraphics[width=0.1\linewidth]{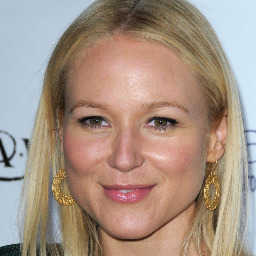}
\includegraphics[width=0.1\linewidth]{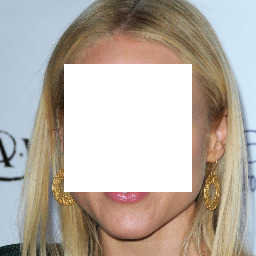}
\includegraphics[width=0.1\linewidth]{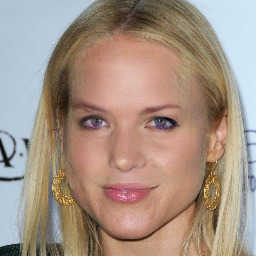}
\hspace{0.mm}
\includegraphics[width=0.1\linewidth]{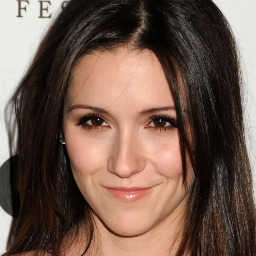}
\includegraphics[width=0.1\linewidth]{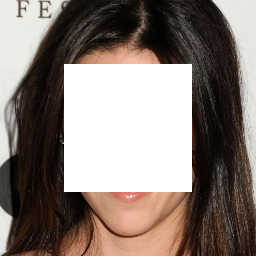}
\includegraphics[width=0.1\linewidth]{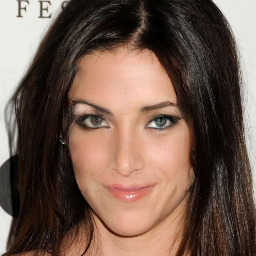}
\hspace{0.mm}
\includegraphics[width=0.1\linewidth]{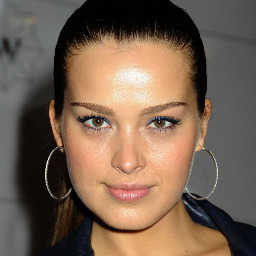}
\includegraphics[width=0.1\linewidth]{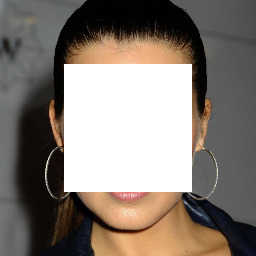}
\includegraphics[width=0.1\linewidth]{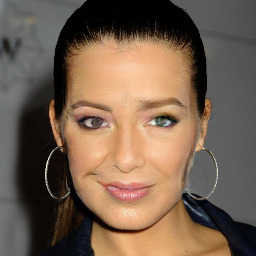}
\hspace{0.mm}

\caption{More inpainting results of our full model with contextual attention on CelebA-HQ faces. Each triad, from left to right, shows original image, input masked image and result image. All input images are masked from validation set (training and validation split is provided in released code). All results are direct outputs from same trained model without post-processing.}
\label{fig:supp_celeba_hq}
\end{figure*}

\textbf{CelebA-HQ~\cite{karras2017progressive}}
We show results from our full model trained on CelebA-HQ dataset in Figure~\ref{fig:supp_celeba_hq}. Note that the original image resolution of CelebA-HQ dataset is \(1024 \times 1024\). We resize image to \(256 \times 256\) for both training and evaluation.

\begin{figure*}[h]
\centering

\includegraphics[width=0.1\linewidth]{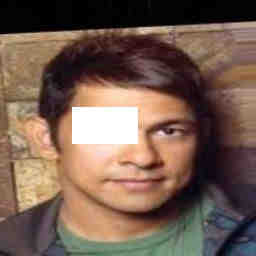}
\includegraphics[width=0.1\linewidth]{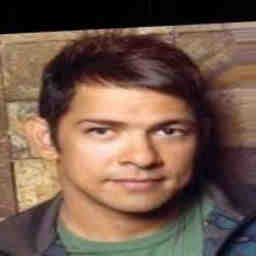}
\includegraphics[width=0.1\linewidth]{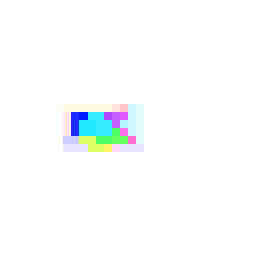}
\hspace{0.mm}
\includegraphics[width=0.1\linewidth]{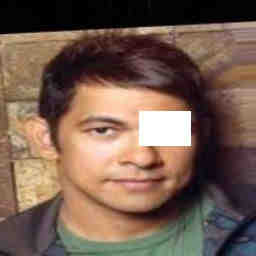}
\includegraphics[width=0.1\linewidth]{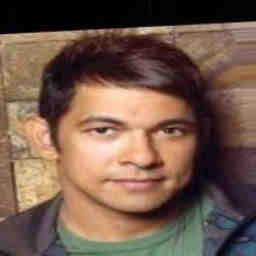}
\includegraphics[width=0.1\linewidth]{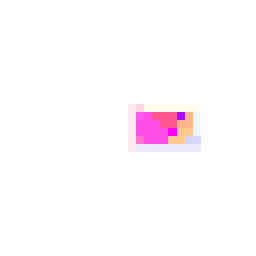}
\hspace{0.mm}
\includegraphics[width=0.1\linewidth]{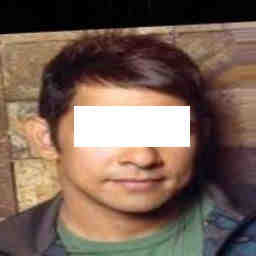}
\includegraphics[width=0.1\linewidth]{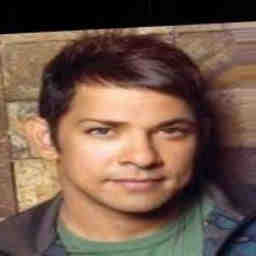}
\includegraphics[width=0.1\linewidth]{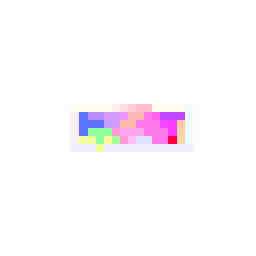}
\hspace{0.mm}
\includegraphics[width=0.1\linewidth]{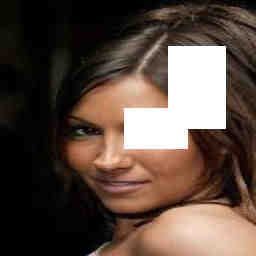}
\includegraphics[width=0.1\linewidth]{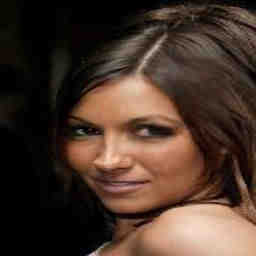}
\includegraphics[width=0.1\linewidth]{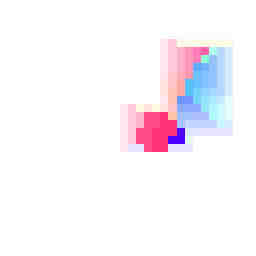}
\hspace{0.mm}
\includegraphics[width=0.1\linewidth]{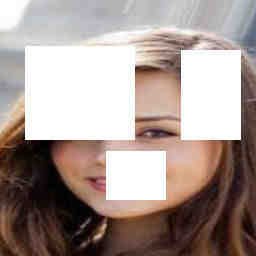}
\includegraphics[width=0.1\linewidth]{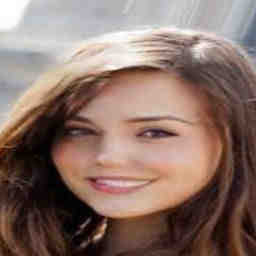}
\includegraphics[width=0.1\linewidth]{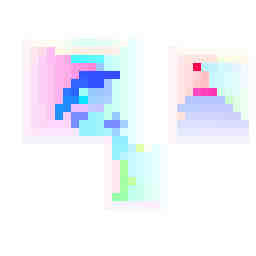}
\hspace{0.mm}
\includegraphics[width=0.1\linewidth]{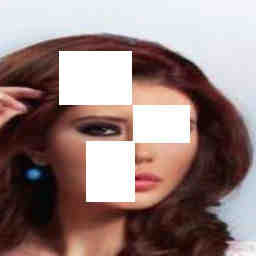}
\includegraphics[width=0.1\linewidth]{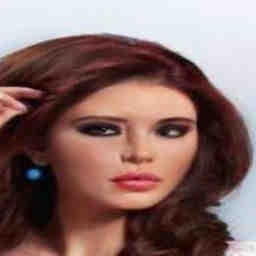}
\includegraphics[width=0.1\linewidth]{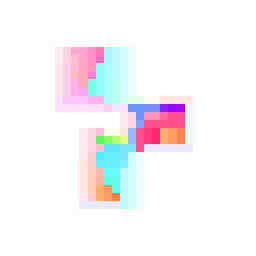}
\includegraphics[width=0.1\linewidth]{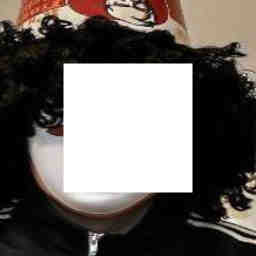}
\includegraphics[width=0.1\linewidth]{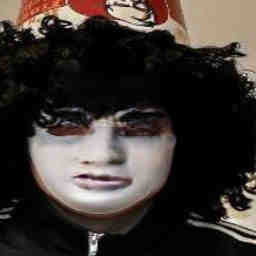}
\includegraphics[width=0.1\linewidth]{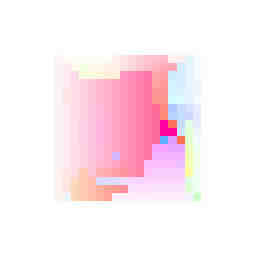}
\hspace{0.mm}
\includegraphics[width=0.1\linewidth]{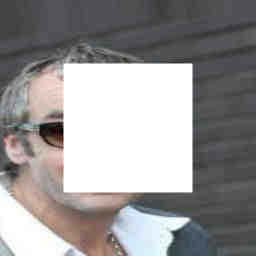}
\includegraphics[width=0.1\linewidth]{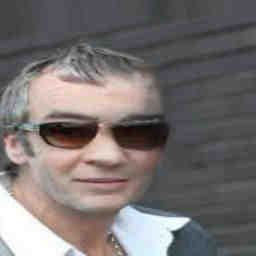}
\includegraphics[width=0.1\linewidth]{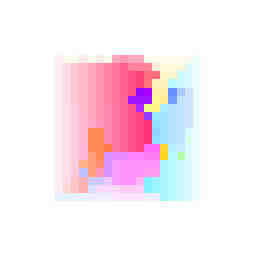}
\hspace{0.mm}
\includegraphics[width=0.1\linewidth]{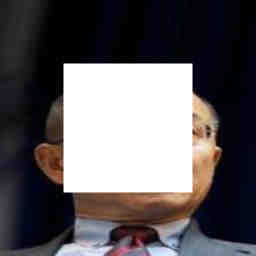}
\includegraphics[width=0.1\linewidth]{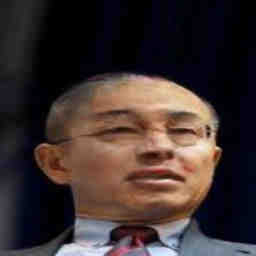}
\includegraphics[width=0.1\linewidth]{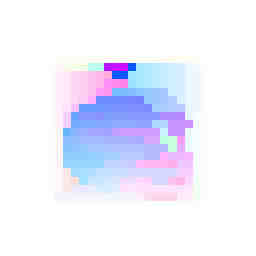}
\includegraphics[width=0.1\linewidth]{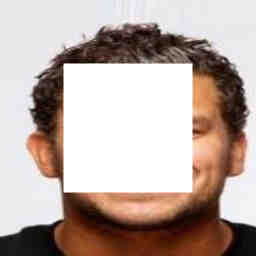}
\includegraphics[width=0.1\linewidth]{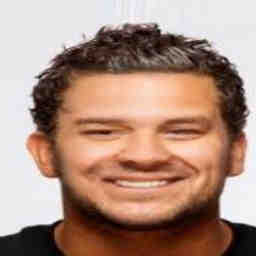}
\includegraphics[width=0.1\linewidth]{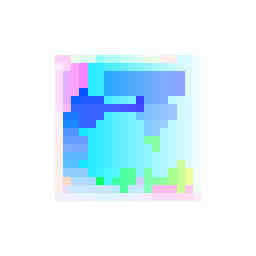}
\hspace{0.mm}
\includegraphics[width=0.1\linewidth]{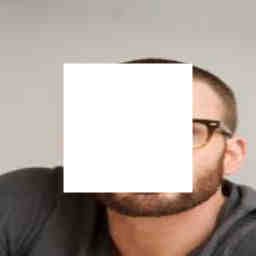}
\includegraphics[width=0.1\linewidth]{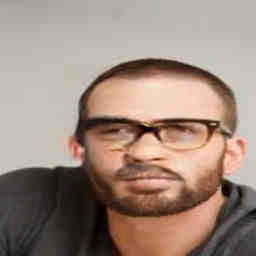}
\includegraphics[width=0.1\linewidth]{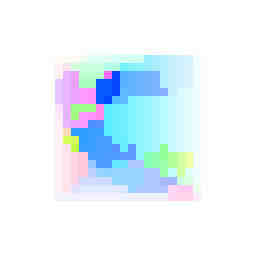}
\hspace{0.mm}
\includegraphics[width=0.1\linewidth]{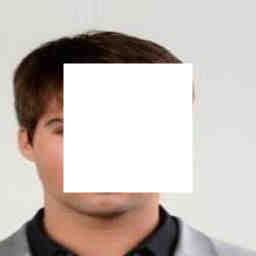}
\includegraphics[width=0.1\linewidth]{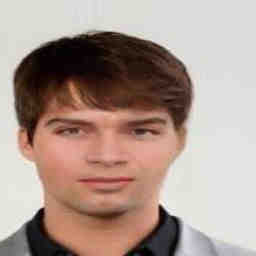}
\includegraphics[width=0.1\linewidth]{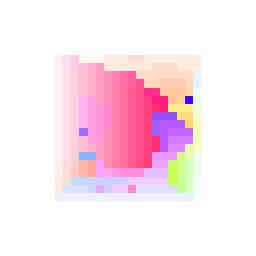}
\includegraphics[width=0.1\linewidth]{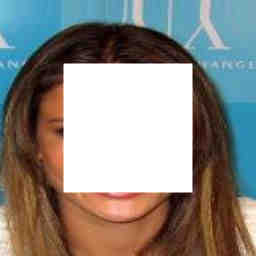}
\includegraphics[width=0.1\linewidth]{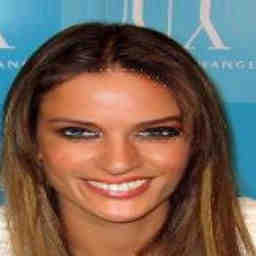}
\includegraphics[width=0.1\linewidth]{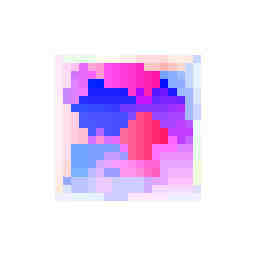}
\hspace{0.mm}
\includegraphics[width=0.1\linewidth]{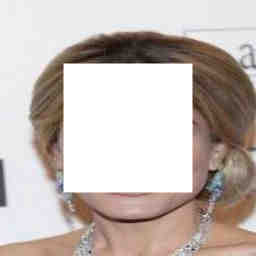}
\includegraphics[width=0.1\linewidth]{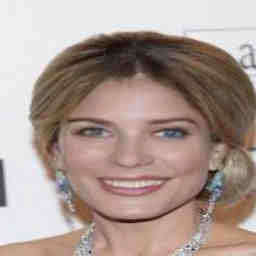}
\includegraphics[width=0.1\linewidth]{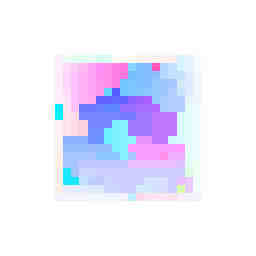}
\hspace{0.mm}
\includegraphics[width=0.1\linewidth]{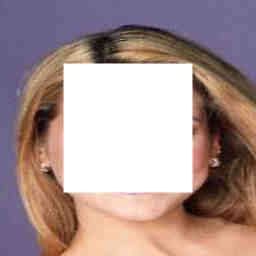}
\includegraphics[width=0.1\linewidth]{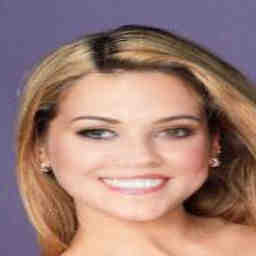}
\includegraphics[width=0.1\linewidth]{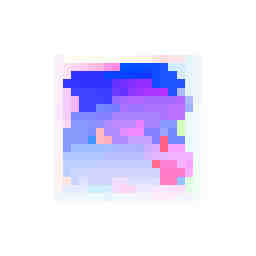}
\includegraphics[width=0.1\linewidth]{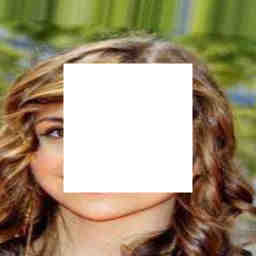}
\includegraphics[width=0.1\linewidth]{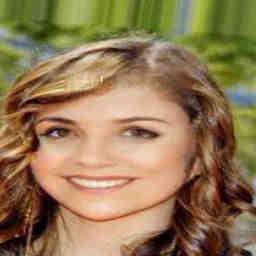}
\includegraphics[width=0.1\linewidth]{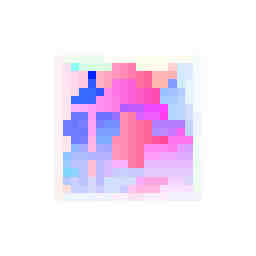}
\hspace{0.mm}
\includegraphics[width=0.1\linewidth]{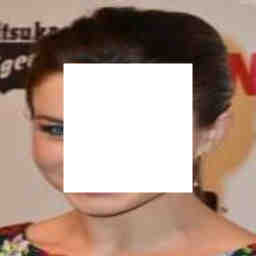}
\includegraphics[width=0.1\linewidth]{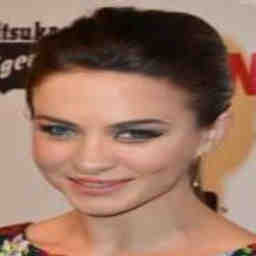}
\includegraphics[width=0.1\linewidth]{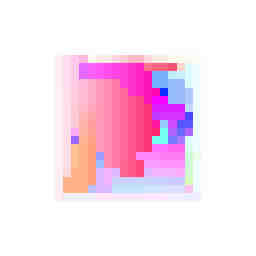}
\hspace{0.mm}
\includegraphics[width=0.1\linewidth]{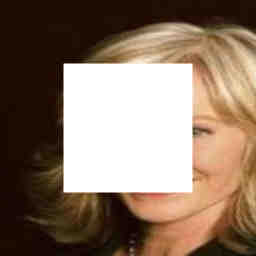}
\includegraphics[width=0.1\linewidth]{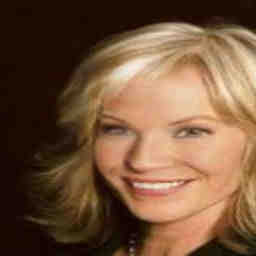}
\includegraphics[width=0.1\linewidth]{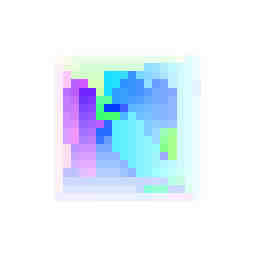}
\includegraphics[width=0.1\linewidth]{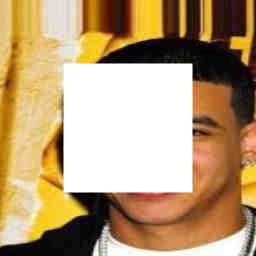}
\includegraphics[width=0.1\linewidth]{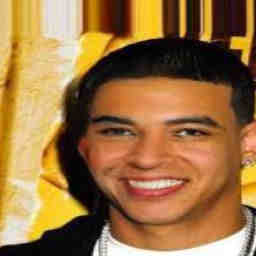}
\includegraphics[width=0.1\linewidth]{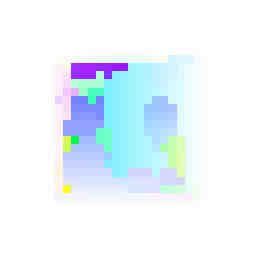}
\hspace{0.mm}
\includegraphics[width=0.1\linewidth]{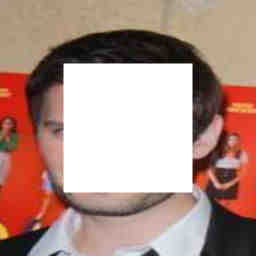}
\includegraphics[width=0.1\linewidth]{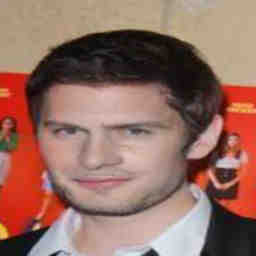}
\includegraphics[width=0.1\linewidth]{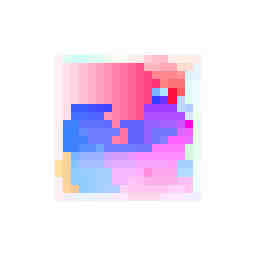}
\hspace{0.mm}
\includegraphics[width=0.1\linewidth]{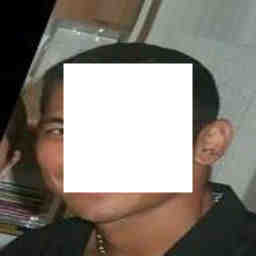}
\includegraphics[width=0.1\linewidth]{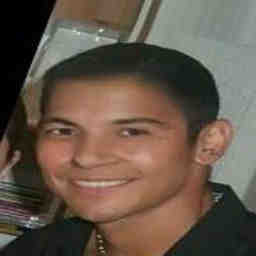}
\includegraphics[width=0.1\linewidth]{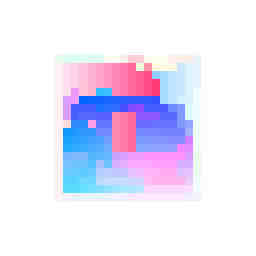}
\includegraphics[width=0.1\linewidth]{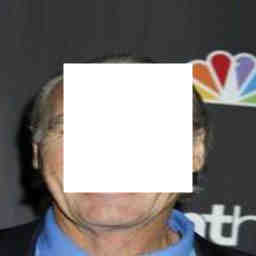}
\includegraphics[width=0.1\linewidth]{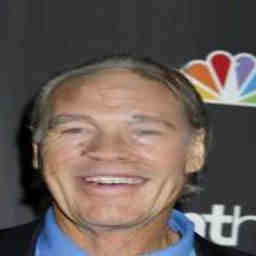}
\includegraphics[width=0.1\linewidth]{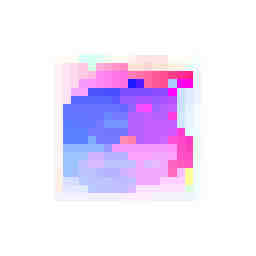}
\hspace{0.mm}
\includegraphics[width=0.1\linewidth]{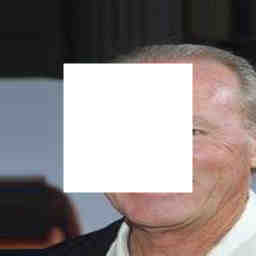}
\includegraphics[width=0.1\linewidth]{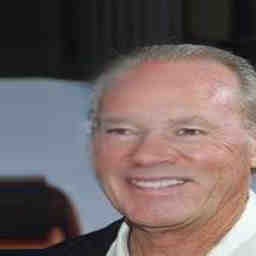}
\includegraphics[width=0.1\linewidth]{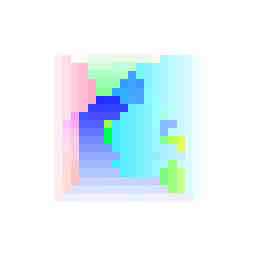}
\hspace{0.mm}
\includegraphics[width=0.1\linewidth]{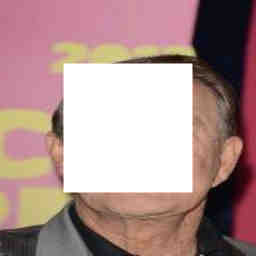}
\includegraphics[width=0.1\linewidth]{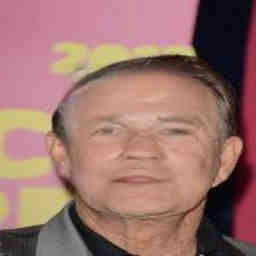}
\includegraphics[width=0.1\linewidth]{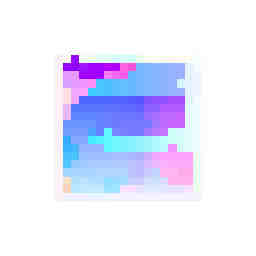}

\caption{More inpainting results of our full model with contextual attention on CelebA faces. Each triad, from left to right, shows input image, result and attention map (upscaled \(4 \times\)). All input images are masked from validation set (face identities are NOT overlapped between training set and validation set). All results are direct outputs from same trained model without post-processing.}
\label{fig:supp_celeba}
\end{figure*}

\textbf{CelebA~\cite{liu2015faceattributes}}
We show more results from our full model trained on CelebA dataset in Figure~\ref{fig:supp_celeba}. Note that the original image resolution of CelebA dataset is \(218 \times 178\). We resize image to \(315 \times 256\) and do a random crop of size \(256 \times 256\) to make face landmarks roughly unaligned for both training and evaluation.

\begin{figure*}[h]
\centering

\includegraphics[width=0.1\linewidth]{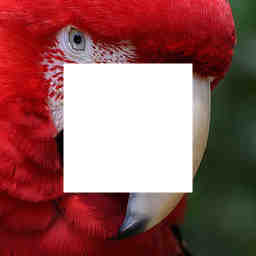}
\includegraphics[width=0.1\linewidth]{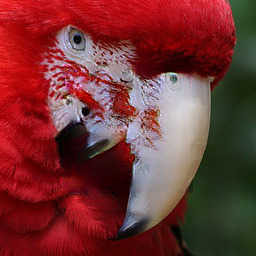}
\includegraphics[width=0.1\linewidth]{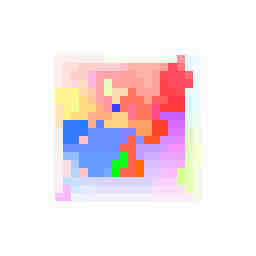}
\hspace{0.mm}
\includegraphics[width=0.1\linewidth]{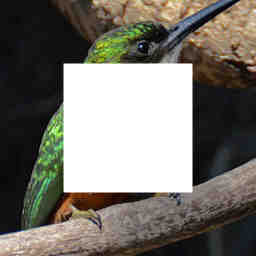}
\includegraphics[width=0.1\linewidth]{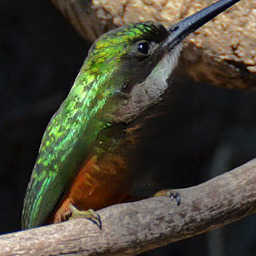}
\includegraphics[width=0.1\linewidth]{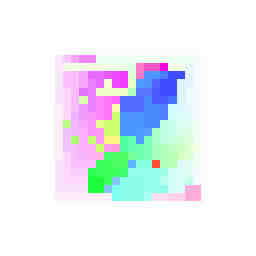}
\hspace{0.mm}
\includegraphics[width=0.1\linewidth]{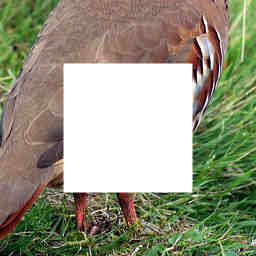}
\includegraphics[width=0.1\linewidth]{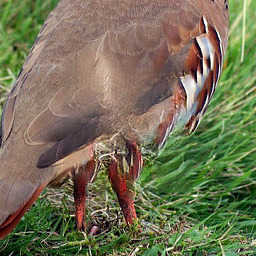}
\includegraphics[width=0.1\linewidth]{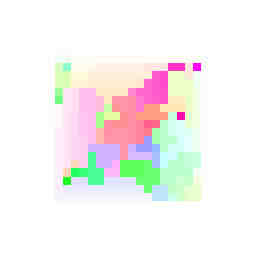}

\includegraphics[width=0.1\linewidth]{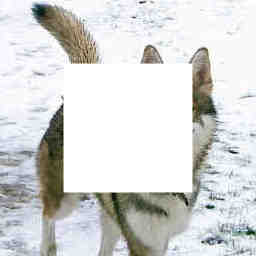}
\includegraphics[width=0.1\linewidth]{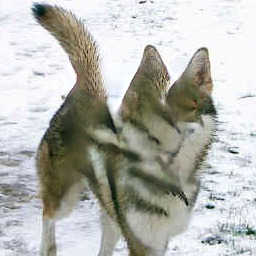}
\includegraphics[width=0.1\linewidth]{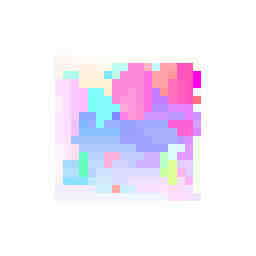}
\hspace{0.mm}
\includegraphics[width=0.1\linewidth]{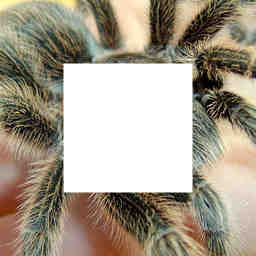}
\includegraphics[width=0.1\linewidth]{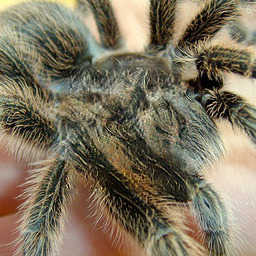}
\includegraphics[width=0.1\linewidth]{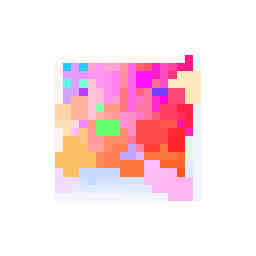}
\hspace{0.mm}
\includegraphics[width=0.1\linewidth]{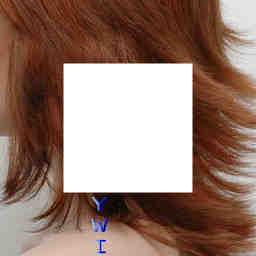}
\includegraphics[width=0.1\linewidth]{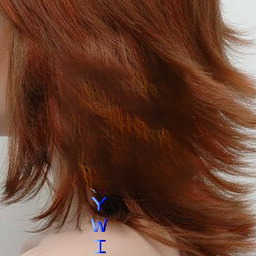}
\includegraphics[width=0.1\linewidth]{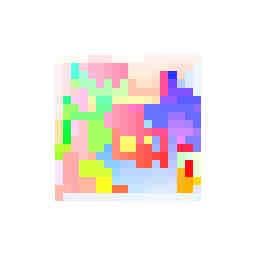}

\includegraphics[width=0.1\linewidth]{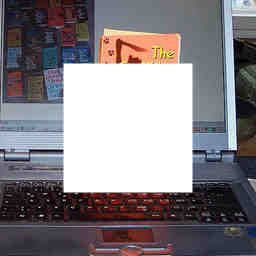}
\includegraphics[width=0.1\linewidth]{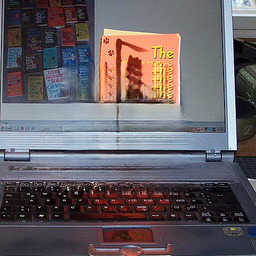}
\includegraphics[width=0.1\linewidth]{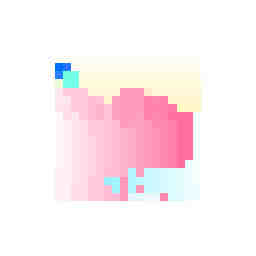}
\hspace{0.mm}
\includegraphics[width=0.1\linewidth]{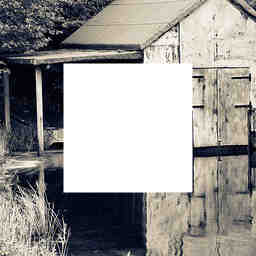}
\includegraphics[width=0.1\linewidth]{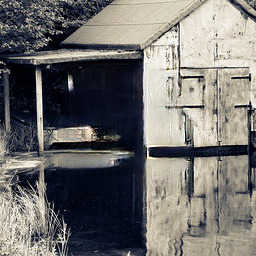}
\includegraphics[width=0.1\linewidth]{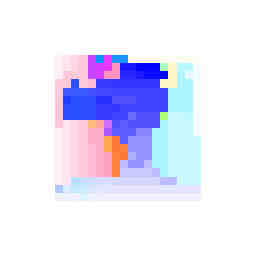}
\hspace{0.mm}
\includegraphics[width=0.1\linewidth]{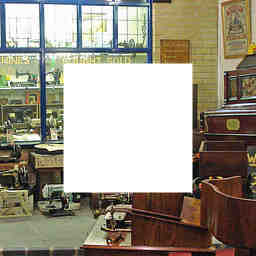}
\includegraphics[width=0.1\linewidth]{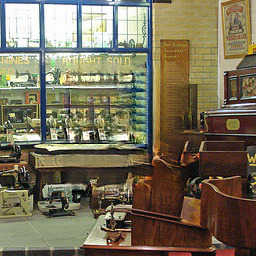}
\includegraphics[width=0.1\linewidth]{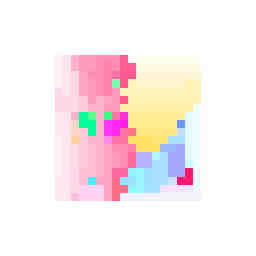}

\includegraphics[width=0.1\linewidth]{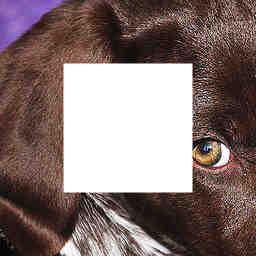}
\includegraphics[width=0.1\linewidth]{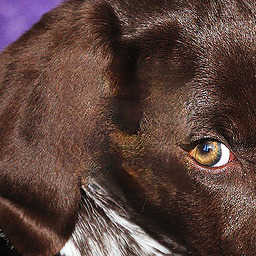}
\includegraphics[width=0.1\linewidth]{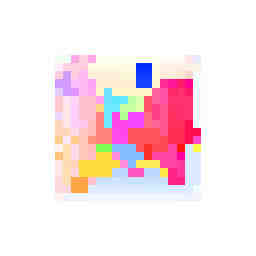}
\hspace{0.mm}
\includegraphics[width=0.1\linewidth]{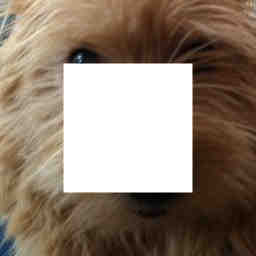}
\includegraphics[width=0.1\linewidth]{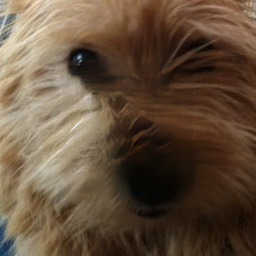}
\includegraphics[width=0.1\linewidth]{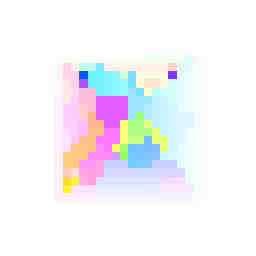}
\hspace{0.mm}
\includegraphics[width=0.1\linewidth]{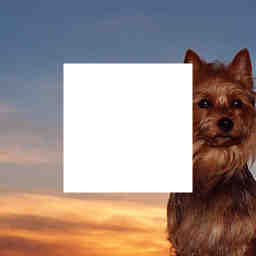}
\includegraphics[width=0.1\linewidth]{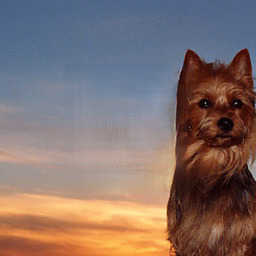}
\includegraphics[width=0.1\linewidth]{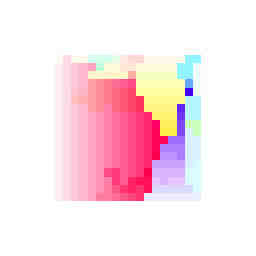}

\includegraphics[width=0.1\linewidth]{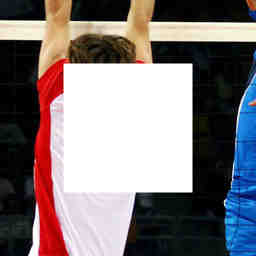}
\includegraphics[width=0.1\linewidth]{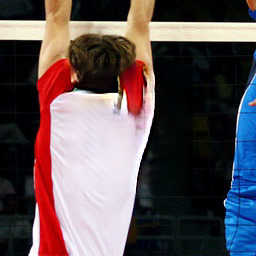}
\includegraphics[width=0.1\linewidth]{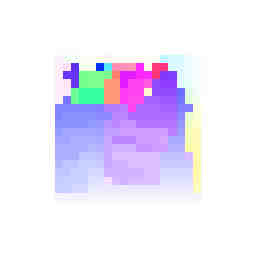}
\hspace{0.mm}
\includegraphics[width=0.1\linewidth]{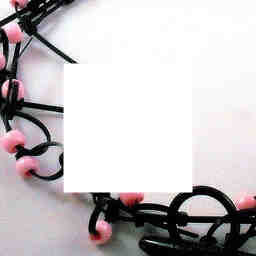}
\includegraphics[width=0.1\linewidth]{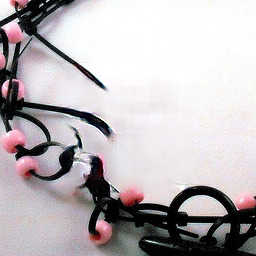}
\includegraphics[width=0.1\linewidth]{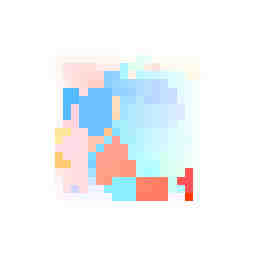}
\hspace{0.mm}
\includegraphics[width=0.1\linewidth]{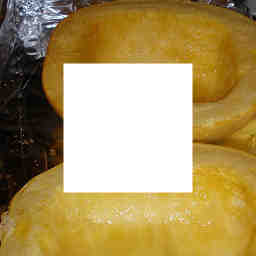}
\includegraphics[width=0.1\linewidth]{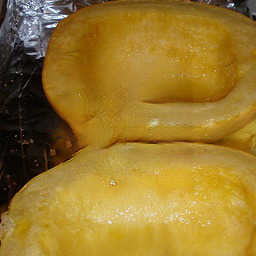}
\includegraphics[width=0.1\linewidth]{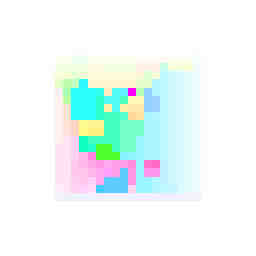}

\includegraphics[width=0.1\linewidth]{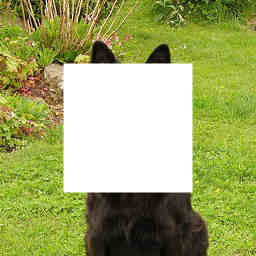}
\includegraphics[width=0.1\linewidth]{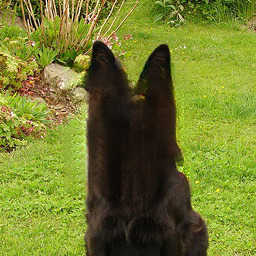}
\includegraphics[width=0.1\linewidth]{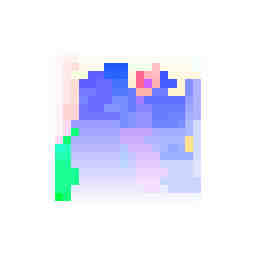}
\hspace{0.mm}
\includegraphics[width=0.1\linewidth]{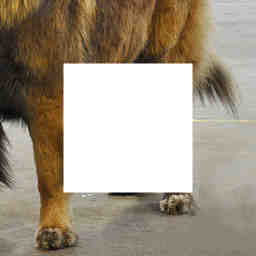}
\includegraphics[width=0.1\linewidth]{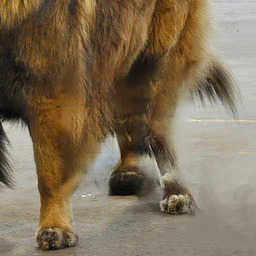}
\includegraphics[width=0.1\linewidth]{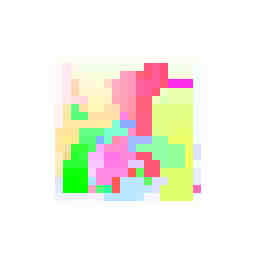}
\hspace{0.mm}
\includegraphics[width=0.1\linewidth]{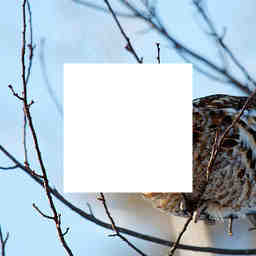}
\includegraphics[width=0.1\linewidth]{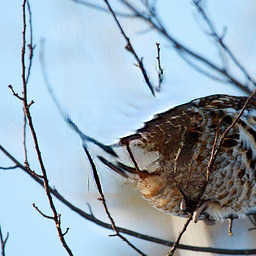}
\includegraphics[width=0.1\linewidth]{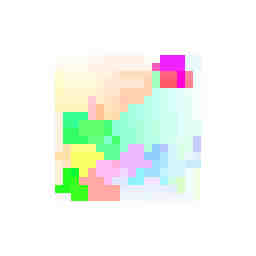}

\includegraphics[width=0.1\linewidth]{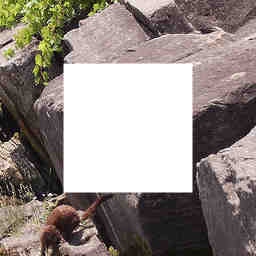}
\includegraphics[width=0.1\linewidth]{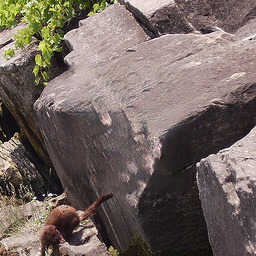}
\includegraphics[width=0.1\linewidth]{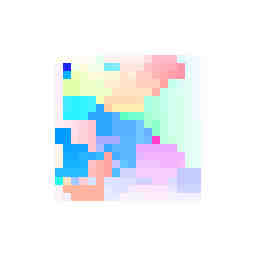}
\hspace{0.mm}
\includegraphics[width=0.1\linewidth]{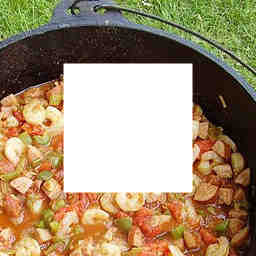}
\includegraphics[width=0.1\linewidth]{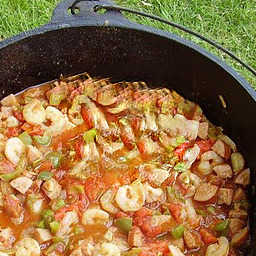}
\includegraphics[width=0.1\linewidth]{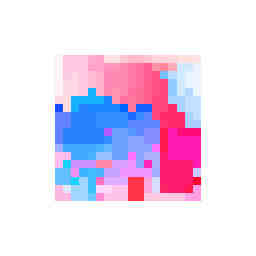}
\hspace{0.mm}
\includegraphics[width=0.1\linewidth]{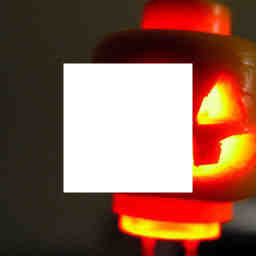}
\includegraphics[width=0.1\linewidth]{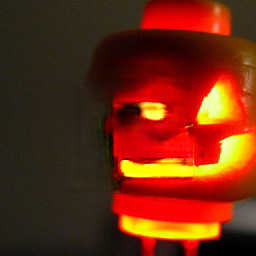}
\includegraphics[width=0.1\linewidth]{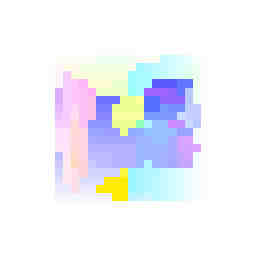}

\includegraphics[width=0.1\linewidth]{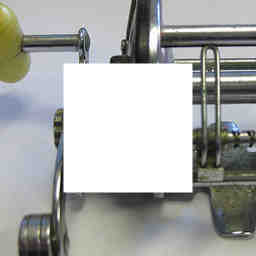}
\includegraphics[width=0.1\linewidth]{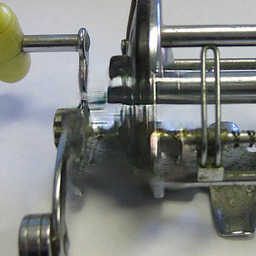}
\includegraphics[width=0.1\linewidth]{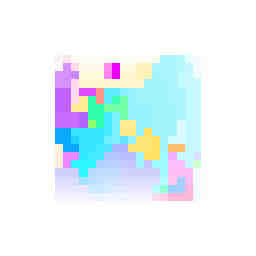}
\hspace{0.mm}
\includegraphics[width=0.1\linewidth]{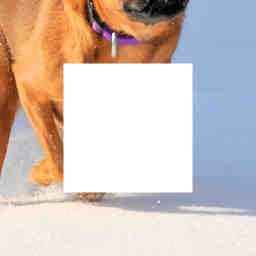}
\includegraphics[width=0.1\linewidth]{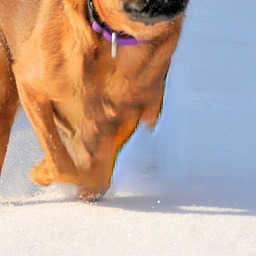}
\includegraphics[width=0.1\linewidth]{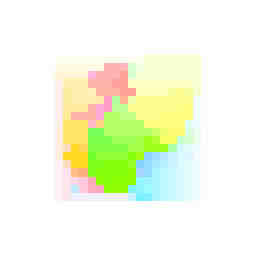}
\hspace{0.mm}
\includegraphics[width=0.1\linewidth]{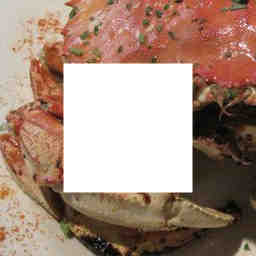}
\includegraphics[width=0.1\linewidth]{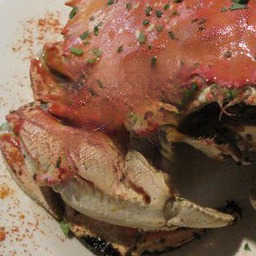}
\includegraphics[width=0.1\linewidth]{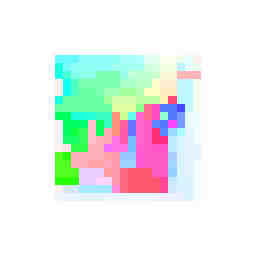}

\caption{More inpainting results of our full model with contextual attention on ImageNet. Each triad, from left to right, shows input image, result and attention map (upscaled \(4 \times\)). All input images are masked from validation set. All results are direct outputs from same trained model without post-processing.}
\label{fig:supp_imagenet}
\end{figure*}

\textbf{ImageNet~\cite{russakovsky2015imagenet}}
We show more results from our full model trained on ImageNet dataset in Figure~\ref{fig:supp_imagenet}.

\begin{figure*}[h]
\centering

\includegraphics[width=0.1\linewidth]{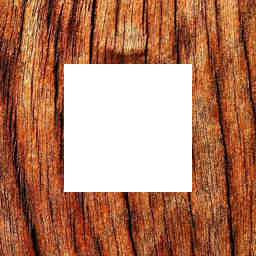}
\includegraphics[width=0.1\linewidth]{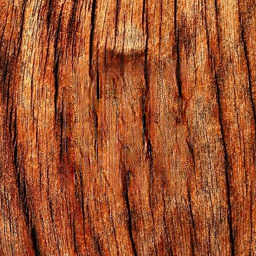}
\includegraphics[width=0.1\linewidth]{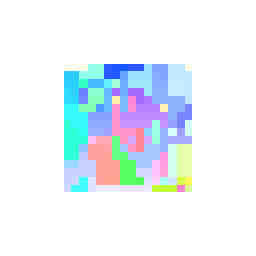}
\hspace{0.mm}
\includegraphics[width=0.1\linewidth]{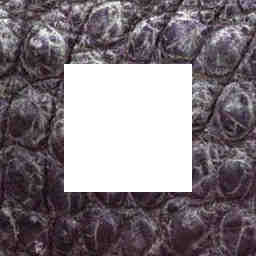}
\includegraphics[width=0.1\linewidth]{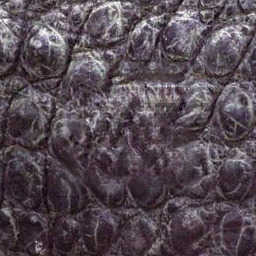}
\includegraphics[width=0.1\linewidth]{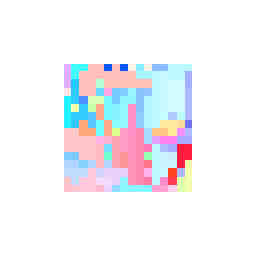}
\hspace{0.mm}
\includegraphics[width=0.1\linewidth]{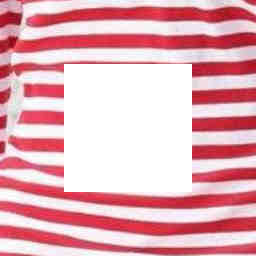}
\includegraphics[width=0.1\linewidth]{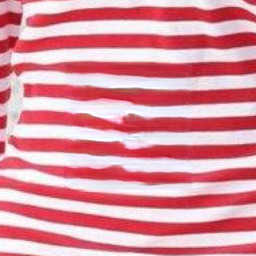}
\includegraphics[width=0.1\linewidth]{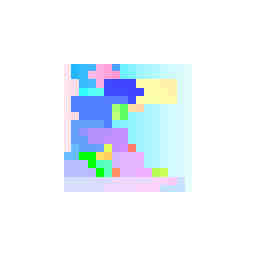}

\includegraphics[width=0.1\linewidth]{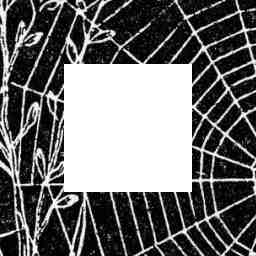}
\includegraphics[width=0.1\linewidth]{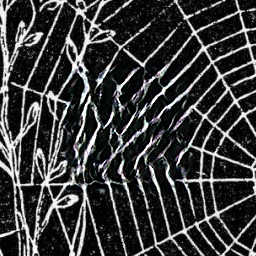}
\includegraphics[width=0.1\linewidth]{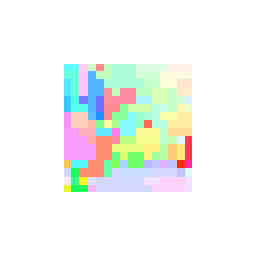}
\hspace{0.mm}
\includegraphics[width=0.1\linewidth]{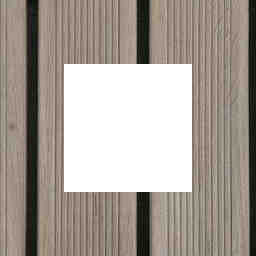}
\includegraphics[width=0.1\linewidth]{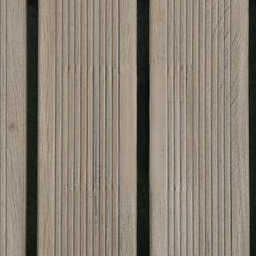}
\includegraphics[width=0.1\linewidth]{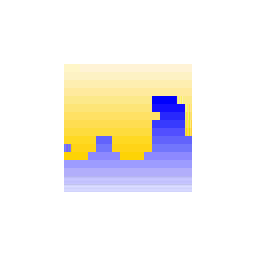}
\hspace{0.mm}
\includegraphics[width=0.1\linewidth]{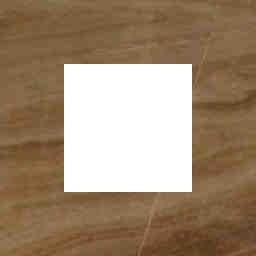}
\includegraphics[width=0.1\linewidth]{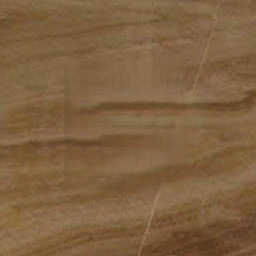}
\includegraphics[width=0.1\linewidth]{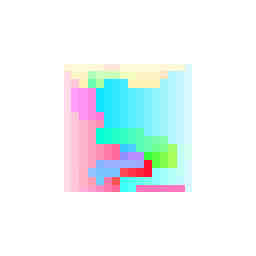}

\caption{More inpainting results of our full model with contextual attention on DTD textures. Each triad, from left to right, shows input image, result and attention map (upscaled \(4 \times\)). All input images are masked from validation set. All results are direct outputs from same trained model without post-processing.}
\label{fig:supp_dtd}
\end{figure*}

\textbf{DTD textures~\cite{cimpoi2014describing}}
We show more results from our full model trained on DTD dataset in Figure~\ref{fig:supp_dtd}.

\section{Comparisons with More Methods}
\begin{figure*}[h]
\centering
\begin{minipage}[t]{\textwidth}
\begin{minipage}[t]{0.24\textwidth}
\centering
\includegraphics[width=\linewidth]{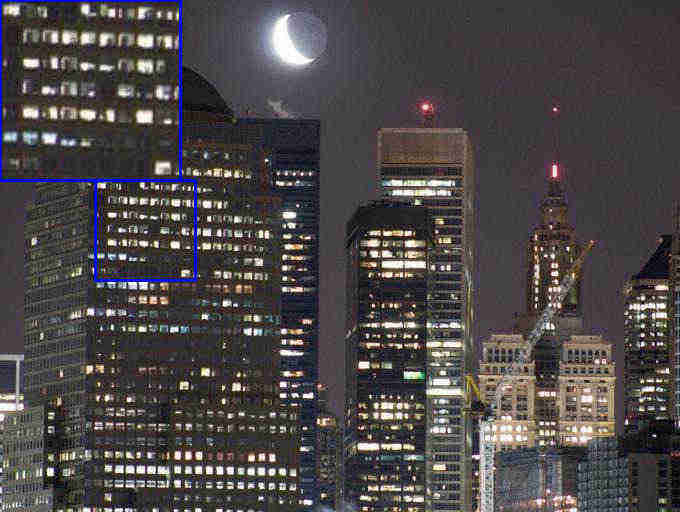}
original image
\end{minipage}
\begin{minipage}[t]{0.24\textwidth}
\centering
\includegraphics[width=\linewidth]{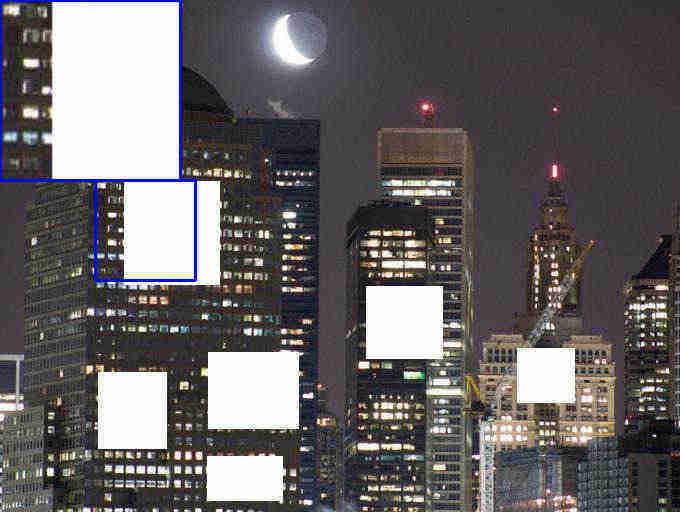}
input image
\end{minipage}
\begin{minipage}[t]{0.24\textwidth}
\centering
\includegraphics[width=\linewidth]{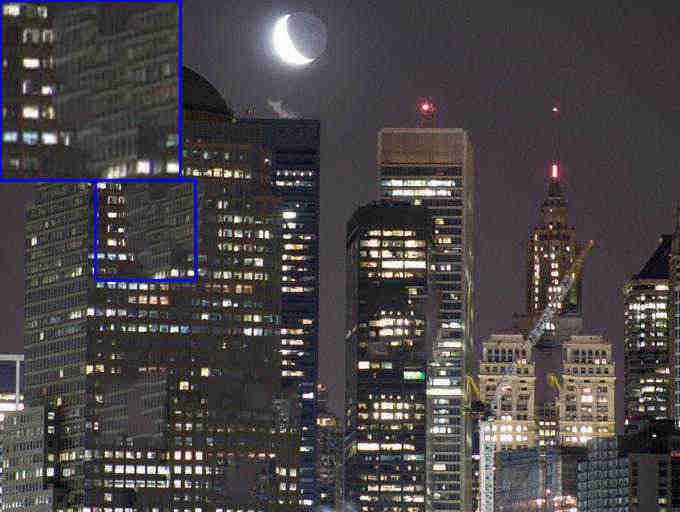}
Content-Aware Fill~\cite{barnes2009patchmatch}
\end{minipage}
\begin{minipage}[t]{0.24\textwidth}
\centering
\includegraphics[width=\linewidth]{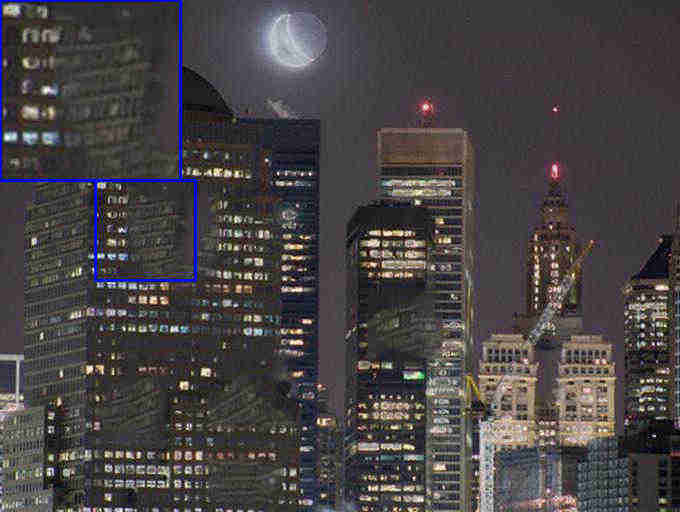}
ImageMelding~\cite{darabi2012image}
\end{minipage}
\end{minipage}

\begin{minipage}[t]{\textwidth}
\begin{minipage}[t]{0.24\textwidth}
\centering
\includegraphics[width=\linewidth]{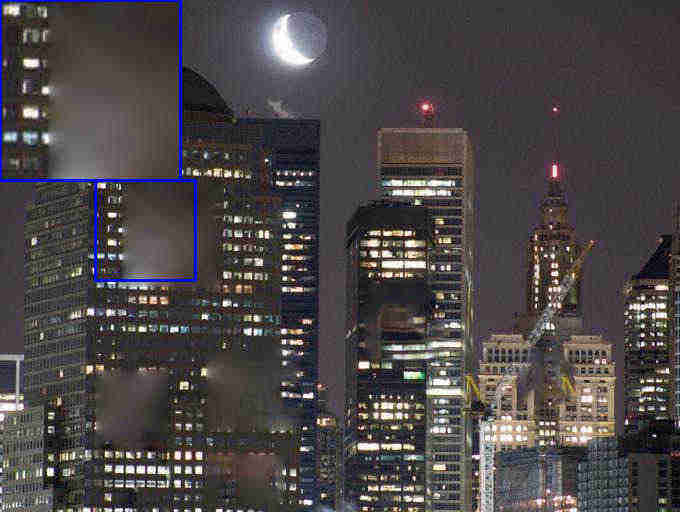}
StructCompletion~\cite{huang2014image}
\end{minipage}
\begin{minipage}[t]{0.24\textwidth}
\centering
\includegraphics[width=\linewidth]{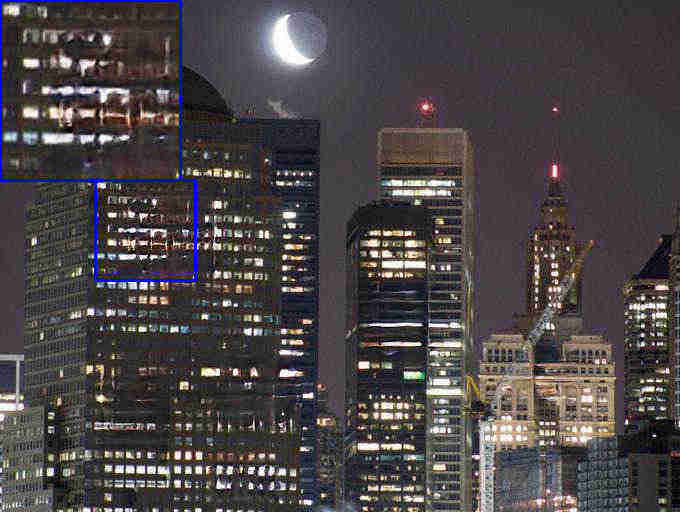}
our baseline model
\end{minipage}
\begin{minipage}[t]{0.24\textwidth}
\centering
\includegraphics[width=\linewidth]{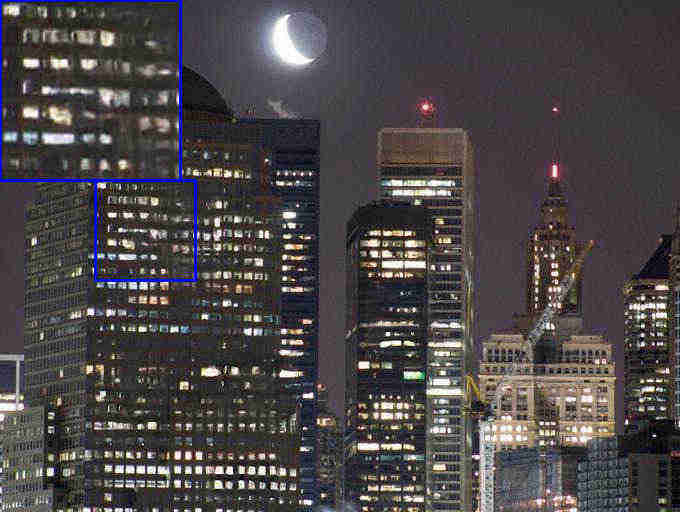}
our full model
\end{minipage}
\begin{minipage}[t]{0.24\textwidth}
\centering
\includegraphics[width=\linewidth]{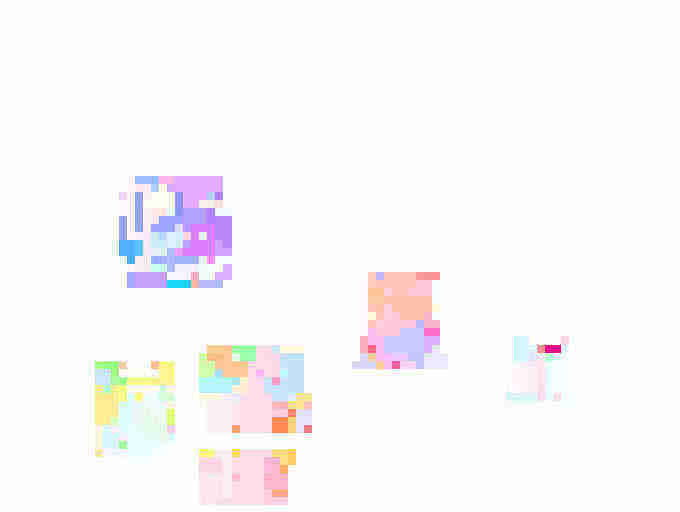}
attention map of full model
\end{minipage}
\end{minipage}

\begin{minipage}[t]{\textwidth}
\begin{minipage}[t]{0.24\textwidth}
\centering
\includegraphics[width=\linewidth]{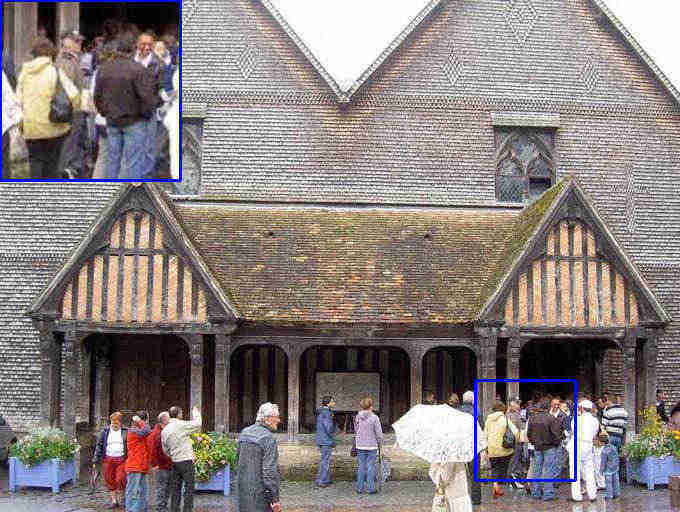}
original image
\end{minipage}
\begin{minipage}[t]{0.24\textwidth}
\centering
\includegraphics[width=\linewidth]{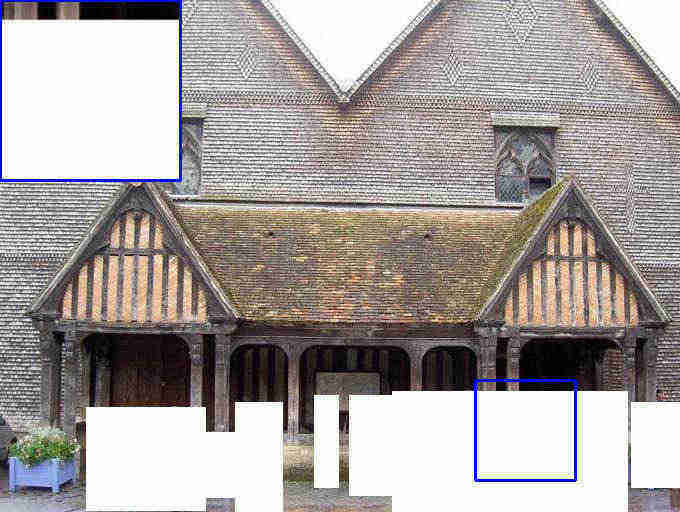}
input image
\end{minipage}
\begin{minipage}[t]{0.24\textwidth}
\centering
\includegraphics[width=\linewidth]{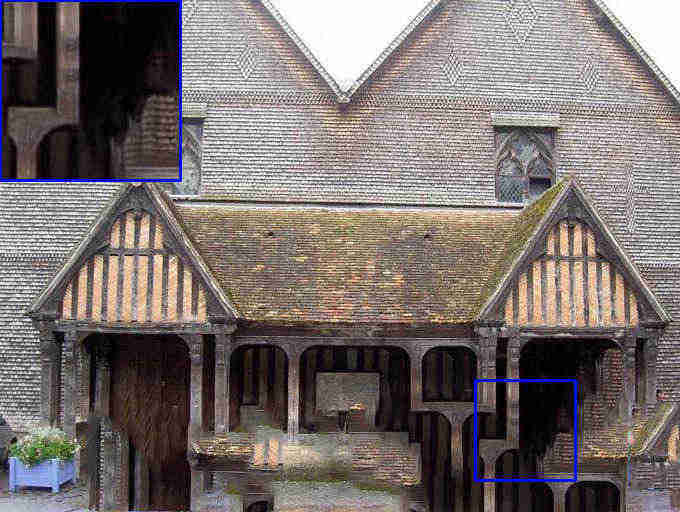}
Content-Aware Fill~\cite{barnes2009patchmatch}
\end{minipage}
\begin{minipage}[t]{0.24\textwidth}
\centering
\includegraphics[width=\linewidth]{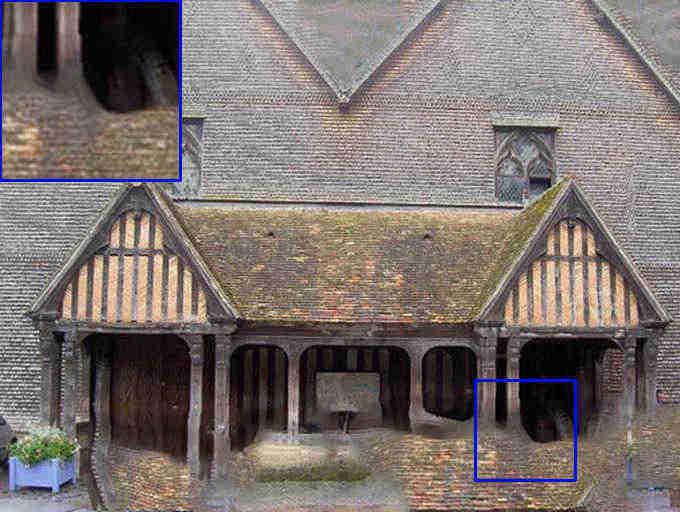}
ImageMelding~\cite{darabi2012image}
\end{minipage}
\end{minipage}

\begin{minipage}[t]{\textwidth}
\begin{minipage}[t]{0.24\textwidth}
\centering
\includegraphics[width=\linewidth]{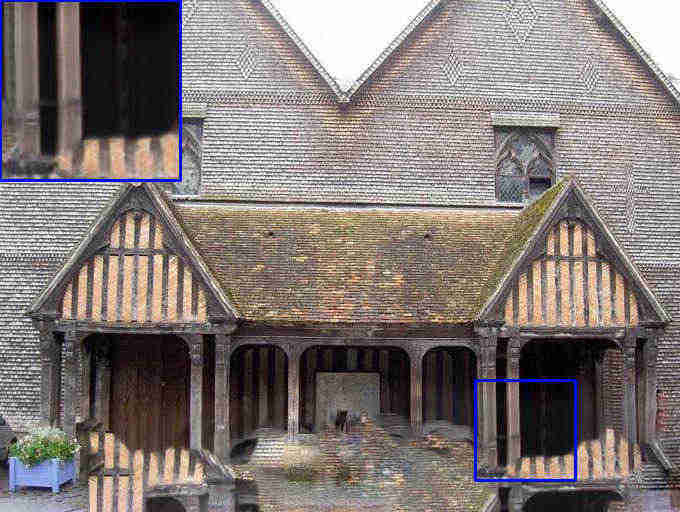}
StructCompletion~\cite{huang2014image}
\end{minipage}
\begin{minipage}[t]{0.24\textwidth}
\centering
\includegraphics[width=\linewidth]{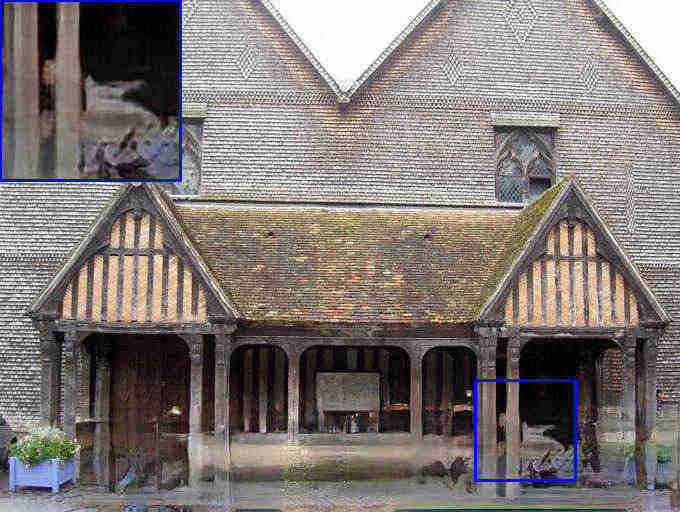}
our baseline model
\end{minipage}
\begin{minipage}[t]{0.24\textwidth}
\centering
\includegraphics[width=\linewidth]{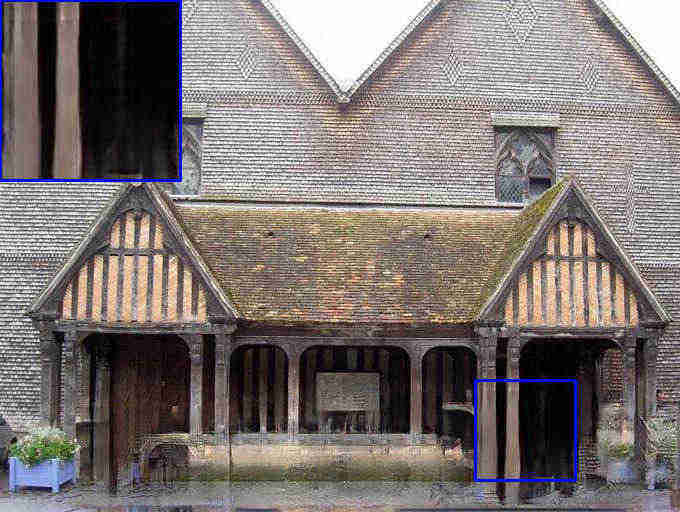}
our full model
\end{minipage}
\begin{minipage}[t]{0.24\textwidth}
\centering
\includegraphics[width=\linewidth]{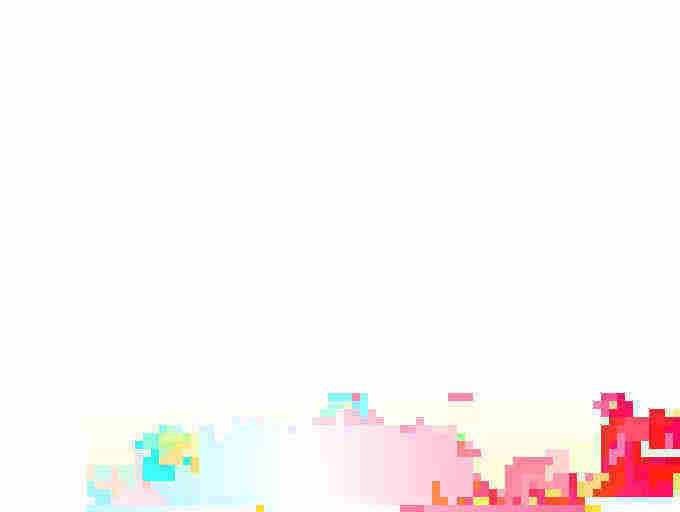}
attention map of full model
\end{minipage}
\end{minipage}

\begin{minipage}[t]{\textwidth}
\begin{minipage}[t]{0.24\textwidth}
\centering
\includegraphics[width=\linewidth]{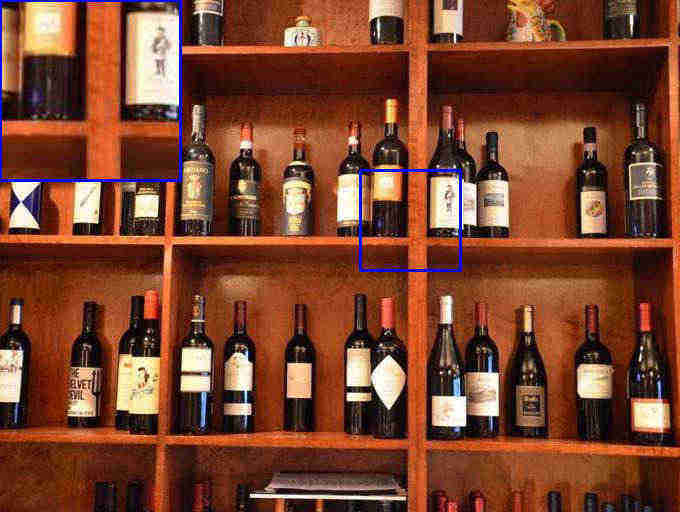}
original image
\end{minipage}
\begin{minipage}[t]{0.24\textwidth}
\centering
\includegraphics[width=\linewidth]{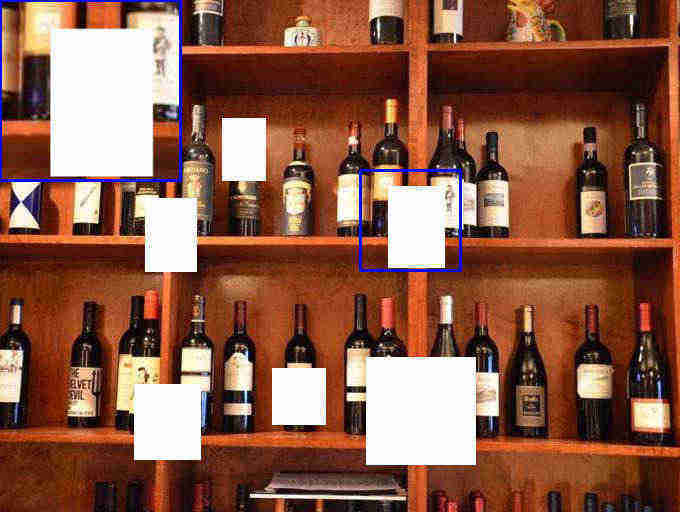}
input image
\end{minipage}
\begin{minipage}[t]{0.24\textwidth}
\centering
\includegraphics[width=\linewidth]{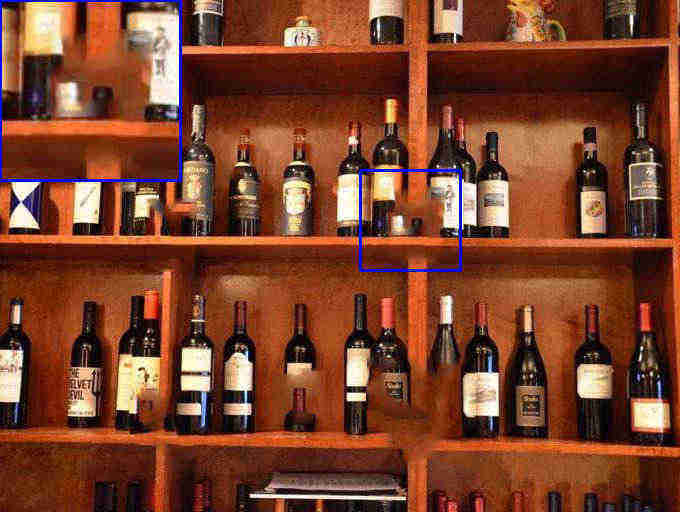}
Content-Aware Fill~\cite{barnes2009patchmatch}
\end{minipage}
\begin{minipage}[t]{0.24\textwidth}
\centering
\includegraphics[width=\linewidth]{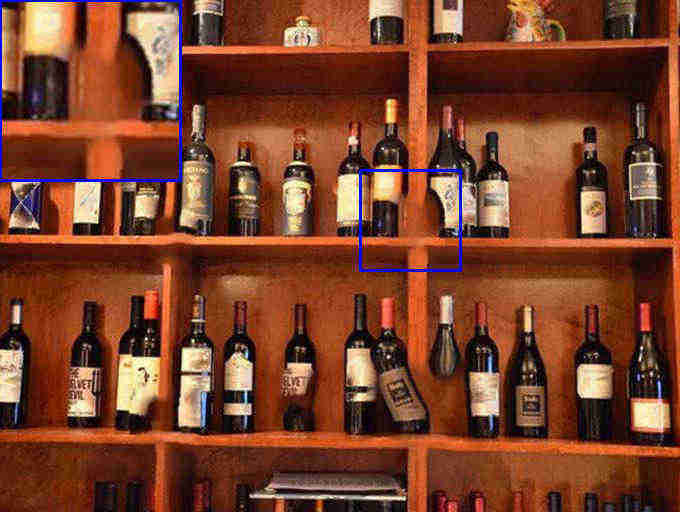}
ImageMelding~\cite{darabi2012image}
\end{minipage}
\end{minipage}

\begin{minipage}[t]{\textwidth}
\begin{minipage}[t]{0.24\textwidth}
\centering
\includegraphics[width=\linewidth]{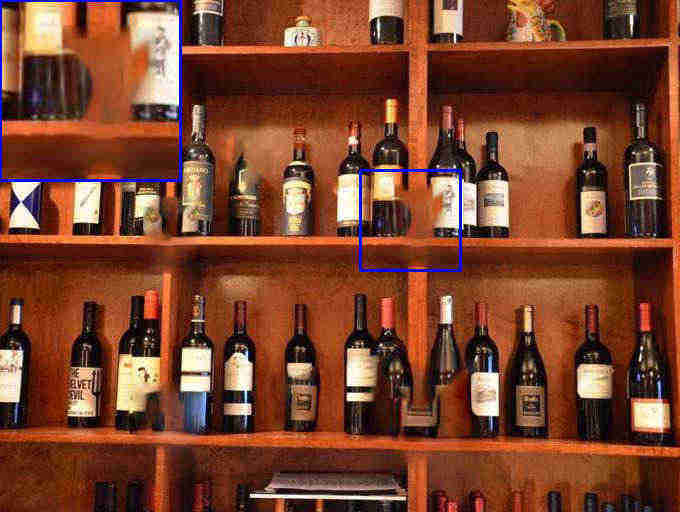}
StructCompletion~\cite{huang2014image}
\end{minipage}
\begin{minipage}[t]{0.24\textwidth}
\centering
\includegraphics[width=\linewidth]{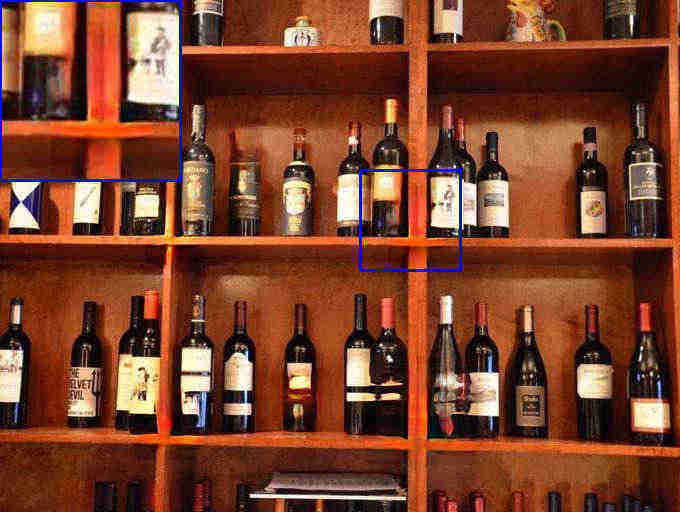}
our baseline model
\end{minipage}
\begin{minipage}[t]{0.24\textwidth}
\centering
\includegraphics[width=\linewidth]{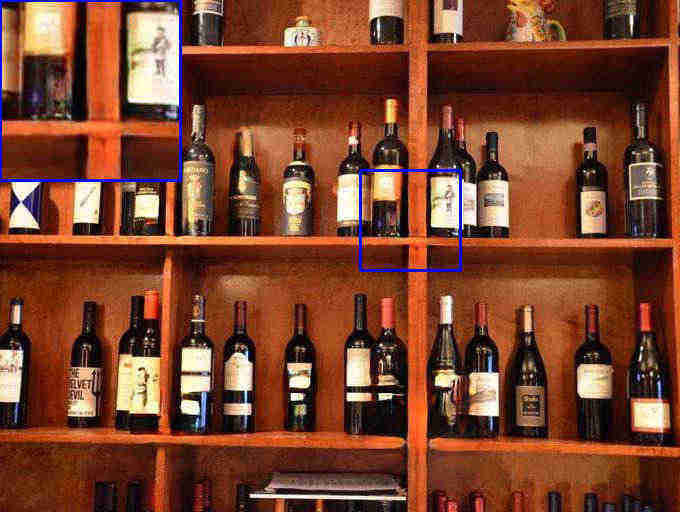}
our full model
\end{minipage}
\begin{minipage}[t]{0.24\textwidth}
\centering
\includegraphics[width=\linewidth]{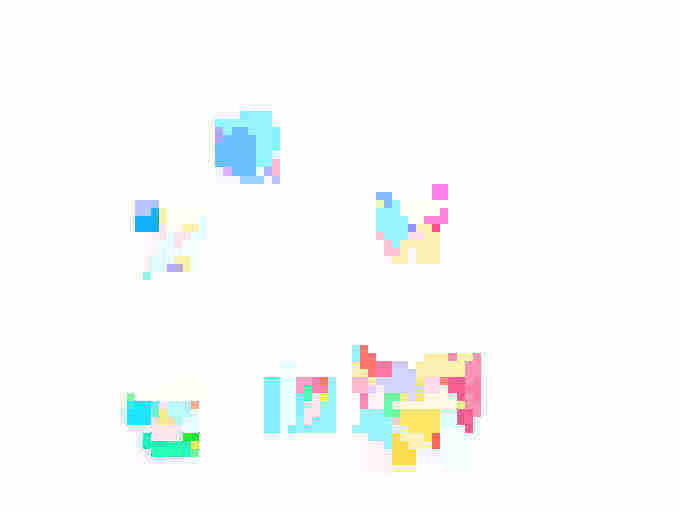}
attention map of full model
\end{minipage}
\end{minipage}

\caption{More qualitative results and comparisons.
All input images are masked from validation set. All our results are direct outputs from same trained model without post-processing. Best viewed with zoom-in.}
\label{fig:supp_places2_comparison1}
\end{figure*}

\begin{figure*}[h]
\centering
\begin{minipage}[t]{\textwidth}
\begin{minipage}[t]{0.24\textwidth}
\centering
\includegraphics[width=\linewidth]{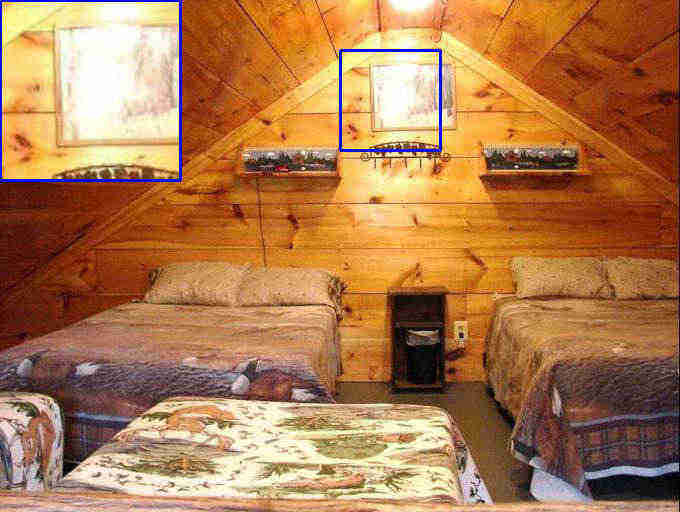}
original image
\end{minipage}
\begin{minipage}[t]{0.24\textwidth}
\centering
\includegraphics[width=\linewidth]{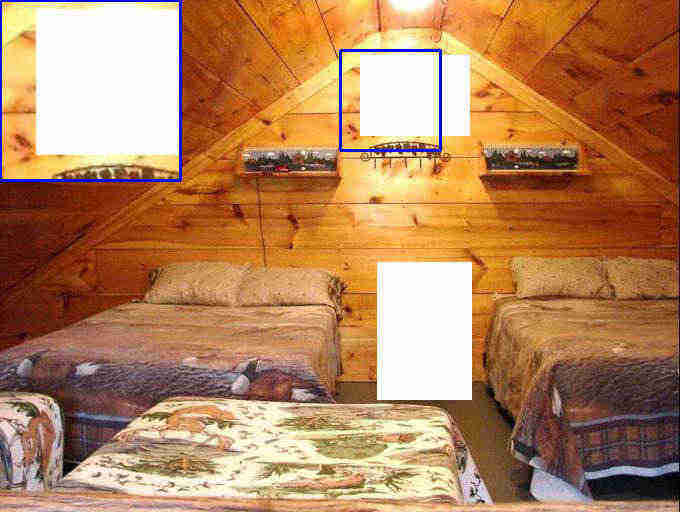}
input image
\end{minipage}
\begin{minipage}[t]{0.24\textwidth}
\centering
\includegraphics[width=\linewidth]{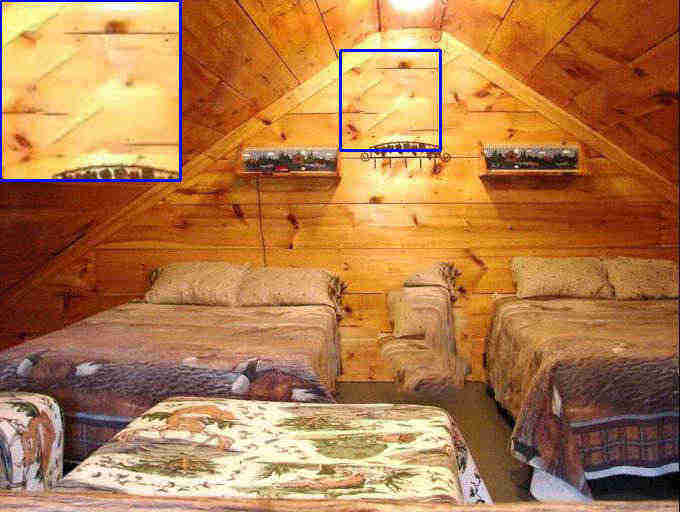}
Content-Aware Fill~\cite{barnes2009patchmatch}
\end{minipage}
\begin{minipage}[t]{0.24\textwidth}
\centering
\includegraphics[width=\linewidth]{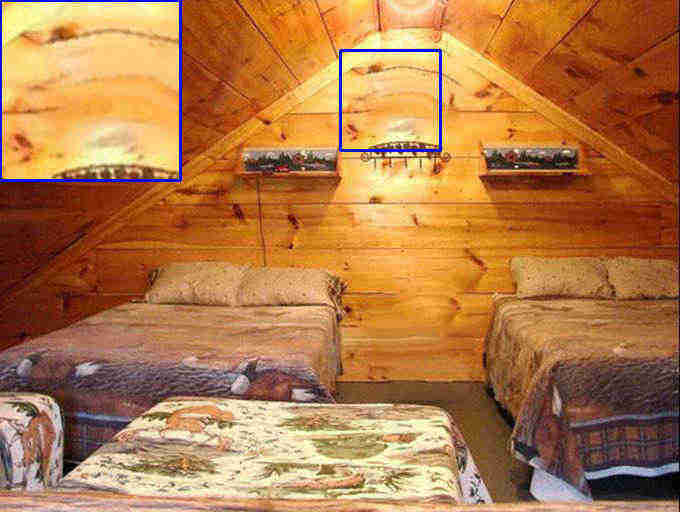}
ImageMelding~\cite{darabi2012image}
\end{minipage}
\end{minipage}

\begin{minipage}[t]{\textwidth}
\begin{minipage}[t]{0.24\textwidth}
\centering
\includegraphics[width=\linewidth]{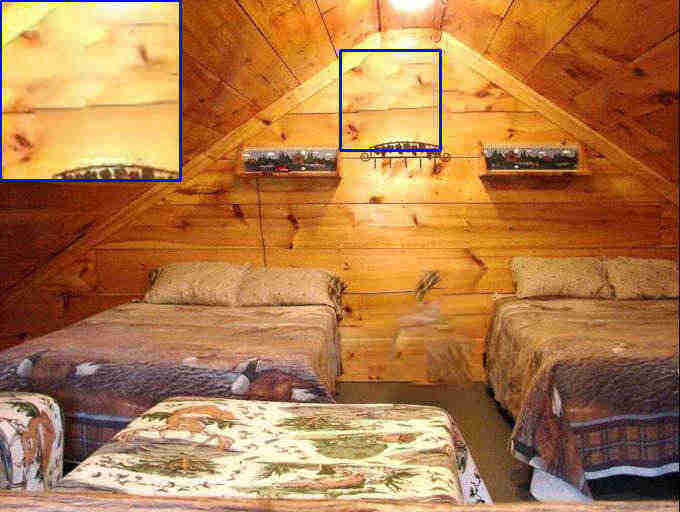}
StructCompletion~\cite{huang2014image}
\end{minipage}
\begin{minipage}[t]{0.24\textwidth}
\centering
\includegraphics[width=\linewidth]{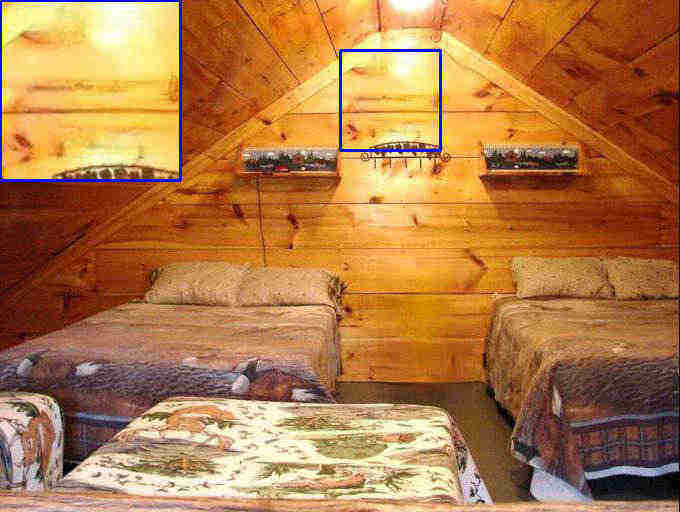}
our baseline model
\end{minipage}
\begin{minipage}[t]{0.24\textwidth}
\centering
\includegraphics[width=\linewidth]{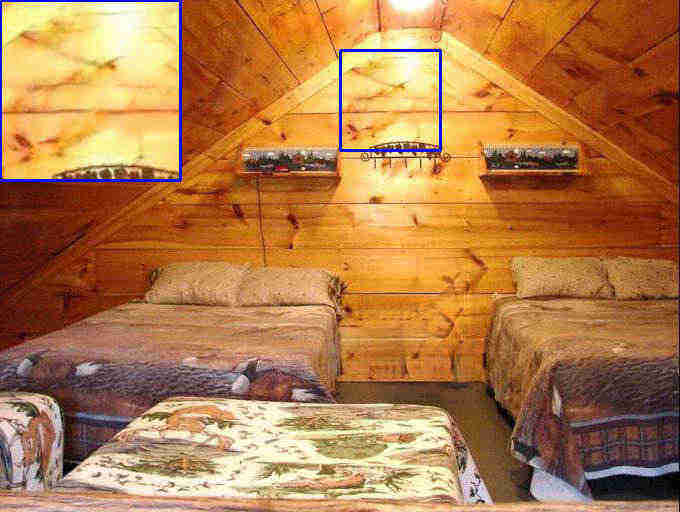}
our full model
\end{minipage}
\begin{minipage}[t]{0.24\textwidth}
\centering
\includegraphics[width=\linewidth]{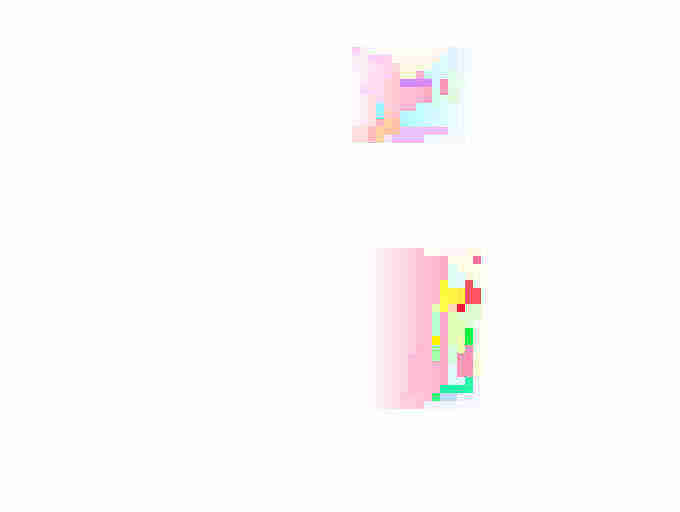}
attention map of full model
\end{minipage}
\end{minipage}

\begin{minipage}[t]{\textwidth}
\begin{minipage}[t]{0.24\textwidth}
\centering
\includegraphics[width=\linewidth]{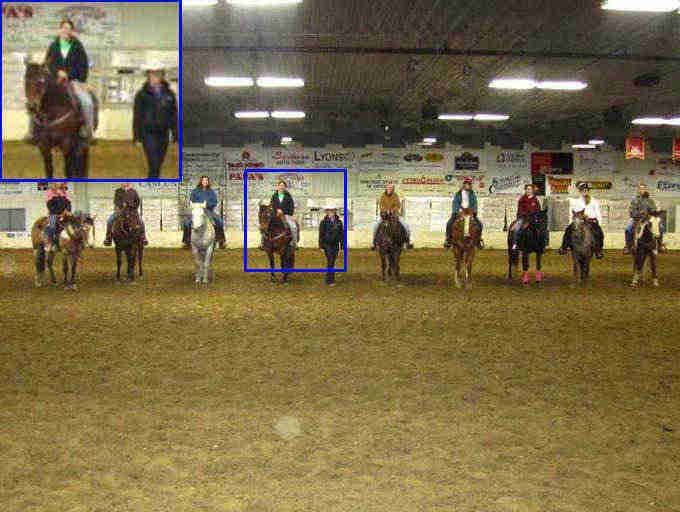}
original image
\end{minipage}
\begin{minipage}[t]{0.24\textwidth}
\centering
\includegraphics[width=\linewidth]{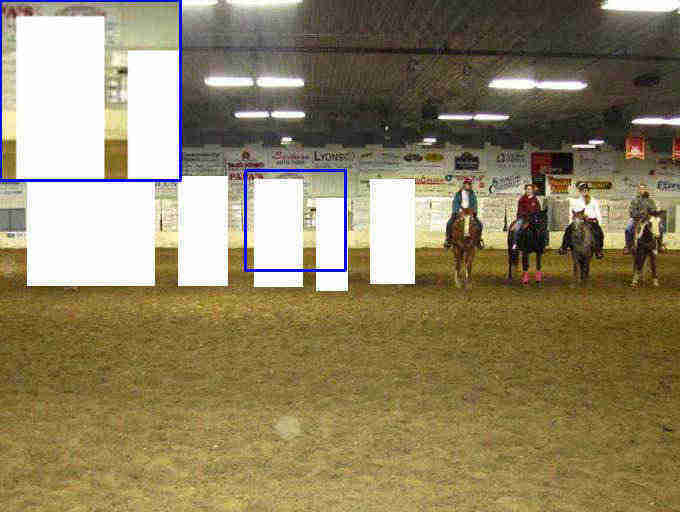}
input image
\end{minipage}
\begin{minipage}[t]{0.24\textwidth}
\centering
\includegraphics[width=\linewidth]{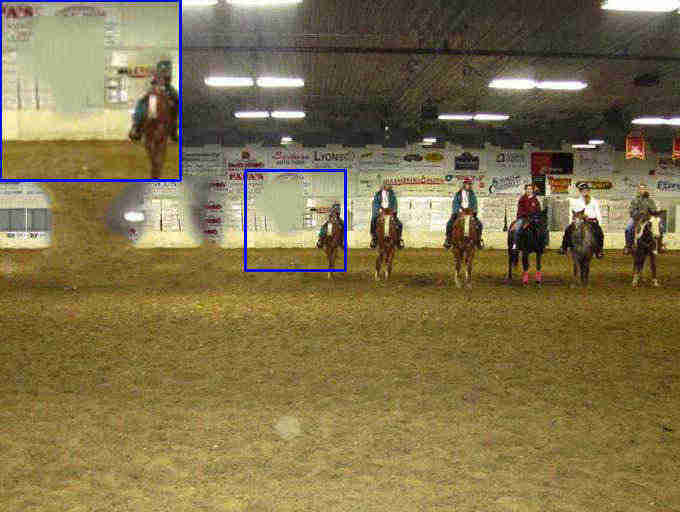}
Content-Aware Fill~\cite{barnes2009patchmatch}
\end{minipage}
\begin{minipage}[t]{0.24\textwidth}
\centering
\includegraphics[width=\linewidth]{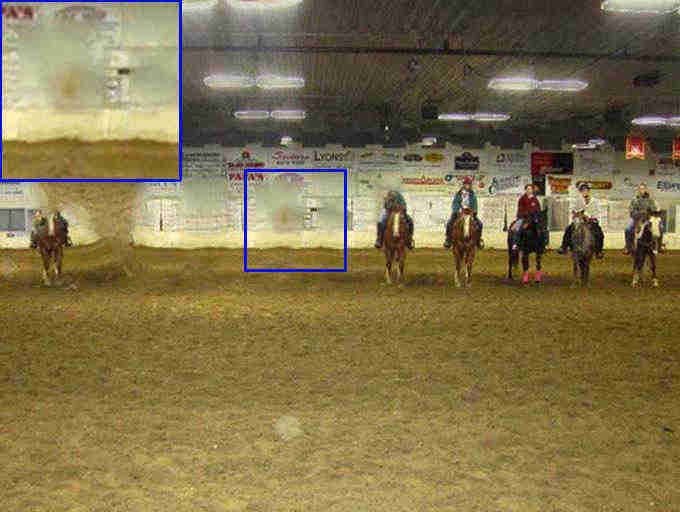}
ImageMelding~\cite{darabi2012image}
\end{minipage}
\end{minipage}

\begin{minipage}[t]{\textwidth}
\begin{minipage}[t]{0.24\textwidth}
\centering
\includegraphics[width=\linewidth]{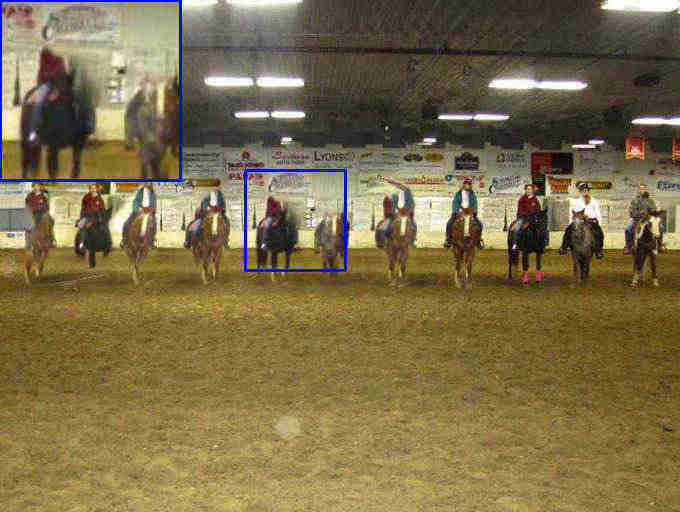}
StructCompletion~\cite{huang2014image}
\end{minipage}
\begin{minipage}[t]{0.24\textwidth}
\centering
\includegraphics[width=\linewidth]{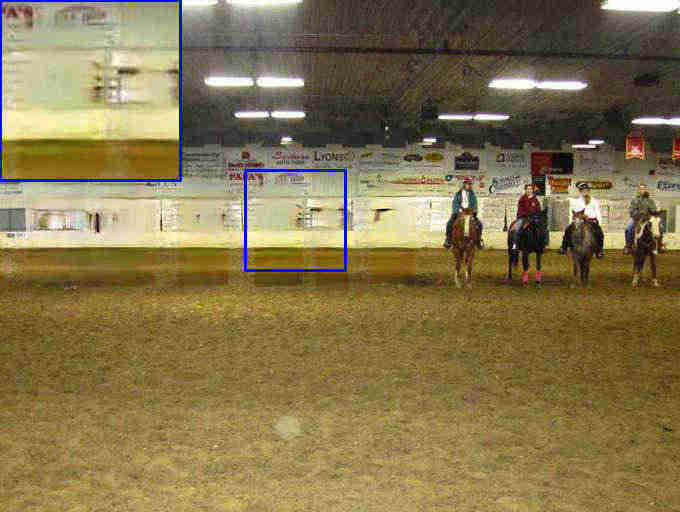}
our baseline model
\end{minipage}
\begin{minipage}[t]{0.24\textwidth}
\centering
\includegraphics[width=\linewidth]{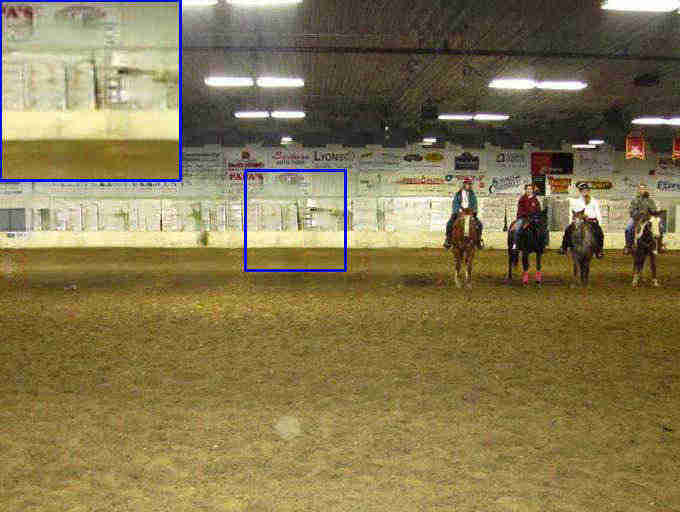}
our full model
\end{minipage}
\begin{minipage}[t]{0.24\textwidth}
\centering
\includegraphics[width=\linewidth]{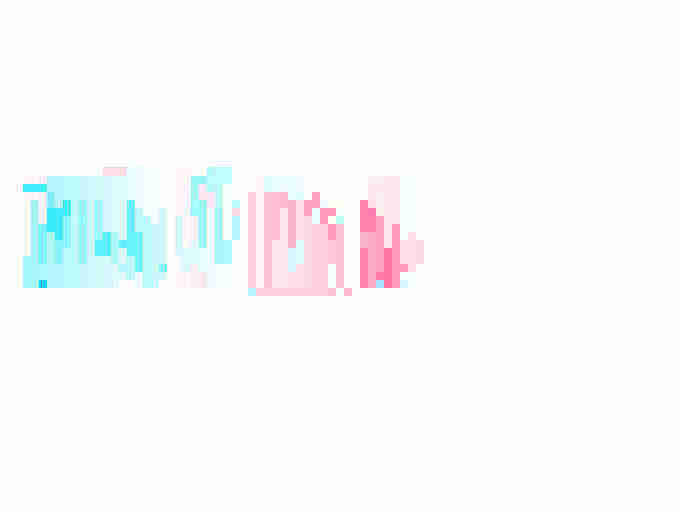}
attention map of full model
\end{minipage}
\end{minipage}

\begin{minipage}[t]{\textwidth}
\begin{minipage}[t]{0.24\textwidth}
\centering
\includegraphics[width=\linewidth]{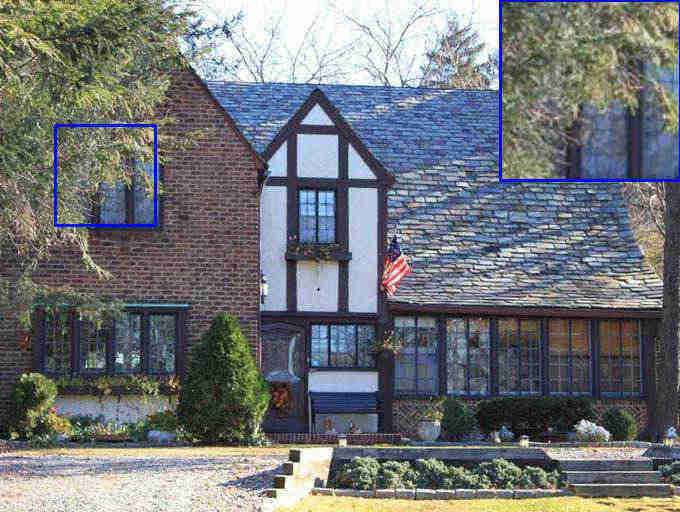}
original image
\end{minipage}
\begin{minipage}[t]{0.24\textwidth}
\centering
\includegraphics[width=\linewidth]{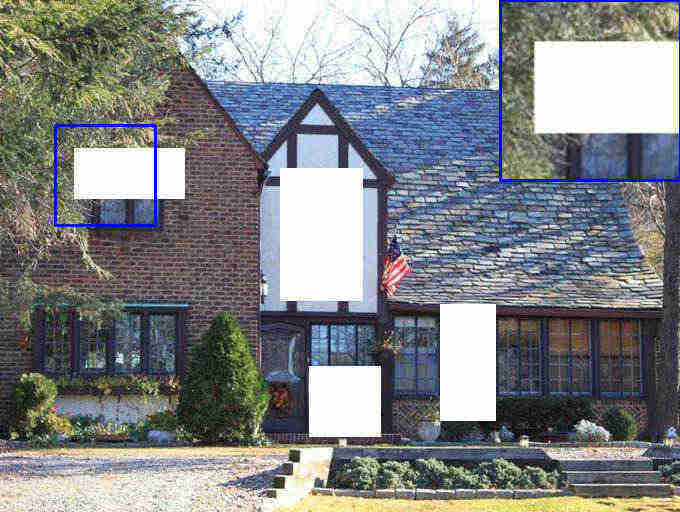}
input image
\end{minipage}
\begin{minipage}[t]{0.24\textwidth}
\centering
\includegraphics[width=\linewidth]{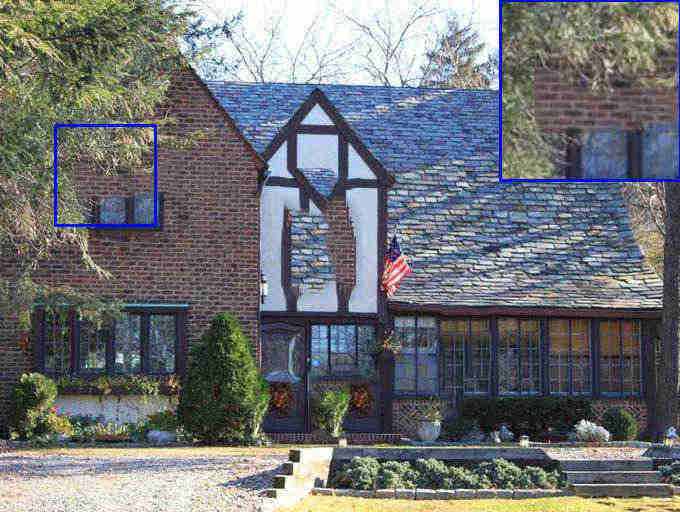}
Content-Aware Fill~\cite{barnes2009patchmatch}
\end{minipage}
\begin{minipage}[t]{0.24\textwidth}
\centering
\includegraphics[width=\linewidth]{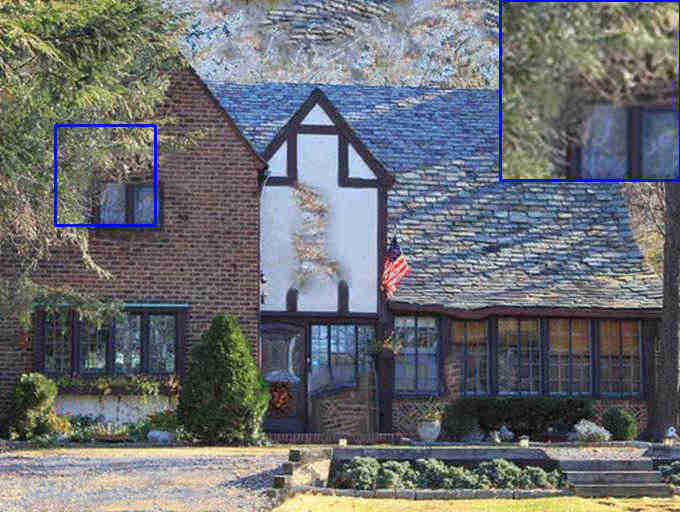}
ImageMelding~\cite{darabi2012image}
\end{minipage}
\end{minipage}

\begin{minipage}[t]{\textwidth}
\begin{minipage}[t]{0.24\textwidth}
\centering
\includegraphics[width=\linewidth]{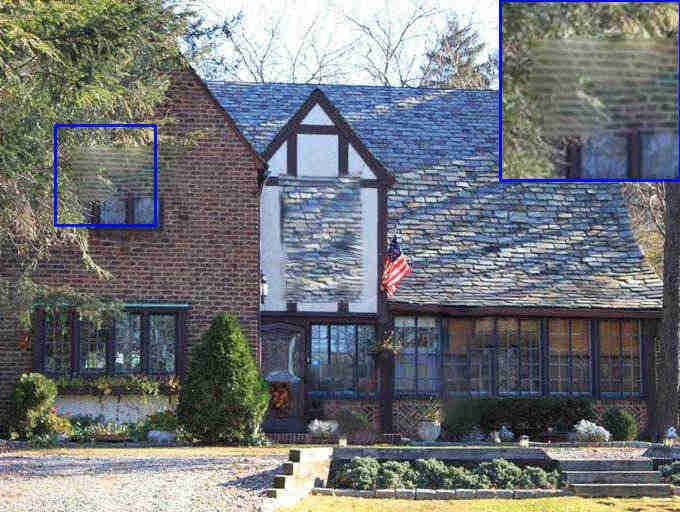}
StructCompletion~\cite{huang2014image}
\end{minipage}
\begin{minipage}[t]{0.24\textwidth}
\centering
\includegraphics[width=\linewidth]{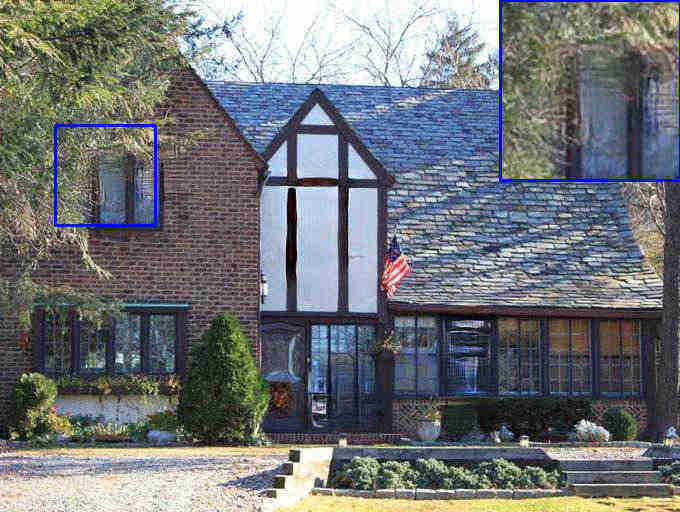}
our baseline model
\end{minipage}
\begin{minipage}[t]{0.24\textwidth}
\centering
\includegraphics[width=\linewidth]{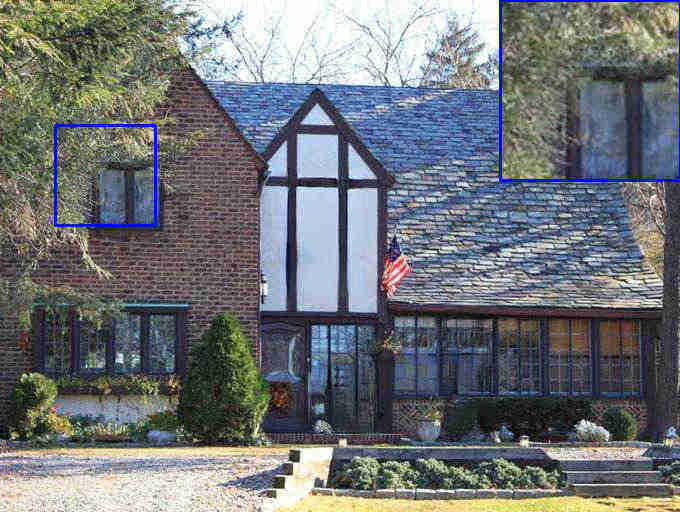}
our full model
\end{minipage}
\begin{minipage}[t]{0.24\textwidth}
\centering
\includegraphics[width=\linewidth]{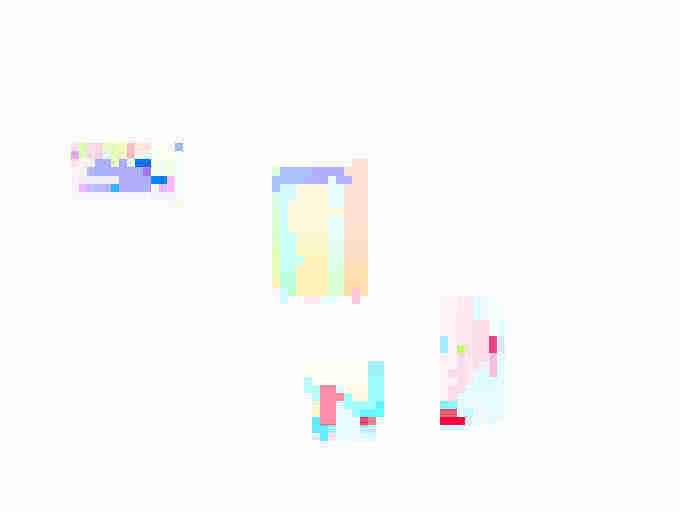}
attention map of full model
\end{minipage}
\end{minipage}

\caption{More qualitative results and comparisons.
All input images are masked from validation set. All our results are direct outputs from same trained model without post-processing. Best viewed with zoom-in.}
\label{fig:supp_places2_comparison2}
\end{figure*}

We show more results for qualitative comparisons with more methods including Photoshop Content-Aware Fill~\cite{barnes2009patchmatch}, Image Melding~\cite{darabi2012image} and StructCompletion~\cite{huang2014image} in Figure~\ref{fig:supp_places2_comparison1} and~\ref{fig:supp_places2_comparison2}. For all these methods, we use default hyper-parameter settings.

\section{More Visualization with Case Study}
\begin{figure*}[t!]
\centering

\includegraphics[width=0.33\linewidth]{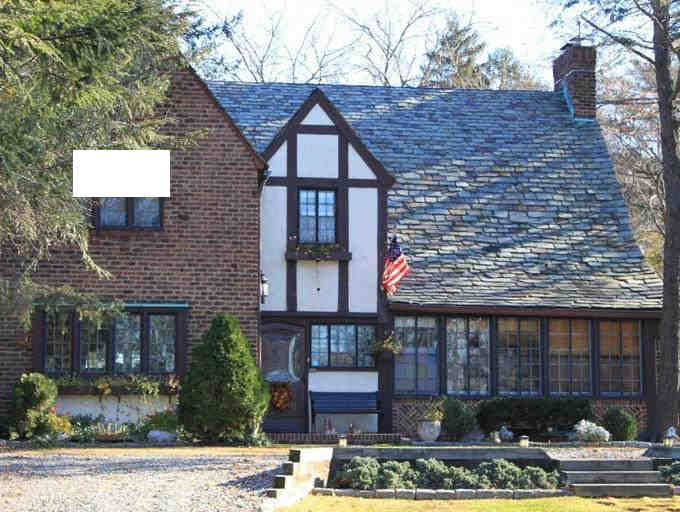}
\includegraphics[width=0.33\linewidth]{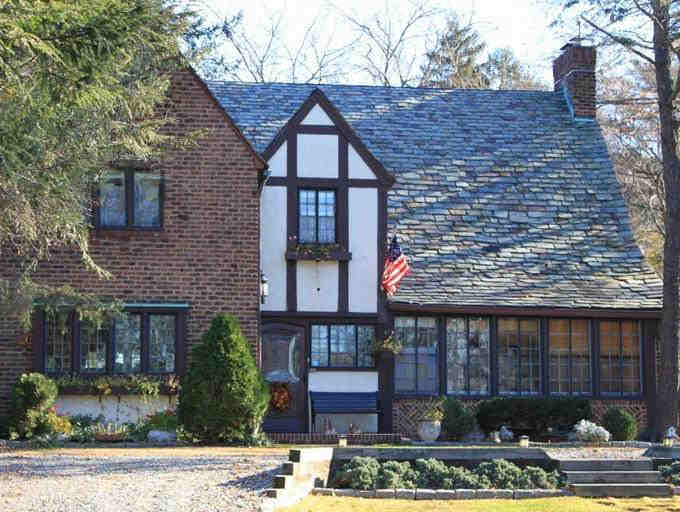}
\includegraphics[width=0.33\linewidth]{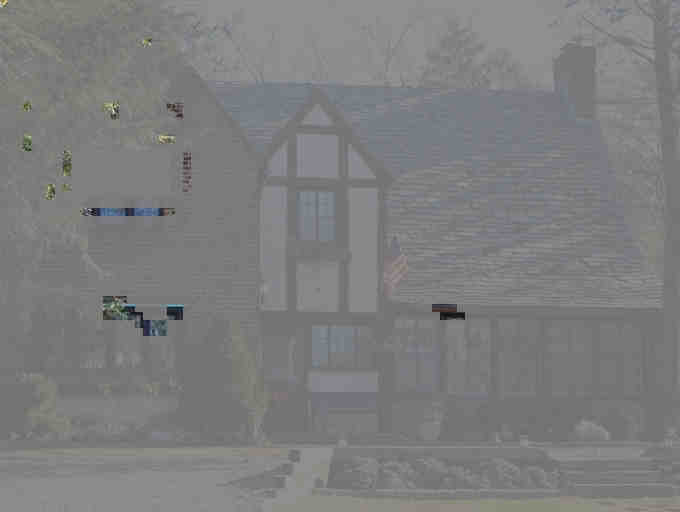}

\hspace{0.mm}

\includegraphics[width=0.33\linewidth]{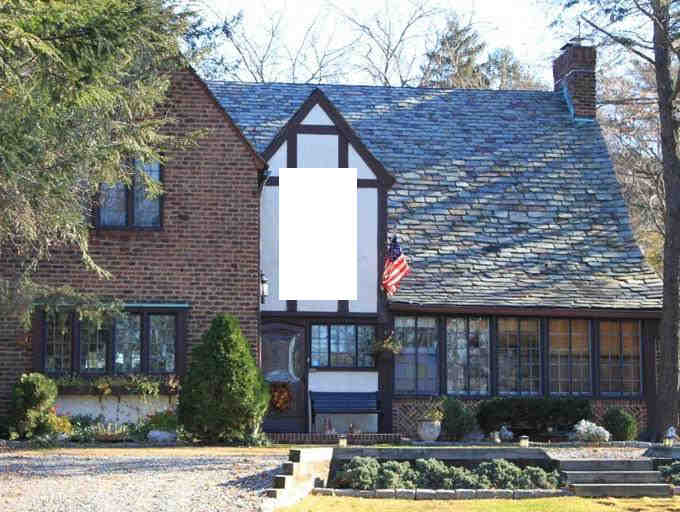}
\includegraphics[width=0.33\linewidth]{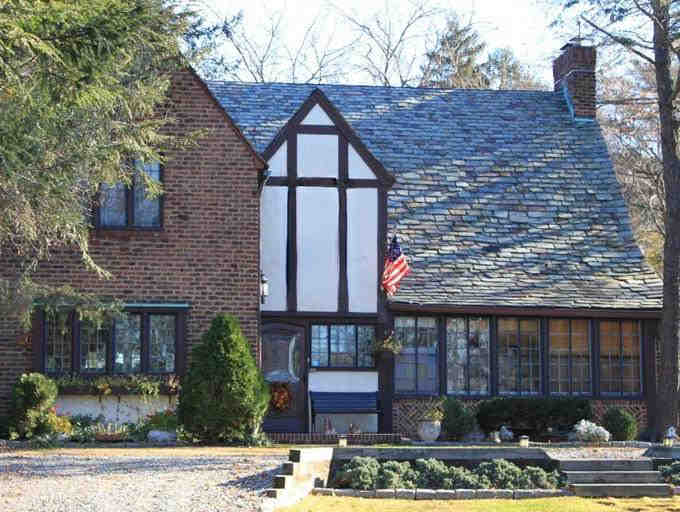}
\includegraphics[width=0.33\linewidth]{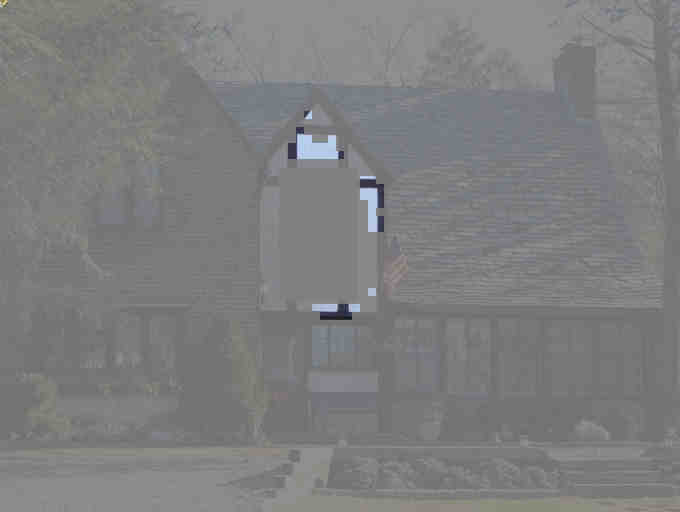}

\includegraphics[width=0.107\linewidth]{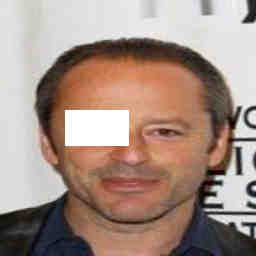}
\includegraphics[width=0.107\linewidth]{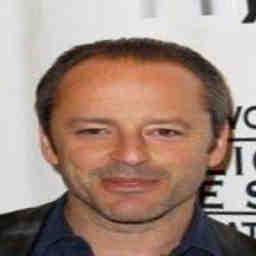}
\includegraphics[width=0.107\linewidth]{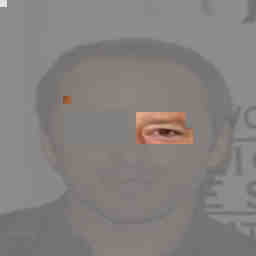}
\includegraphics[width=0.107\linewidth]{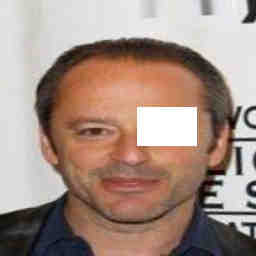}
\includegraphics[width=0.107\linewidth]{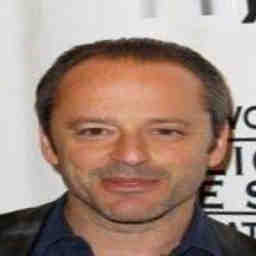}
\includegraphics[width=0.107\linewidth]{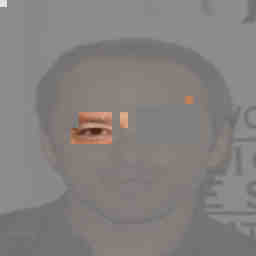}
\includegraphics[width=0.107\linewidth]{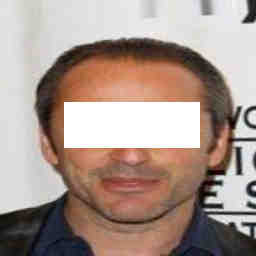}
\includegraphics[width=0.107\linewidth]{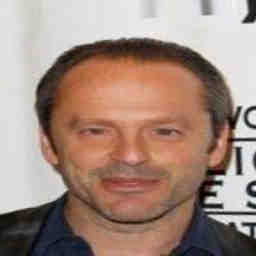}
\includegraphics[width=0.107\linewidth]{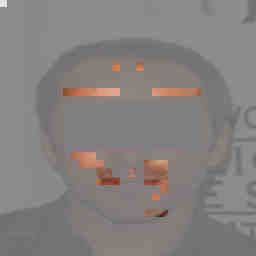}

\caption{Visualization (highlighted regions) on which parts in input image are attended. Each triad, from left to right, shows input image, result and attention visualization.}
\label{fig:supp_visualization}
\end{figure*}

In addition to attention map visualization, we visualize which parts in the input image are being attended for pixels in holes. To do so, we highlight the regions that have the maximum attention score and overlay them to input image. As shown in Figure~\ref{fig:supp_visualization}, the visualization results given holes in different locations demonstrate the effectiveness of our proposed contextual attention to borrow information at distant spatial locations.

\section{Network Architectures}
In addition to Section 3, we report more details of our network architectures. For simplicity, we denote them with K (kernel size), D (dilation), S (stride size) and C (channel number).

\textbf{Inpainting network}
Inpainting network has two encoder-decoder architecture stacked together, with each encoder-decoder of network architecture:

K5S1C32 - K3S2C64 - K3S1C64 - K3S2C128 - K3S1C128 - K3S1C128 - K3D2S1C128 - K3D4S1C128 - K3D8S1C128 - K3D16S1C128 - K3S1C128 - K3S1C128 - resize (\(2 \times\)) - K3S1C64 - K3S1C64 - resize (\(2 \times\)) - K3S1C32 - K3S1C16 - K3S1C3 - clip.

\textbf{Local WGAN-GP critic}
We use Leaky ReLU with \(\alpha = 0.2\) as activation function for WGAN-GP critics.

K5S2C64 - K5S2C128 - K5S2C256 - K5S2C512 - fully-connected to 1.

\textbf{Global WGAN-GP critic}
K5S2C64 - K5S2C128 - K5S2C256 - K5S2C256 - fully-connected to 1.

\textbf{Contextual attention branch}
K5S1C32 - K3S2C64 - K3S1C64 - K3S2C128 - K3S1C128 - K3S1C128 - contextual attention layer - K3S1C128 - K3S1C128 - concat.



\end{document}